\pdfoutput=1
\PassOptionsToPackage{numbers, sort&compress}{natbib}  % https://tex.stackexchange.com/questions/641619/sortcompress-is-not-working
\documentclass{article}

\usepackage[preprint]{neurips_2023}

\usepackage[utf8]{inputenc} % allow utf-8 input
\usepackage[T1]{fontenc}    % use 8-bit T1 fonts
\usepackage{hyperref}       % hyperlinks
\usepackage{url}            % simple URL typesetting
\usepackage{booktabs}       % professional-quality tables
\usepackage{amsfonts}       % blackboard math symbols
\usepackage{nicefrac}       % compact symbols for 1/2, etc.
\usepackage{microtype}      % microtypography
\usepackage{xcolor}         % colors

\usepackage{graphicx}
\usepackage{caption}
\usepackage{subcaption}
\usepackage{gensymb}
\usepackage{amsmath}
\usepackage{bbm}
\usepackage{bm}
\newcommand{\mnist}{\textsc{mnist}}
\newcommand{\emnist}{\textsc{emnist}}
\newcommand{\cifar}{\textsc{cifar-}\oldstylenums{10}}
\newcommand{\facescrub}{\textsc{facescrub}}

\usepackage[toc,page,header]{appendix}
\usepackage{minitoc}
\title{Modularity Trumps Invariance \\ for Compositional Robustness}

\author{
  Ian Mason \\%\thanks{Use footnote for providing further information
  Massachusetts Institute of Technology \\
  \texttt{imason@mit.edu} \\
  \And
  Anirban Sarkar \\
  Massachusetts Institute of Technology \\
  \texttt{anirbans@mit.edu} \\
  \And
  Tomotake Sasaki \\
  Artificial Intelligence Laboratory \\
  Fujitsu Limited \\
  \texttt{tomotake.sasaki@fujitsu.com} \\
  \And
  Xavier Boix \\
  Massachusetts Institute of Technology \\
  Fujitsu Research of America, Inc.\\
  \texttt{xboix@fujitsu.com}
}

\begin{document}
\doparttoc % Tell to minitoc to generate a toc for the parts
\faketableofcontents % Run a fake tableofcontents command for the partocs
\part{}\vspace{-10mm}

\maketitle

\begin{abstract}
  % V3
  By default neural networks are not robust to changes in data distribution. This has been demonstrated with simple image corruptions, such as blurring or adding noise, degrading image classification performance. Many methods have been proposed to mitigate these issues but for the most part models are evaluated on single corruptions. In reality, visual space is compositional in nature, that is, that as well as robustness to elemental corruptions, robustness to compositions of corruptions is also needed. In this work we develop a compositional image classification task where, given a few elemental corruptions, models are asked to generalize to compositions of these corruptions. That is, to achieve \emph{compositional robustness}. %The elemental corruptions form in-distribution training domains and the compositions form out-of-distribution test domains.
  We experimentally compare empirical risk minimization with an invariance building pairwise contrastive loss and, counter to common intuitions in domain generalization, achieve only marginal improvements in compositional robustness by encouraging invariance. To move beyond invariance, following previously proposed inductive biases that model architectures should reflect data structure, we introduce a modular architecture whose structure replicates the compositional nature of the task. We then show that this modular approach consistently achieves better compositional robustness than non-modular approaches. We additionally find empirical evidence that the degree of invariance between representations of `in-distribution' elemental corruptions fails to correlate with robustness to `out-of-distribution' compositions of corruptions.
  % Add a concluding remark? These findings provide more evidence for modules and more questions for invariance??
\end{abstract}

% https://esajournals.onlinelibrary.wiley.com/doi/full/10.1002/bes2.1258
% https://www.nature.com/articles/d41586-018-02404-4
% Could look for Successful Scientific Writing. Matthews, Bowen and Matthews. http://eugene.yakovis.com/doc/Matthews%20Bowen%20Matthews%202000%20(raw%20OCR).pdf

\section{Introduction}
\vspace{-2mm}

Biologically intelligent systems show a remarkable ability to generalize beyond their training stimuli, that is to learn new concepts from no, or few, examples by combining previously learned concepts \citep{ito2022compositional, lake2019human, piantadosi2016compositional, shulz2016probing}. In contrast, artificial neural networks are surprisingly brittle, failing to recognize known categories when presented with images with fairly minor corruptions \citep{dodge2017study, geirhos2021partial, hosseini2017google, hendrycks2019benchmarking, jang2021noise}. To improve robustness many methods have been proposed for learning more robust representations, including data augmented training techniques \citep{hendrycks2020augmix, hendrycks2021many, jang2021noise, yun2019cutmix, zhang2018mixup}, and encouraging invariant representations or predictions \citep{huang2022robustness, kim2019learning, sinha2021consistency, von2021self}. 

\begin{figure*}
    \centering
    \includegraphics[trim={0 17cm 0 0}, width=\textwidth]{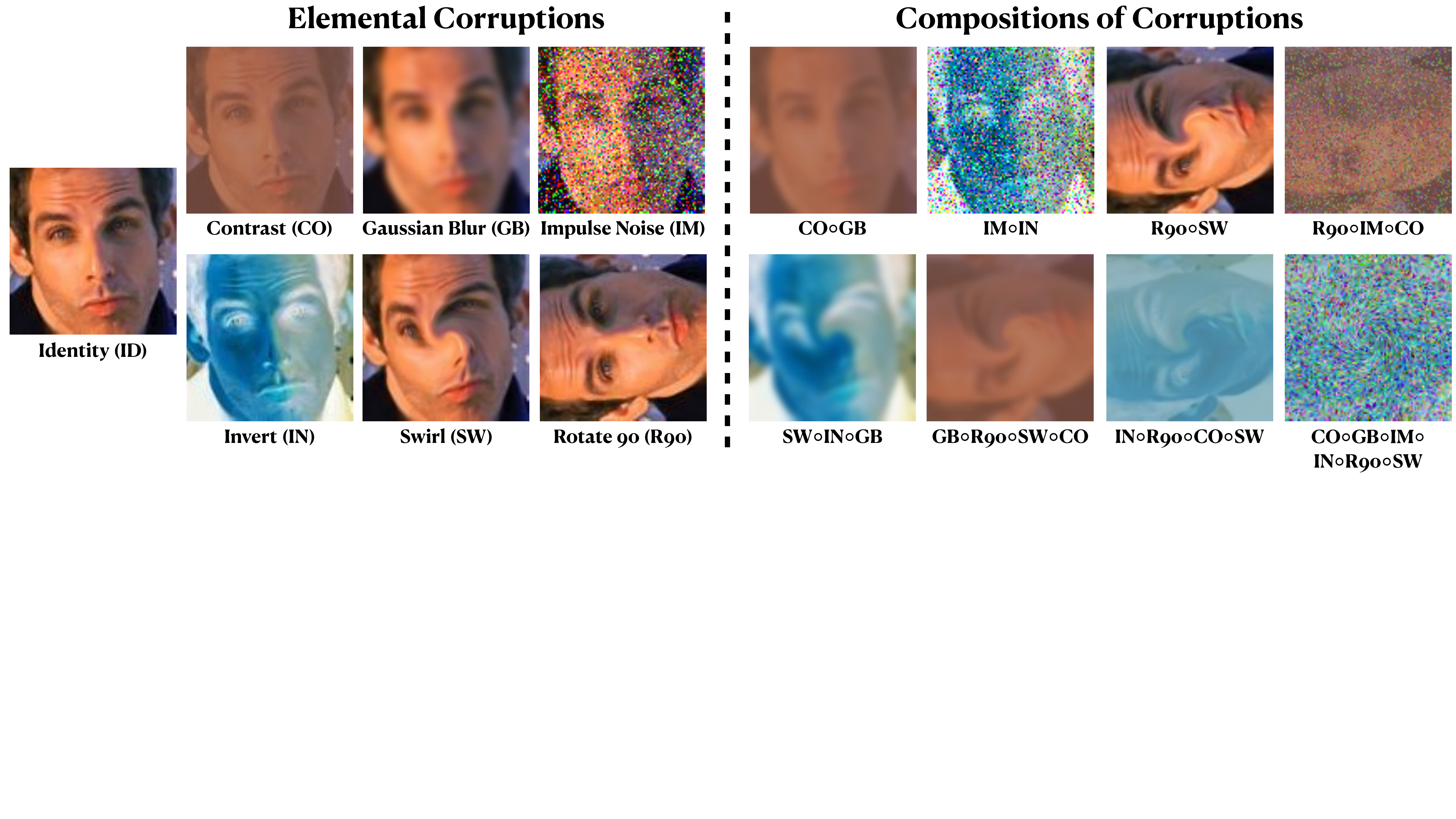}
    \vspace{-3.5mm}
    \caption{The compositional robustness task. A model is trained jointly on images corrupted with elemental corruptions (left) and evaluated on images corrupted with compositions of these corruptions (right). Shown is \emph{all} $7$ elemental corruptions and a \emph{subset} of the $160$ compositions of corruptions.}
    \vspace{-4.5mm}
    \label{fig:benchmark-facescrub}
\end{figure*}

However, when the robustness of these methods is evaluated it tends to be on single corruptions of the type seen in ImageNet-C \citep{hendrycks2019benchmarking}. In reality, the space of possible corruptions is compositional. If we draw a loose correspondence between corruptions and real world weather conditions, with noise akin to rain on a windshield, blur as fog and a contrast change as a change in brightness, we see it is in fact possible to have rain, fog and bright sun simultaneously. In this work we extend the notion of robustness over corruptions to robustness over \emph{compositions of corruptions}. We construct a compositional image classification task where a neural network is trained on single \emph{elemental corruptions} and evaluated on \emph{compositions} of these corruptions (Figure~\ref{fig:benchmark-facescrub}). Importantly, this is not an adversarial or no-free-lunch task, as we want the AI systems we develop to be capable of compositional generalization \citep{bahdanau2019systematic, chalmers1990fodor, fodor1988connectionism, goyal2022inductive, lake2018generalization, lake2017building, mendez2022review}. 

If natural visual data can be decomposed into a set of elemental functions (or mechanisms \citep{peters2017elements, parascandolo2018learning}), we do not yet know how to find them. The compositional robustness task we create allows us to experiment with a compositional structure where the underlying elemental functions are known. By studying the behaviors of neural networks under this structure, we aim to gain insights into how we might develop methods for better compositional robustness. Such insights could be applied to create systems that generalize more robustly or allow for lower data collection costs, needing only to collect or synthesize the elemental corruptions instead of the exponentially large number of compositions. Finally, this task creates a new domain generalization task on which we can evaluate the generality of proposed methods for domain generalization. In domain generalization parlance, a system is trained on data from multiple training domains (the elemental corruptions), and then evaluated on data from a related set of test domains (the compositions), from which no data samples are seen during training. 

To better understand how neural networks behave on out-of-distribution compositional data we evaluate different methods for domain generalization on this task. Firstly, we explore empirical risk minimization (ERM), which has been shown to be a strong baseline when correctly tuned \citep{gulrajani2021search}. Secondly, we evaluate a setup where invariance between the same image under different corruptions is explicitly encouraged using the contrastive loss \citep{chen2020simclr, gutmann2010nce, hadsell2006dimensionality}, since a central theme in domain generalization has been to encourage the learning of invariant representations \citep{ahmed2021systematic, albuquerque2019generalizing, arjovsky2019irm, dou2019domain, li2018deep, ghifary2015domain, kim2021selfreg, li2018domain, motiian2017unified, sakai2022three, creager2021environment}. Finally, we introduce a modular architecture to better reflect the compositional structure of the task \citep{pfeiffer2023modular}. Here, rather than all parameters jointly modelling all corruptions, each elemental image corruption is `undone' by a separate module in latent space.

Counter to our initial expectations we find that training to encourage invariant representations with the contrastive loss offers only minor improvements in terms of out-of-distribution accuracy, whilst the modular architecture consistently outperforms other methods. Additionally, we find that the degree of invariance between representations of elemental corruptions fails to correlate with performance on out-of-distribution compositions of corruptions. At their narrowest interpretation, these results empirically show that for compositional robustness, when training domains consist only of the elemental components, modular approaches tend to outperform monolithic (non-modular) approaches. At their broadest interpretation our results question whether encouraging non-trivially\footnote{The trivial case with constant representations has maximal invariance but cannot achieve good generalization.} invariant representations is sufficient to achieve compositional domain generalization. This indicates that there is still work to be done on understanding the additional properties required for compositional robustness and suggests more modular architectures as a promising candidate for one such property. % such as uncovering the neural mechanisms that lead to invariant representations in neural networks and the brain. 

% \paragraph{Paper overview and approach??}
% This paper proceeds as follows...

% The idea of this paper isn't that this is the best way to `solve vision' in general, rather that by studying the compositional case in a controlled manner we may be able to gain insights into how we might be able to compose mechanisms, and develop methods for better composition which can then hopefully be applied to create systems that generalise more robustly.

% What questions do we ask. What answers do we give.
% Questions: Do neural networks compositionally generalise. What mechanisms might allow for compositional generalization? How might we improve compositional generalisation?

\vspace{-2mm}
\section{Related Work}
\vspace{-2mm}
% Not sure but something like: whilst in general it is impossible to get the causal graph from data, Kun Zhang has results that it is possible to determine if you make some assumptions. However, rather than determining the graph, this work asks, if I know exactly what my mechanisms are, how should I model them (and is evaluated in terms of performance on novel compositions). Inspired by Goyal and Bengio we take the approach that they should be spare and at the appropriate level of abstraction.

% One solution/only solution(?) for compositionality is invariance for the composed input. Can argue intuitively, or on invariance in bio systems or domain generalisation (IRM etc?). Final sentence of paragraph - strong evidence for presence of selectivity & invariance  in bio systems (hubel and wiesel etc.)

% In some sense the question going round and round in circles is can we expect a meta-distribution as in ProbDG or do we want to learn structure some other way?
% Where some other way might be "In object recognition, the “visual world” can be considered as decomposing into views (e.g. perspectives or lighting conditions)". From: https://arxiv.org/pdf/1508.07680.pdf

We now briefly recap related works from the areas of domain generalization, invariant representations, modularity, compositional generalization and robustness.

\textbf{Domain Generalization and Invariant Representations.}  % Two paras
The creation of models that are robust to unseen changes in data distribution is the work of domain generalization. Given certain training domains, the aim of domain generalization is to build models that can generalize to related unseen test domains. One common approach is to encourage the learning of invariant representations between training domains whilst achieving high performance \citep{ahmed2021systematic, albuquerque2019generalizing, arjovsky2019irm, dou2019domain, ghifary2015domain, kim2021selfreg, li2018domain, li2018deep, motiian2017unified, sakai2022three, creager2021environment}, with the idea that this will lead to invariant representations between training and test domains and hence good generalization performance. However, this relies on an implicit assumption that we have sufficient training domains that are reasonably representative samples from some meta-distribution of domains (this has been made explicit in some works \cite{krueger2021out, eastwood2022probable}). It is not clear that this will be true in general, and arguably replaces the problematic assumption of \emph{i.i.d} data with an equally problematic assumption of \emph{i.i.d} domains. What's more, such generalist approaches may be unable to take structure amongst training domains into account. It should be noted that there has also been substantial work on encouraging invariance for the related task of domain adaptation where (unlabelled) data from test domains is available \citep{ganin2016dann, tzeng2017adversarial, sun2016deepcoral, long2015learning, long2018cdan, eastwood2022sourcefree}. Despite being motivated by theoretical work \citep{ben2007domain, ben2010theory}, %and intuitions that invariant representations should give good performance, 
the central role of invariance in domain adaptation and generalization has been questioned \citep{zhao2019learning, johansson2019support, rosenfeld2021risks, shen2022connect, akuzawa2020adversarial}. In Section~\ref{sec:invariance} we discuss the limitations of encouraging invariance for compositional robustness.

\textbf{Relational Inductive Biases and Modularity.} % Equivariance, ICMs, Modularity, Disentanglement
A closely related approach to learning robust representations aims to take advantage of explicit structure in data. These relational inductive biases \citep{battaglia2018relational} aim to include knowledge about entities and the relations between them into neural network architectures. For example, we can encode that entities should not change under certain transformations by building invariance to these transformations into our architectures. Work on equivariance beyond translation explicitly creates such robustness \citep{cohen2016group, weiler2018learning, worrall2017harmonic} but is usually formulated in terms of group actions \citep{cohen2019general} so is limited to invertible transformations. More general approaches aim to uncover structure by decomposing data into independent (causal) mechanisms \citep{peters2017elements, parascandolo2018learning, scholkopf2021towards, goyal2021recurrent, goyal2022inductive} or disentangled factors of variation \citep{chen2016infogan, higgins2017beta, kim2018disentangling, roth2023disentanglement, eastwood2018framework, locatello2019challenging, schott2021visual, montero2021role, montero2022lost}. Ways to explicitly model decomposable structures in data include pre-training on primitive components \citep{ito2022compositional} % Also Amir Rahimi's paper on simple questions 
and using modular architectures to encode structure \citep{jaderberg2015spatial, andreas2016learning, andreas2016neural, damario2021modular, bahdanau2019systematic, goyal2021recurrent, goyal2022coordination, mendez2021lifelong, madan2022ood, carvalho2023composing, pfeiffer2023modular}. In contrast, in this work we know how the data structure decomposes and explore the performance of modular and non-modular architectures on the recomposition of known elemental components.

% However, there has been limited research into how these models behave on compositional data and little is known about the invariances learned by such systems.

% Compositions and other types of generalisation for factors of variation. Motivate why it is needed.
\textbf{Compositional Generalization.} The visual world is compositional \citep{bahdanau2019closure, lake2015human, lake2017building, romaszko2017vision, krishna2017visual}. Whilst much has been made of compositionality in language (linguistic compositionality) and reasoning (conceptual compositionality)
\citep{andreas2016learning, battaglia2018relational, furrer2020compositional, hu2017learning, johnson2017clevr, lake2018generalization, lake2017building, livska2018memorize, mendez2022review, qiu2022improving, xie2022coat, schmidhuber1990towards}, compositional robustness has received relatively little attention. Recent AI systems still fail on compositional tasks \citep{keysers2020measuring, lake2018generalization, schott2021visual, van2004lack} where the space of generalization grows exponentially with the number of elemental components. Whilst practically it is not possible to sample all combinations of elemental components, one interpretation of large models~\citep{geirhos2021partial, kaplan2020scaling, radford2021learning} is that they aim to sample densely enough to generalize to unseen combinations. However, for real world data, it is unclear how big the compositional space is and how densely we need to sample, with this being particularly pertinent if the data distribution is high-dimensional \citep{geiger2020perspective, wainwright2019high}
% geiger - section 1.2. Waingwright section 1.2.3.
or fat tailed \citep{taleb2020statistical}.
% taleb - chapter 3
To that end, several works have analyzed controlled settings, aiming to understand the best settings for training in order to achieve the best generalization \citep{ahmed2021systematic, cooper2021out, schott2021visual}.

% The fundamental research question this work is concerned with is how it is possible to generalize over such compositional spaces. 

% Informally a problem with scale is that we can never see everything because fat tails: https://twitter.com/filippie509/status/1559583211612622848
% I suspect the figures in the tweet thread are from Taleb https://www.youtube.com/watch?v=nDY_fh2TVlI

% Possible caveat: the answer is probably somewhere in the middle, you don't see only independent mechanisms and then have to compose them, you do see some combinations. We take the extreme view because it is under-explored and it is easier to disentangle what is scale and what is composition by only looking at composition. By applying learned lessons from studying exclusively compositional generalisation the idea is to better generalize/transfer knowledge to new combinations & classes (evaluated using methods like Avi Cooper's) and thus work better when used at scale (less data, better generalisation). [Going back along the arrow compositional generalization -> scale. Methods like CLIP aim to go the other way along the arrow, with enough scale -> compositional generalisation.

\textbf{Robustness Over Image Corruptions.} 
Whilst the aforementioned work aims to improve the robustness of neural networks, many have worked specifically on improving robustness for common image corruptions and adversarial examples \citep{hendrycks2019benchmarking, jang2021noise, hendrycks2020augmix, hendrycks2021many, yun2019cutmix, zhang2018mixup, huang2022robustness, kim2019learning, sinha2021consistency, von2021self, baidya2021combining}. However, the majority of previous works are evaluated only on single corruptions, ignoring the true compositional space formed by the corruptions. 

\vspace{-2mm}
\section{Methods}
\label{sec:methods}
\vspace{-2mm}

\subsection{A Framework for Evaluating Compositional Robustness}
\label{sec:setup}
\vspace{-2mm}

We design a framework for evaluating compositional robustness on any dataset for image classification. We first create elemental components by applying six different corruptions separately to all images. These corruptions along with the original, \emph{Identity (ID)}, data create $7$ training domains. We use the corruptions \emph{Contrast (CO)}, \emph{Gaussian Blur (GB)}, \emph{Impulse Noise (IM)}, \emph{Invert (IN)}, \emph{Rotate 90$\degree$ (R90)} and \emph{Swirl (SW)}, seen in Figure~\ref{fig:benchmark-facescrub} (left). We choose these corruptions to include a mixture of long-range and local effects as well as invertible and non-invertible corruptions. A further exploration of the choice and parameter settings of corruptions is given in Appendix~\ref{app:corruption-choice}. % Note that the motivation for developing this framework is to better understand how compositionality is handled in neural networks rather than to achieve a direct solution for these particular corruptions; some of the corruptions can be inverted with one line of code. % for example to undo the invert corruption we simply need to calculate $1 - pixel\_value$ for each pixel.

To test compositional robustness we create images from compositions of the elemental corruptions, see Figure~\ref{fig:benchmark-facescrub} (right).  We consider every possible permutation of compositions of two corruptions (excluding \emph{Identity}) giving $^6\!P_2 = 30$ possible compositions.  For compositions of more than two corruptions we sample the possible permutations to approximately balance the contributions of compositions containing different numbers of elemental corruptions (the sampling process is described in Appendix~\ref{app:corruption-choice}). This creates $40$ possible compositions of $3$ corruptions and $30$ possible compositions for each of $4$, $5$ and $6$ corruptions. Altogether the compositions form $160$ test domains. The task we then try to solve is to achieve the highest classification accuracy on images from the $160$ compositional test domains whilst training only on the $7$ elemental training domains.

\subsection{Monolithic Approaches}
\label{sec:monolithic}
\vspace{-2mm}
A domain generalization task consists of data from related domains or environments $\mathcal{D}_e = \{(\bm{x}_e^{(i)}, y_e^{(i)})\}_{i=1}^{N_e}$, with $e \in \mathcal{E}_{all}$, where $\mathcal{E}_{all}$ is the set of all domains we wish to generalize to and $N_e$ the number of datapoints in domain $e$. However, during training we only have access to a subset of domains $\mathcal{E}_{tr} \subset \mathcal{E}_{all}$. For our task, $\mathcal{E}_{tr}$ is the set of elemental training domains, $|\mathcal{E}_{tr}| = 7$, and $\mathcal{E}_{all}$ additionally includes the compositional test domains, $|\mathcal{E}_{all}| = 167$. As we use the same set of base images to create corrupted images, the number of datapoints, $N_e$, is the same across all domains.

For a neural network $f_{\bm{\theta}}$ parameterized by $\bm{\theta}$, we aim to find parameters, $\bm{\theta}^{*}$, from parameter space $\bm{\Theta}$, that optimize loss function $\mathcal{L}$, on training domains $\mathcal{E}_{tr}$. The accuracy of $f_{\bm{\theta^{*}}}$ is then evaluated on the test domains. Monolithic approaches share all parameters, $\bm{\theta}^{*}$, over all domains where,
\vspace{-2mm}
\begin{equation}
    \bm{\theta}^{*} = \underset{\bm{\theta} \in \bm{\Theta}} {\operatorname{argmin}}   \sum_{e \in \mathcal{E}_{tr}} \sum_{i=1}^{N_e} \mathcal{L}(f_{\bm{\theta}}(\bm{x}_e^{(i)}, y_e^{(i)})).
    \label{eq:domain-gen}
\end{equation}
The first approach we evaluate is Empirical Risk Minimization (ERM), training all parameters jointly to minimize some risk function over training domains. We set $\mathcal{L}$ to be the mean cross entropy loss.

% \begin{equation}
%     \mathcal{L}(f_{\theta}(x, y)) = \frac{1}{N} \sum_{c=1}^C -\mathbb{I}(y=c)\log(\sigma(f_{\theta}(x))_c,   % \sum - p_i \log q_i is cross entropy. p_i is one-hot label vector, q_i is prediction.
%     \label{eq:cross-entropy}
% \end{equation}

% with $\sigma$ representing the softmax operation, $N$ the total number of training examples, $C$ the number of categories, $\mathbb{I}$ an indicator function that is $1$ when $y=c$ and $0$ otherwise, and, slightly overloading notation, $x_c$ representing the $c^{th}$ entry in vector $x$.

% \paragraph{Contrastive Training}
The second approach we evaluate is contrastive training. A standard domain generalization approach is to encourage invariance between representations on the training domains \citep{zhou2022domain} and since we have paired data between domains we can explicitly encourage invariance using the contrastive loss \citep{chen2020simclr, gutmann2010nce, hadsell2006dimensionality}. Note that the availability of paired data creates a best-case set up for the learning of invariant representations and that learning a representation that is invariant for paired images from different domains would satisfy the invariance encouraging criteria of previous works \citep{arjovsky2019irm, li2018deep, dou2019domain}. 

We follow the SimCLR contrastive training formulation \citep{chen2020simclr}, taking $B$ datapoints from each elemental training domain (created from the same base images) to get a minibatch of size $B|\mathcal{E}_{tr}|$.  Applying an additional index to each of the domains in $\mathcal{E}_{tr}$ to get $\mathcal{E}_{tr} = \{e_d \}_{d=1}^{D}$, positive pairs come from pairs of the same image under different corruptions $(\bm{x}_{e_r}^{(i)}, \bm{x}_{e_s}^{(i)}), r \neq s$, and negative pairs from all other pairs in the minibatch $(\bm{x}_{e_r}^{(i)}, \bm{x}_{e_s}^{(j)}), i \neq j$. We apply the contrastive loss on representations from the penultimate layer of $f_{\bm{\theta}}$, notating the representation for $\bm{x}_{e}^{(i)}$ as $\bm{z}_{e}^{(i)}$. Using cosine similarity, $\text{sim}(\bm{u},\bm{v}) = \bm{u}^T \bm{v} / \|\bm{u}\|  \|\bm{v}\|$, to measure similarity between representations we define the loss for a positive pair in the minibatch as 
\vspace{-1mm}
\begin{equation}
    \ell(\bm{x}_{e_r}^{(i)},\bm{x}_{e_s}^{(i)}) = - \log \frac{\exp(\text{sim}(\bm{z}_{e_r}^{(i)}, \bm{z}_{e_s}^{(i)}) / \tau)}{\sum_{d=1}^{D}\sum_{k=1}^{B} \mathbbm{1}[k \neq i] \exp(\text{sim}(\bm{z}_{e_r}^{(i)}, \bm{z}_{e_d}^{(k)}) / \tau)}, 
    \label{eq:contrastive}
\end{equation}
where $\tau$ is a temperature parameter and $\mathbbm{1}[k \neq i]$ is an indicator function equal to $1$ when $k \neq i$ and $0$ otherwise. We compute this loss across all positive pairs in the minibatch to encourage invariant representations. To learn to classify, we additionally include the cross entropy loss to arrive at,
\vspace{-1mm}
\begin{equation}
    \mathcal{L}(f_{\bm{\theta}}(\bm{x}_{e_r}^{(i)}, y_{e_r}^{(i)})) = - \sum_{c=1}^C \mathbbm{1}[y_{e_r}^{(i)}=c]\log(\sigma(f_{\bm{\theta}}(\bm{x}_{e_r}^{(i)}))_c + \lambda \sum_{\substack{e_s \in \mathcal{E}_{tr} \\ s \neq r}} \ell(\bm{x}_{e_r}^{(i)},\bm{x}_{e_{s}}^{(i)}).
    \label{eq:ce-contrastive}
\end{equation}
Here the first term is the cross entropy loss, with $C$ the total number of categories, $\mathbbm{1}[y=c]$ an indicator function that is $1$ when $y=c$ and $0$ otherwise, $\sigma$ the softmax operation, and, in a slight overloading of notation, subscript $c$ represents the $c^{th}$ entry of the log-softmax vector. $\lambda$ is a hyper-parameter weighting the influence of the cross entropy and contrastive terms. Note also, as described above, Equation~\ref{eq:ce-contrastive} is calculated on a minibatch rather than over all datapoints simultaneously. 

To evaluate the monolithic approaches on compositions of corruptions we simply calculate classification accuracy on the domains in $\mathcal{E}_{all}$.

\subsection{A Modular Approach}
\vspace{-2mm}
The final approach we evaluate is a modular architecture, as it has been argued that modularity is a key feature of robust, intelligent systems \citep{goyal2022inductive, mahowald2023dissociating}. 
For each elemental corruption we add one module to our network which aims to `undo' the corruption in latent space. In practice these modules are intermediate layers that operate on hidden representations to map the representation of a corrupted image to the representation of the same image when uncorrupted. To make this possible modules are designed to have input and output features with the same shape. When classifying a test image corrupted with a composition of elemental corruptions we sequentially apply the modules for each corruption present in the composition. For example, if we are testing on the composition \emph{IN}$\circ$\emph{GB} we apply both the module trained on the \emph{Invert} corruption and the module trained on the \emph{Gaussian Blur} corruption. Modules that are located in-between earlier layers of the network are applied first, if modules are in the same layer we apply the module which appears first in the permutation ordering (Section~\ref{sec:setup}).

To formalize this idea, we split network parameters $\bm{\theta}$ into one set of parameters shared over all domains, $\bm{\theta}_{\textit{shared}}$, and an additional set of domain specific module parameters for each training domain $\{\bm{\theta}_e\}_{e \in \mathcal{E}_{tr}}$, similar to residual adaptation \citep{rebuffi17adapters, rebuffi2018efficient}. In practice $\bm{\theta}_{\textit{shared}}$ parameterizes a neural network and $\bm{\theta}_e$ the intermediate layers that can be inserted when working with domain $e$. 

To train this system we first train parameters $\bm{\theta}_{\textit{shared}}$ on \emph{Identity} data using the cross entropy loss. We then freeze $\bm{\theta}_{\textit{shared}}$ and train separate modules parameterized by $\bm{\theta}_e$ on data from each elemental training domain $e \in \mathcal{E}_{tr}$ along with paired \emph{Identity} data. Since we encourage the modules to `undo' corruptions, we use the loss function from Equation~\ref{eq:ce-contrastive} with minor modifications. Firstly, the set of domains for the contrastive loss %(Equation~\ref{eq:ce-contrastive}, second term)\
is limited to only the relevant elemental training domain and the \emph{Identity} domain. %(or equivalently, to train $\psi_e$, we define $\mathcal{E}_{tr}$ to be $\{e, \emph{Identity}\}$). 
Secondly, for the \emph{Identity} data, latent representation $\bm{z}$ is from the layer at which the module is inserted and for the corrupted data from the output of the module, spatially flattening the feature map if required (as opposed to from the penultimate network layer as described when introducing  Equation~\ref{eq:contrastive} in Section~\ref{sec:monolithic}). % Intuitively this encourages the module to `undo' the corruption in latent space.
Appendix~\ref{app:modules} contains a graphical depiction of this process.

% \textbf{Module Location}
An important design choice for any modular approach is how to choose where to locate the modules, with recent works observing that different domain changes should be dealt with in different neural network layers \citep{eastwood2022unitlevel,  royer2020flexible, lee2023surgical}. We take a very simple approach, training separate modules between each layer of the network parameterized by $\bm{\theta}_{\textit{shared}}$ for $5$ epochs. We then select the module with the best in-distribution accuracy on a held-out validation set as the module to train to completion. This is similar to using adaptation speed \citep{bengio2019meta, le2021analysis} as a proxy to discover modular decompositions, although in practice we find if we use adaptation times substantially smaller than $5$ epochs we can erroneously select module locations that do not achieve optimal in-distribution performance.

\subsection{Measuring the Invariance of Learned Representations}
\vspace{-2mm}
Since encouraging invariance is so prominent in the domain generalization literature \citep{zhou2022domain} we also empirically investigate the role of invariant representations in generalizing to unseen compositions of corruptions. We create two invariance scores following the methods of Madan et al. \citep{madan2022ood}, with full details along with an illustrative example in Appendix~\ref{app:invariance-scores}. These per-neuron scores are calculated for every neuron in the penultimate layer of the network (after applying modules if applicable), and the median score over all neurons is reported. Loosely the \emph{elemental invariance score}, is the maximum difference in neuron activation amongst the elemental corruptions normalized to lie between 0 and 1, with the idea that this score should be high when all elemental corruptions activate a neuron in a similar way (i.e.\ the neuron is invariant to the elemental corruptions). We additionally calculate the \emph{composition invariance score}, which also lies between $0$ and $1$ and measures how similarly a neuron activates on a composition when compared to the closest elemental corruption in the composition. We choose the closest elemental corruption because, to achieve high accuracy, it should be sufficient for a neuron to activate similarly on the composition and one elemental corruption, even if the elemental corruptions as a whole do not activate invariantly.

\subsection{Datasets, Architectures and Training Procedure}
\vspace{-2mm}
We evaluate each training approach on three different datasets for image classification: \emnist~\citep{cohen2017emnist}, an extended \mnist~with $47$ handwritten character classes; \cifar~\citep{krizhevsky2009cifar}, a simple object recognition dataset with $10$ classes, and \facescrub~\citep{ng2014facescrub}, a face-recognition dataset. For \facescrub~we follow \citep{vogelsang2018potential} removing classes with fewer than 100 images, resulting in $388$ classes, with each class representing an individual identity. We train using stochastic gradient descent with momentum $0.9$ and weight decay $5\times 10^{-4}$, learning rate is set using a grid search over $\{1, 10^{-1}, 10^{-2}, 10^{-3}\}$ and contrastive loss weighting, $\lambda$, over $\{10, 1, 10^{-1}, 10^{-2}\}$, with the best setting selected based on the performance on a validation set of the \emph{training} domains \citep{gulrajani2021search}. $\tau$ from Equation \ref{eq:contrastive} is set to $0.15$ in all experiments. We use a batch size of $256$ (or the nearest multiple of $|\mathcal{E}_{tr}|$ for the contrastive loss) and train for a maximum of $200$ epochs, using early stopping on the held out validation set. Each dataset is run over three seeds from which we select one seed to report the most pedagogical results. \cifar\ and \facescrub\ images are augmented with random cropping and flipping, ensuring positively paired examples receive exactly the same augmentation. For \emnist~we use a simple convolutional network with a LeNet-like \citep{lecun1998gradient} architecture with modules made up from convolutional layers. For \cifar~we use ResNet18 \citep{he2016resnet} without the first max pooling layer, wherever possible using ResNet blocks as modules. For \facescrub~we use Inception-v3 \citep{szegedy2016inception} without the auxiliary classifier. As with ResNet we use additional Inception-v3 layers as modules wherever possible. For full architectural details see Appendix~\ref{app:implementation}. 

\section{Results}
\label{sec:results}
\vspace{-2mm}

\begin{figure}[tb]
    \centering
    % \vspace{-6mm}
    \begin{subfigure}[t]{0.32\textwidth}
        \centering
        \includegraphics[width=\linewidth]{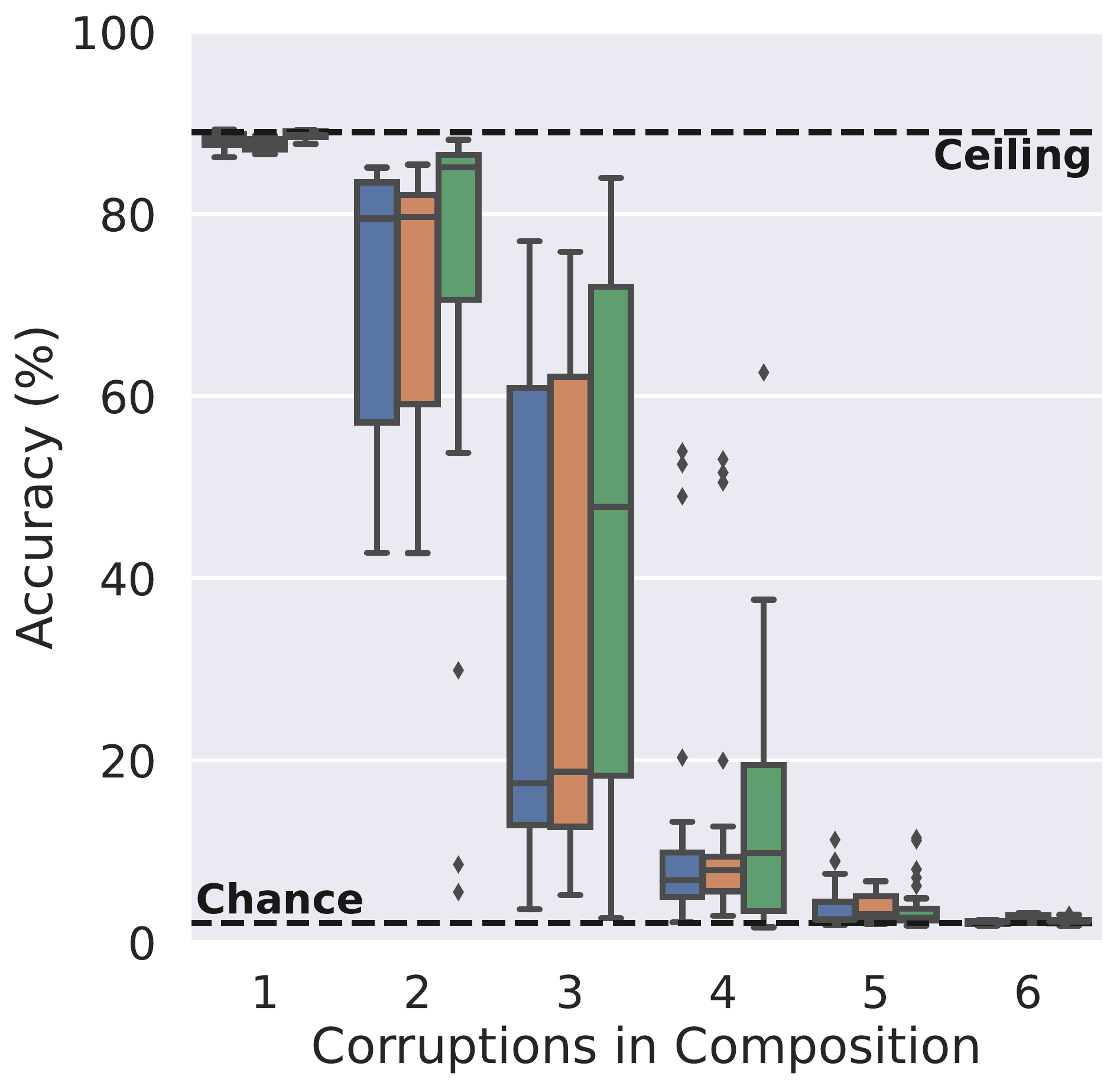} \\
        \caption{EMNIST}
        \label{fig:comp-em}
    \end{subfigure}%
    \hfill
    \begin{subfigure}[t]{0.32\textwidth}
        \centering
        \includegraphics[width=\linewidth]{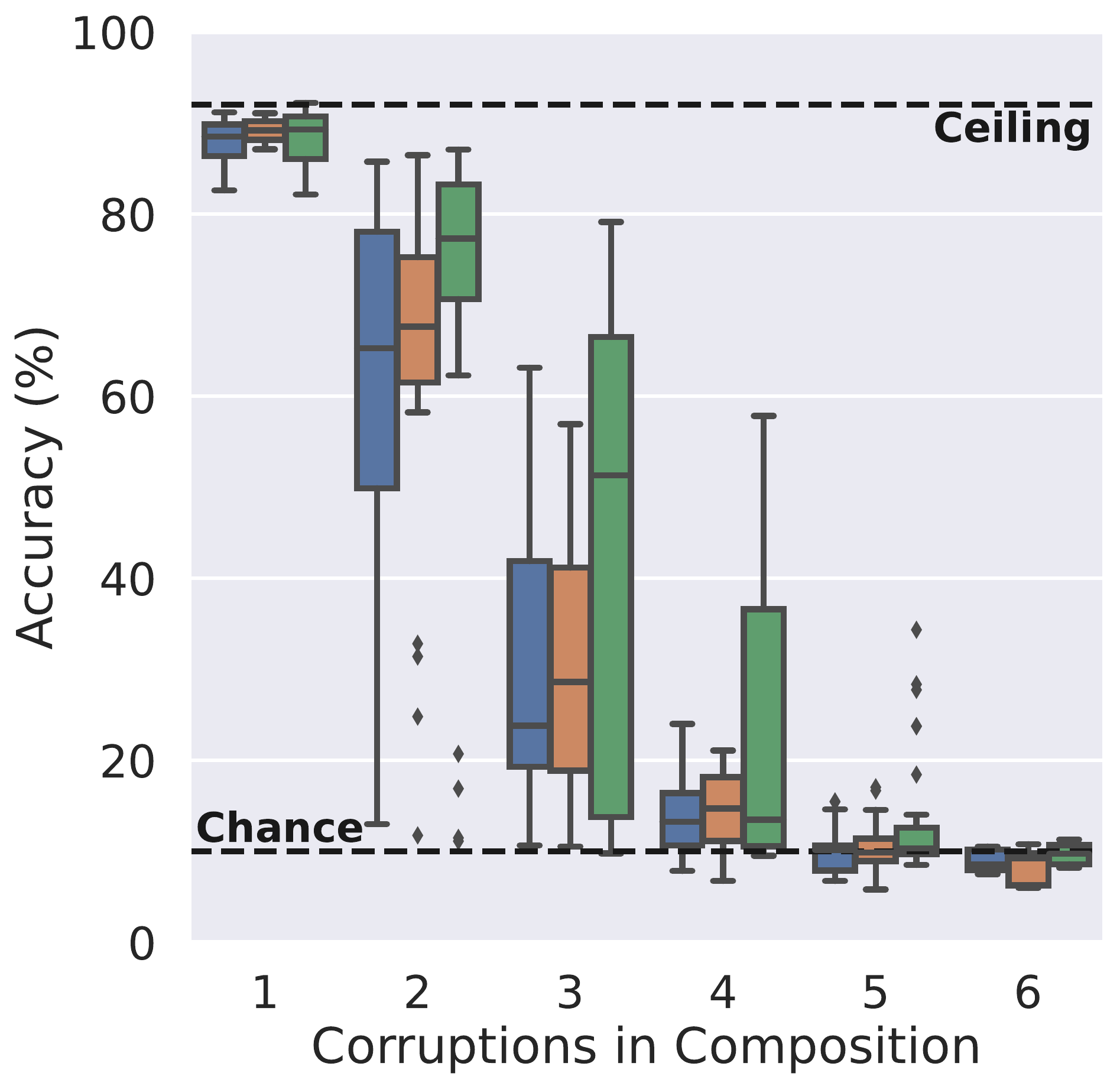} \\
        \caption{CIFAR-10}
        \label{fig:comp-cf}
    \end{subfigure}%
    \hfill
    \begin{subfigure}[t]{0.32\textwidth}
        \centering
        \includegraphics[width=\linewidth]{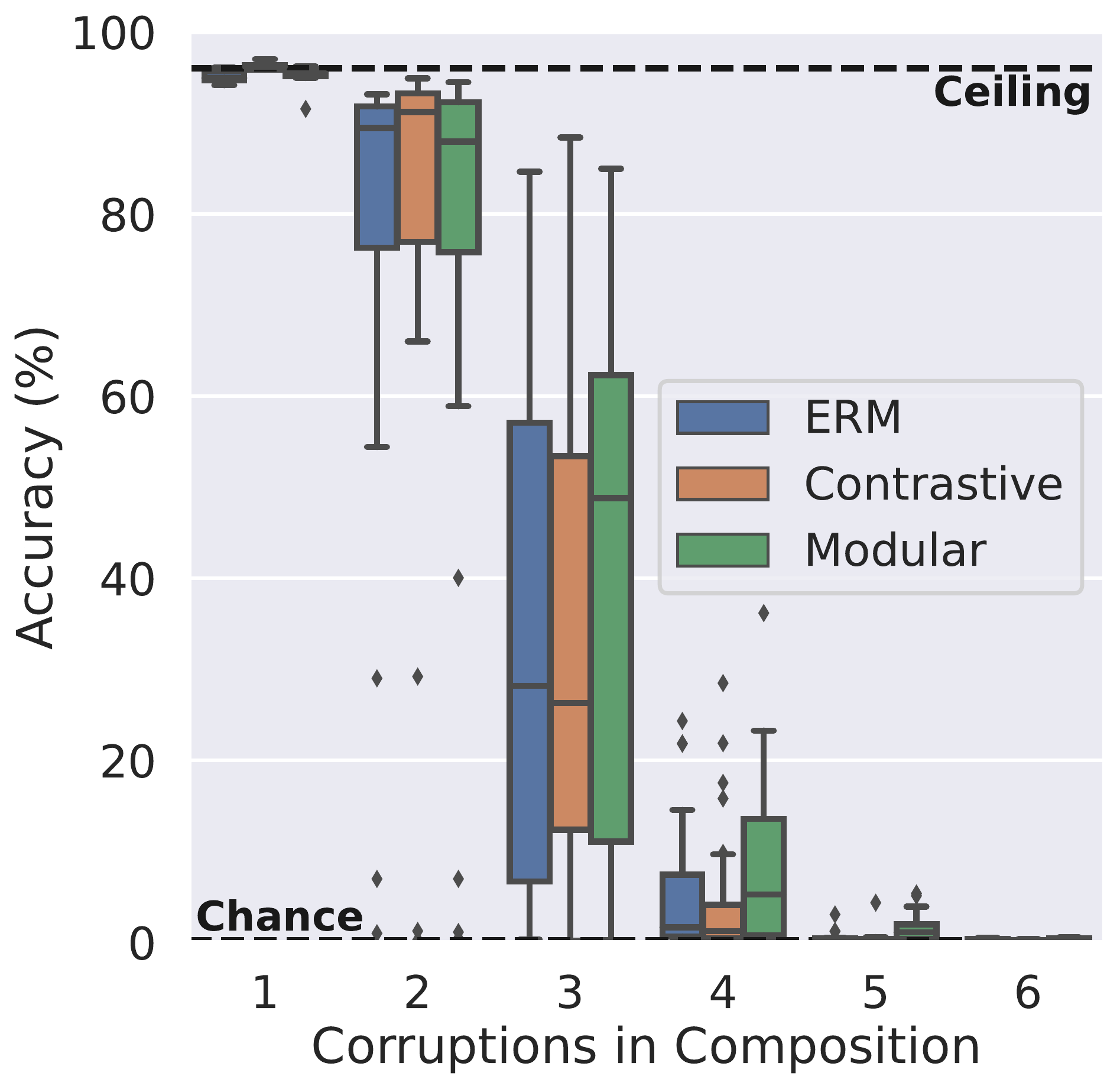} \\
        \caption{FACESCRUB}
        \label{fig:comp-fs}
    \end{subfigure}
    \vspace{-1.5mm}
    \caption{Evaluating compositional robustness on different datasets. Evaluation domains are divided into groups depending on the number of elemental corruptions making up a composition. Different colored boxes (left to right in each triple) show the performance of ERM, contrastive training, and the modular approach. Ceiling accuracy is determined by a model trained and tested on \emph{Identity} data.}
    \label{fig:comp}
    \vspace{-4.5mm}
\end{figure}

\looseness=-1 In this section we evaluate the compositional robustness of the different training approaches, first by examining the accuracy of different methods on unseen compositions of corruptions. We additionally explore the relationship between compositional robustness and invariance amongst representations of elemental corruptions. We end on the practical limitations of the approaches we consider in this study. 

\subsection{Monolithic Approaches Show Limited Compositional Robustness}
\vspace{-2mm}
Figure~\ref{fig:comp} shows the classification accuracies of each of the three approaches for each of the three datasets. The evaluation domains, $\mathcal{E}_{all}$, are divided into groups depending on how many elemental corruptions are in the composition applied to images in a domain. Across all methods and datasets we see domains with $1$ corruption achieve very good, near ceiling, performance. This is not surprising as this represents the accuracy on the elemental training domains. A granular view for each of the $167$ domains for every method can be seen in heat maps in Appendix~\ref{app:heat-maps}.

In Figure~\ref{fig:comp} the blue and orange box plots show the performance of ERM and contrastive training respectively, for which we can observe some general trends. Firstly, accuracy on compositions drops as the number of elemental corruptions in a composition increases, with compositions of $5$ or $6$ corruptions rarely performing above chance level. Intuitively, as each additional corruption makes the image harder to recognize (see Figure~\ref{fig:benchmark-facescrub}), it makes sense that this pattern emerges. Perhaps more surprisingly, both methods achieve accuracy far above chance for compositions of $2$ corruptions and perform relatively well for compositions of $3$ corruptions despite these domains being outside of the training distribution. We also see that the contrastive training approach makes only minor improvements over ERM, with the most improvement for \cifar. This runs counter to our assumption that encouraging invariance amongst training domains would increase compositional robustness. Finally, we note that neither method optimally solves the task,  some compositions of $2$ corruptions contain only invertible corruptions, yet neither method reaches ceiling performance for any composition of $2$ corruptions.

% On the other hand, we see that no method optimally solves the task, in our set-up both \emph{Invert} and \emph{Rotate90} are easily invertible corruptions, since we know all models achieve good accuracy for arbitrary training corruption $\textit{C}$ any composition of the form $\textit{C} \circ \textit{R90} \circ \textit{IN}$ or $\textit{C} \circ \textit{IN} \circ \textit{R90}$ should be solvable at ceiling level by inverting \emph{Invert} and \emph{Rotate90}, yet no method achieves ceiling performance for any domain containing a composition of $3$ or more corruptions.

% General trends (across all methods)
% - Acc goes down as number of elementals in composition increases. 5+ seems very hard.
% - Everyone does reasonably on compositions of 2. Not too bad on 3. Include expectation that OoD wouldn't work and that it is a bit surprising that even cross entropy can do well on compositions of 2.

\subsection{The Modular Approach Achieves the Best Compositional Robustness}
\vspace{-2mm}
Comparing all three training approaches, we observe that the modular approach outperforms both ERM and contrastive training, with higher mean performance in almost all cases in Figure~\ref{fig:comp}. The only exception is on compositions of $2$ corruptions for \facescrub, where the modular approach is marginally outperformed by contrastive training. These results demonstrate that  the monolithic approaches are unable to learn to modularize the structure of the task in the same way as the modular approach, since they do not achieve the same performance levels. Additionally we can observe that explicitly modularizing the modelling of elemental corruptions outperforms the direct encouragement of invariance in terms of compositional robustness. 
% Specific trends
% - Modular is best. Particularly compostions of three. A bit on 4.
%   - Only exception is FS compositions of 2. Despite most classes Facescrub seems easiest - something about faces, higher resolution?
% - Relatively limited improvement from Contrastive

\subsection{In-Distribution Invariance Does Not Correlate With Compositional Robustness}
\label{sec:invariance}
\vspace{-2mm}

\begin{figure}[tb]
    \centering
    \begin{subfigure}[t]{0.8\textwidth}
        \centering
        \includegraphics[trim={0 60cm 0 0},clip,width=\linewidth]{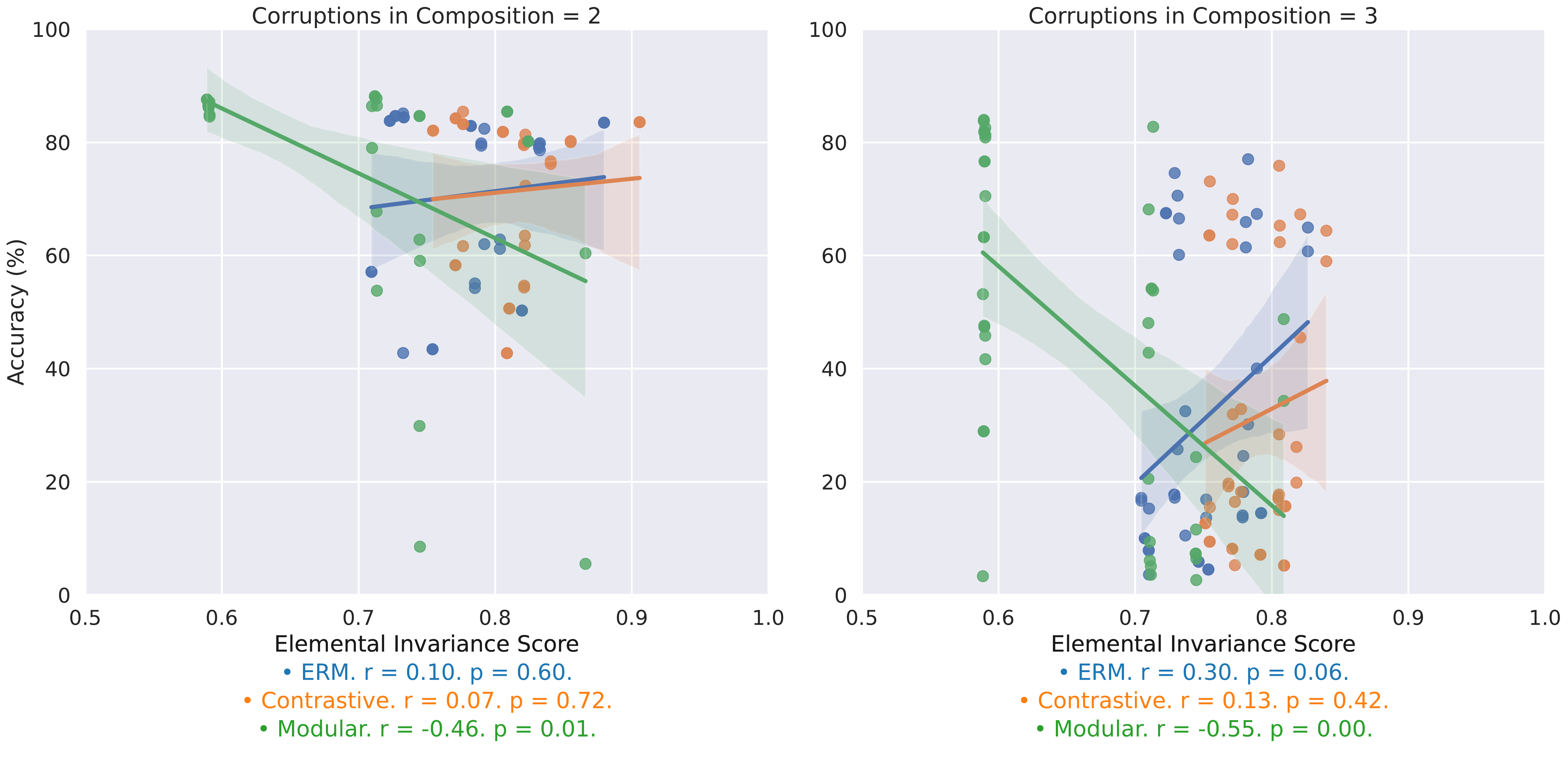} \\
    \end{subfigure}
    \begin{subfigure}[t]{0.8\textwidth}
        \centering
        \includegraphics[trim={0 60cm 0 0},clip,width=\linewidth]{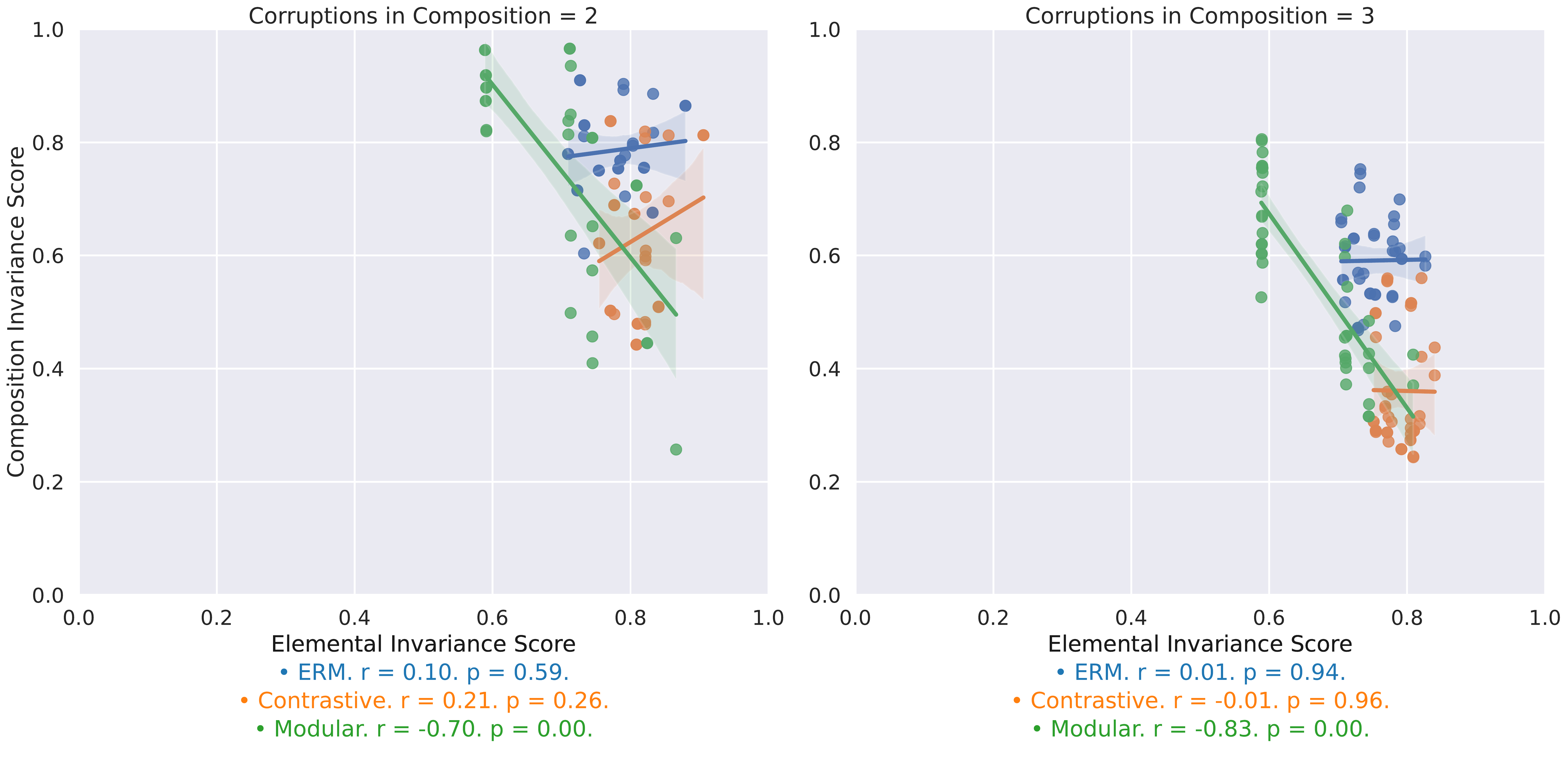} \\
    \end{subfigure}
    \begin{subfigure}[t]{0.8\textwidth}
        \centering
        \includegraphics[trim={0 60cm 0 0},clip,width=\linewidth]{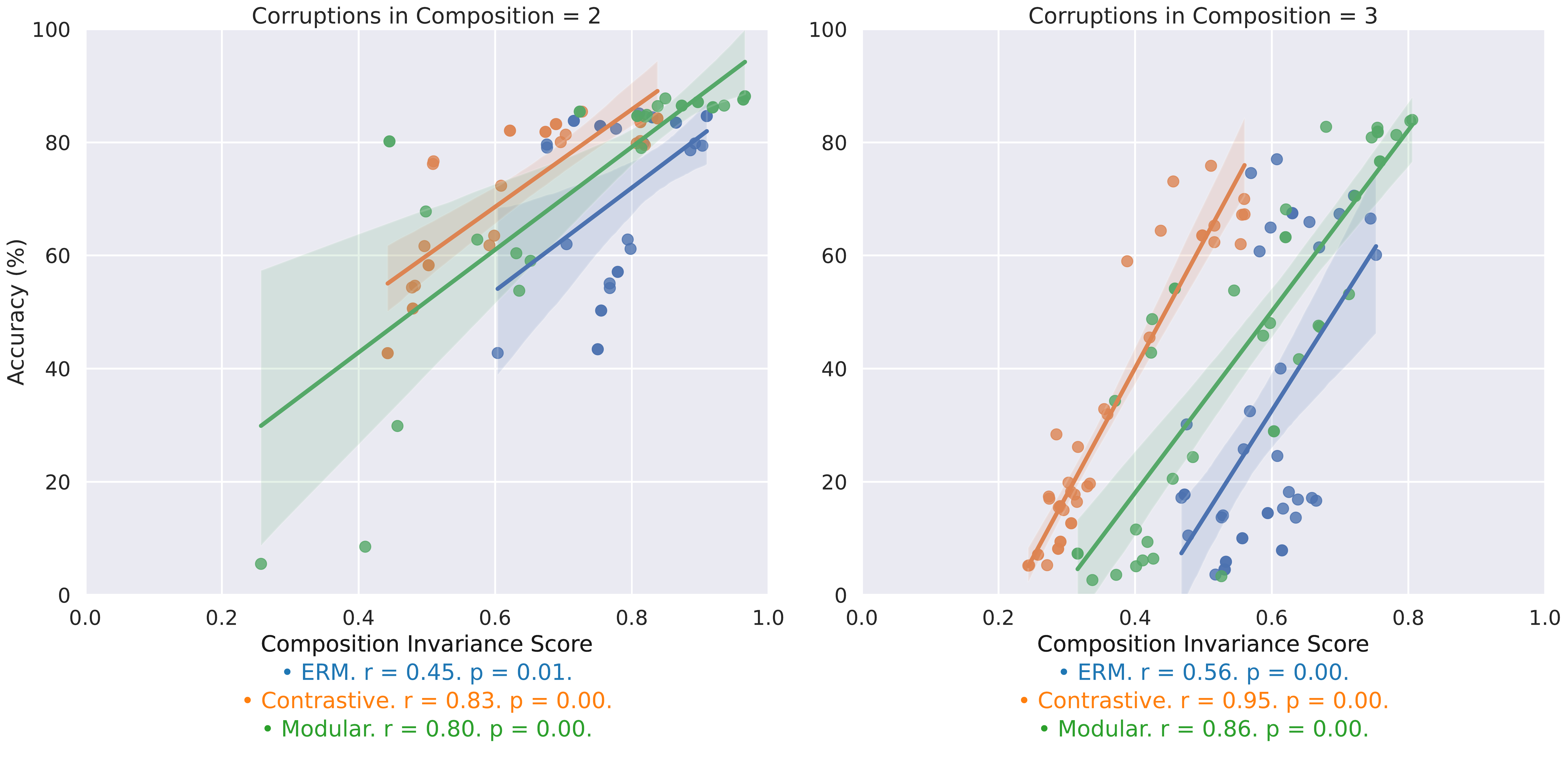} \\
    \end{subfigure}
    \vspace{-1.5mm}
    \caption{Correlating invariance scores with compositional robustness for \emnist. Row one shows the level of representational invariance amongst elemental corruptions fails to correlate with compositional robustness (accuracy). Row two shows the lack of dissemination of invariance between elemental corruptions and compositions. Row three plots composition invariance scores against compositional robustness. Columns show subsets of evaluation domains depending on the number of elemental corruptions making up a composition (as in Figure~\ref{fig:comp}).}
    \label{fig:invariance-emnist}
    \vspace{-6.5mm}
\end{figure}

To investigate our findings further we examine the invariance scores for the different approaches. We again split test domains by the number of elemental corruptions they include %\footnote{This avoids a natural correlation that arises. As we add more elemental corruptions the activations grid used to calculate the invariance scores adds additional rows. So, as the number of corruptions increases the invariance score  decreases as there is more chance to have corruptions be further away from each other.} 
and plot correlations for \emnist~in Figure~\ref{fig:invariance-emnist}. Figure~\ref{fig:invariance-emnist}, top row, plots the elemental invariance score against accuracy on compositions. Interestingly we observe no meaningful correlation between elemental invariance scores and accuracies on compositional test domains, with high p-values and low r-values. This runs counter to our initial expectations based on the ubiquity of invariant representation learning in the domain generalization literature. For our compositional task, these results indicate that encouraging invariance between representations on the training domains may be insufficient to achieve robustness. We even see some points for the modular approach (in the upper left of the plots) that achieve higher accuracy than ERM or contrastive training achieve on any domain yet have lower invariance scores. 

We also note that contrastive training only slightly increases the observed invariance between elemental corruptions, with a small rightward shift of points when compared to ERM. One possible reason for this smaller than expected increase may be because we set hyper-parameters on the training domains \citep{gulrajani2021search} and high contrastive weights take away from in-distribution performance. Alternatively, there has been some discussion on whether the contrastive loss improves performance because of increased invariance or by other mechanisms \cite{shen2022connect,sakai2022three}.

Row three of Figure~\ref{fig:invariance-emnist} shows strong positive correlations between the composition invariance score and accuracy on compositions. This is as expected, since a high composition invariance score indicates a similar representation between compositions and elemental corruptions (which all achieve good accuracy). However, in row two of Figure~\ref{fig:invariance-emnist} we again see limited, or even negative, correlations between elemental and composition invariance scores. This demonstrates that invariance built on elemental training domains may fail to transfer to invariance on compositional test domains, so we cannot consistently improve the composition invariance score by encouraging elemental invariance.

By and large these trends are consistent over datasets (Appendix~\ref{app:invariance}) and seeds (Appendix~\ref{app:seeds}). A notable exception is the negative correlation for the modular approach in row two of Figure~\ref{fig:invariance-emnist} is not seen in other datasets. We also observe a positive correlation between elemental invariance score and accuracy for ERM on \cifar. On \cifar, the encouraging of invariance with contrastive training builds slightly more invariant representations but then correlation between elemental invariance and accuracy disappears.

\subsection{Practical Limitations}
\label{sec:critical-analysis}
\vspace{-2mm}

\begin{figure}[tb]
    \centering
    % \vspace{-6mm}
    \begin{subfigure}[t]{0.32\textwidth}
        \centering
        \includegraphics[width=\linewidth]{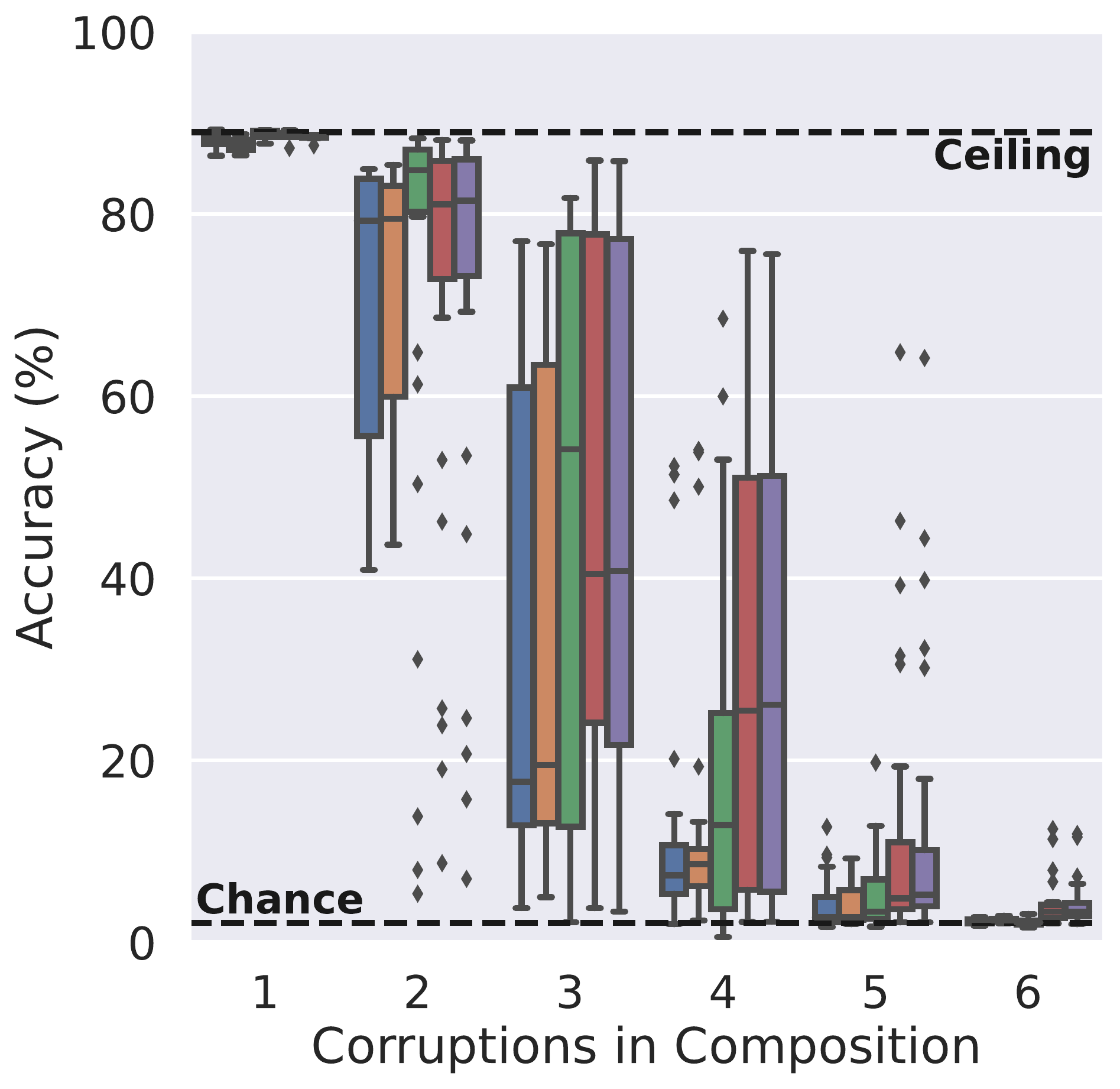} \\
        \caption{EMNIST}
        \label{fig:ablation-em}
    \end{subfigure}%
    \hfill
    \begin{subfigure}[t]{0.32\textwidth}
        \centering
        \includegraphics[width=\linewidth]{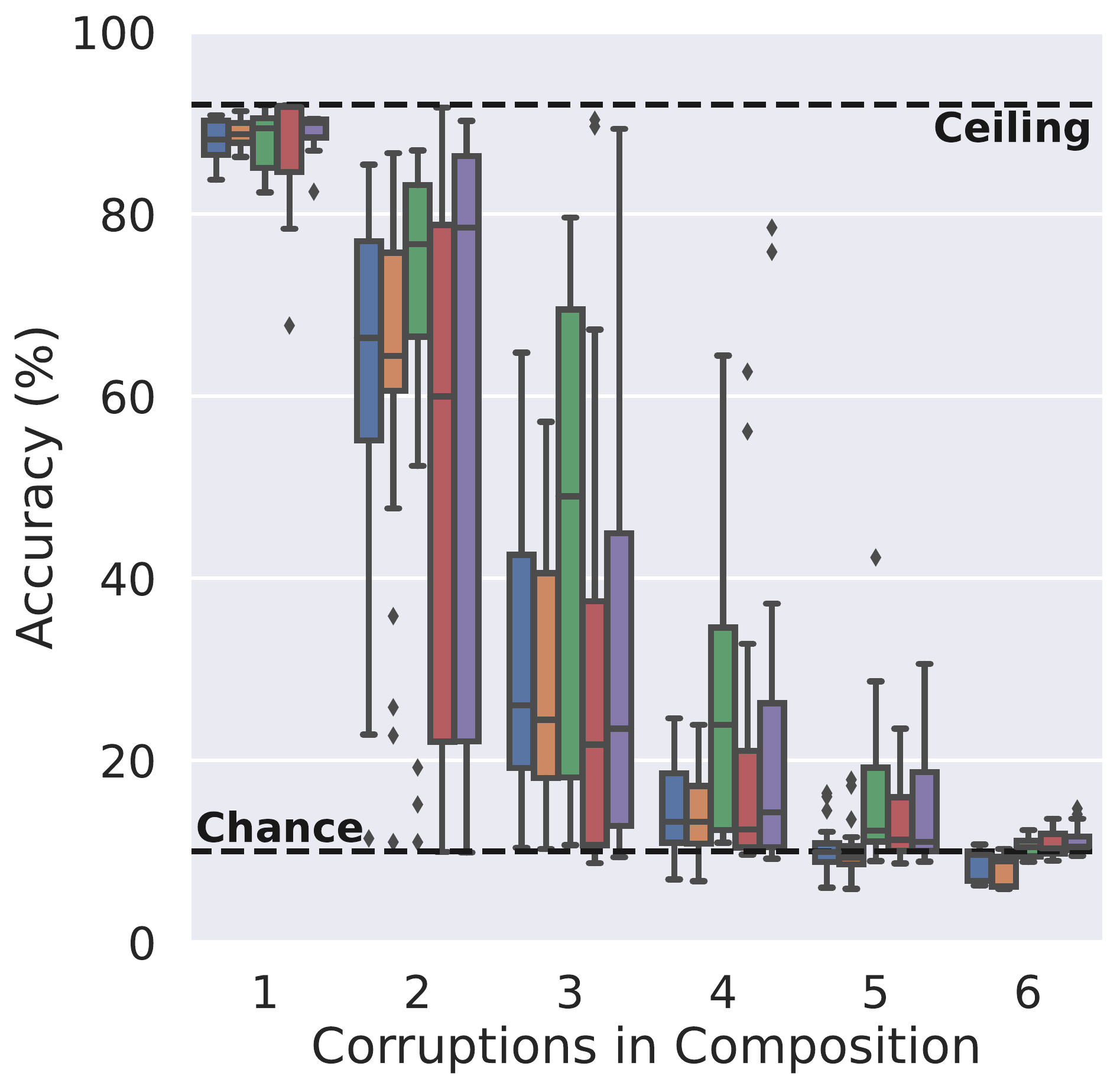} \\
        \caption{CIFAR-10}
        \label{fig:ablation-cf}
    \end{subfigure}%
    \hfill
    \begin{subfigure}[t]{0.32\textwidth}
        \centering
        \includegraphics[width=\linewidth]{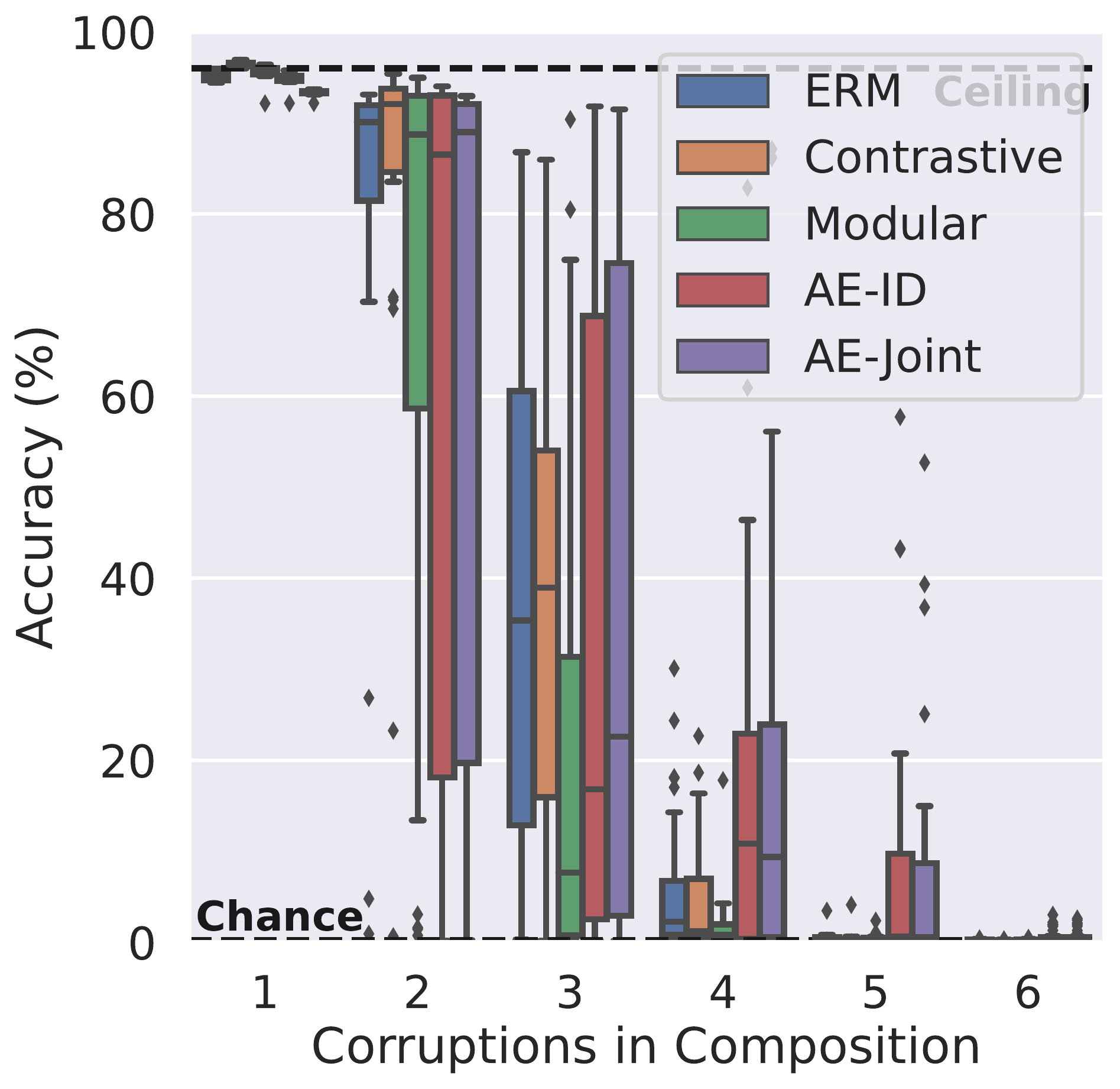} \\
        \caption{FACESCRUB}
        \label{fig:ablation-fs}
    \end{subfigure}
    \vspace{-1.5mm}
    \caption{Evaluating the compositional robustness of alternate training approaches and variance over seeds. The first three boxes in each quintuple are the same as in Figure~\ref{fig:comp} but trained with a different random seed. The remaining boxes (red and purple) show the performance of a modular approach using auto-encoders to undo each elemental corruption.}
    \label{fig:ablation}
    \vspace{-4.5mm}
\end{figure}

The aim of this work is to provide greater understanding of the factors that influence compositional robustness in neural networks. In particular, it is not our aim to provide an oven-ready method for improving compositional robustness. Nevertheless we now show some additional experiments to briefly highlight some of the practical limitations of the modular approach taken in this study.

Firstly, compared to the monolithic approaches, the modular approach has substantially higher variance over seeds. This is primarily due to variance in the selection of the module locations. Figure~\ref{fig:ablation} shows results for the same methods as in Figure~\ref{fig:comp} trained with a different random seed. Although better in some cases (\cifar, compositions of $5$), we see things can also be substantially worse (\facescrub). Whilst module location may have little effect on the in-distribution accuracy, putting modules in the optimum location had a large impact on compositional robustness and should be a focus of future work on modularity.

We also evaluate an alternative modular method where every corruption is handled in image space, that is, we train auto-encoders using mean squared error to transform a corrupted image into the corresponding \emph{Identity} image. To handle compositions we chain together auto-encoders for the relevant elemental corruptions, aiming to sequentially undo the corruptions to arrive at a clean image. We train two possible classifiers to use on the images outputted by this approach; the first minimizes cross entropy loss on clean data (AE-ID) and the second jointly on the outputs of the auto-encoders for all training corruptions (AE-Joint). The compositional robustness of these methods is shown in Figure~\ref{fig:ablation}. In general these methods perform relatively poorly on smaller numbers of corruptions (compositions of $2$ or $3$), this is largely because, as with the modular approach, the auto-encoders are sensitive to the ordering in which they are applied on a composition. On the other hand, the auto-encoders can often outperform all methods for larger numbers of corruptions, indicating that there likely exist methods that can achieve better compositional robustness than the methods we evaluate in this work.

Finally, apart from ERM all of the evaluated methods require paired data between domains which is an unrealistic expectation in practical applications. Additionally, for modular approaches we must know which corruptions are applied in a given test domain in order to apply the correct modules. Another interesting angle for future, more practically minded, solutions is to remove these assumptions.

% Left for Future Work:
% Automating, discussion of recurrence, level of abstraction. Even if invariance saturates for modules can still use other angles e.g. module capacity or location to try and improve things. Improving the issues in Critical Analysis section.

% What is the best that can be done? There are so many design choices here. Architectures of modules, contrastive loss weight, pass through or not for modules, number of modules, do the modules see all the corruptions or only the relevant ones, train on everything first then add modules, etc etc.

% We have provided a benchmark - can we do better?

\vspace{-2mm}
\section{Discussions}
\vspace{-2mm}
We end with several discussions on different interpretations of this work and links to larger questions that may motivate future work.

\textbf{What is the structure of natural data?}
In our compositional robustness framework we see only the elemental factors of variation (elemental corruptions) during training. In reality, whilst it is likely not possible to see every composition, most real-world data will contain an unstructured sampling of the compositional space. This assumes however, that it is possible to decompose data from the environment into elemental factors of variation \citep{chen2016infogan, higgins2017beta, kim2018disentangling, roth2023disentanglement} or independent (causal) mechanisms \citep{peters2017elements, scholkopf2021towards, parascandolo2018learning}. At present it remains unknown if there exists a practically sized set of elemental transformations from which all visual stimuli can be composed, but if such a set exists, the ideas presented in this work suggest that modular architectures may be able to model this space more efficiently than large monolithic models.
% I.e. for monolithic approaches you might just have to see loads and loads of data (including a lot of samples of the compositions). For modular approaches you may need fewer samples of compositions due to better compositional robustness.

\textbf{Learning to decompose from data.}
If there exists a set of elemental transformations from which all visual stimuli can be composed, and we are to make use of modularity as an inductive bias to model them, we must \emph{learn} how to decompose datasets into their constituent factors and how to modularize knowledge in the appropriate semantic spaces \citep{goyal2022inductive}. In this work we have shown that modular approaches have the potential to surpass previous approaches if the decomposition is available and progress has been made on finding appropriate semantic spaces \citep{royer2020flexible, eastwood2022unitlevel}. The learning of decompositions remains an open problem \citep{parascandolo2018learning, bengio2019meta, locatello2019challenging, goyal2021recurrent}.
%, it has been noted that is the general case factors of variation cannot be recovered from unsupervised data alone \citep{locatello2019challenging}.

% 2. Can we learn structure from data (or data and intervention).
    % Many extensions of the methods we laid out, overcoming the limitations discussed in ablation study.
    % Links to ICM and disentanglement
    % There are many different frameworks to work with (causality, modularity, disentanglement)
 
% Stuff like recurrent indep mechanisms. How far can we push modules. Can we learn modules? Do we need more? 
% Capturing causal factors from the environment etc.

\textbf{How modular should neural networks be?}
The modular approach taken in this study uses neural network layers as modules which are manually assigned to handle specific corruptions, yet we have also experimented with monolithic networks and with using entirely separate networks for each corruption (Section~\ref{sec:critical-analysis}). Even if we are able to decompose data into constituent factors, there remains a question of what degree of modularity should be used to model these factors. There have been recent exciting empirical studies in this direction \citep{damario2021modular, madan2022ood, yamada2022transformer} but no consensus has yet been reached.
% note: if modules are better than full networks suggests there is some benefit to modelling the 'shared factors'. That is, a shared model of the canonical distribution in the ICM framework.

\textbf{How far will invariance take us?}
Our results, and the results of others \citep{zhao2019learning, johansson2019support, rosenfeld2021risks, shen2022connect}, raise questions about whether encouraging invariance alone is sufficient to achieve domain generalization in general. We know that invariance is a key factor for robust generalization but we do not yet know how far invariance will be able to take us. Perhaps we simply need to better understand and implement the neural mechanisms that allow invariances to build \citep{anselmi2016unsupervised, poggio2016visual, scholkopf2021towards, goyal2022inductive}, or we may need to further explore learning representations that are only invariant over certain dimensions \citep{kong2022partial, shen2022connect, sun2023invariance}.
% 4. Limitations of invariance / invariance to what? Or an empirical or scientific study of compositionality in neural networks 
% Does Deephys talk about invariance to what? (i.e. invariance to shape or to color - depends on the task - don't want to throw away all the information. This I think is kind of the point of kong2022partial).

\vspace{-2mm}
\section{Conclusion}
\vspace{-2mm}
Since the visual space containing all corruptions is compositional in nature, we have introduced a new framework to evaluate the compositional robustness of different models. We have observed that modular approaches outperform monolithic approaches on this task, even when invariant representations are encouraged. For domain generalization tasks with compositional structure our results raise questions about the efficacy of encouraging invariance without further inductive biases. This work represents only a first step in understanding how neural networks behave under compositional structures, further research is needed into developing methods that make fewer assumptions about the information available at test time and that can work with large unstructured datasets where factors of variation are unknown. %It is also desirable that we develop a better theoretical understanding of compositionality and modularity.

\subsubsection*{Reproducibility statement} 
The code to reproduce the results herein is publicly available at the following GitHub repository: \url{https://github.com/ianxmason/compositional-robustness}. The experimental setup is described in Section~\ref{sec:methods}.

\subsubsection*{Acknowledgements} 
We would like to thank members of the Sinha Lab for Developmental Research and Fujitsu Research for helpful comments and feedback during the development of this project. In particular, Amir Rahimi, Hojin Jang, Avi Cooper, Ece {\"O}zkan Elsen, Hisanao Akima and Serban Georgescu. This work was supported by Fujitsu Limited (Contract No. 40009568).

\small
\bibliographystyle{unsrt}
\bibliography{compositions}

\begin{thebibliography}{100}

\bibitem{ito2022compositional}
Takuya Ito, Tim Klinger, Douglas~H Schultz, John~D Murray, Michael~W Cole, and
  Mattia Rigotti.
\newblock Compositional generalization through abstract representations in
  human and artificial neural networks.
\newblock In {\em Advances in Neural Information Processing Systems}, 2022.

\bibitem{lake2019human}
Brenden~M Lake, Tal Linzen, and Marco Baroni.
\newblock Human few-shot learning of compositional instructions.
\newblock In {\em Proceedings of the 41st Annual Conference of the Cognitive
  Science Society}, 2019.

\bibitem{piantadosi2016compositional}
Steven Piantadosi and Richard Aslin.
\newblock Compositional reasoning in early childhood.
\newblock {\em PLOS ONE}, 11(9):e0147734, 2016.

\bibitem{shulz2016probing}
Eric Schulz, Josh Tenenbaum, David~K Duvenaud, Maarten Speekenbrink, and
  Samuel~J Gershman.
\newblock Probing the compositionality of intuitive functions.
\newblock In {\em Advances in Neural Information Processing Systems}, 2016.

\bibitem{dodge2017study}
Samuel Dodge and Lina Karam.
\newblock A study and comparison of human and deep learning recognition
  performance under visual distortions.
\newblock In {\em 2017 26th International Conference on Computer Communication
  and Networks (ICCCN)}, pages 1--7. IEEE, 2017.

\bibitem{geirhos2021partial}
Robert Geirhos, Kantharaju Narayanappa, Benjamin Mitzkus, Tizian Thieringer,
  Matthias Bethge, Felix~A Wichmann, and Wieland Brendel.
\newblock Partial success in closing the gap between human and machine vision.
\newblock {\em Advances in Neural Information Processing Systems},
  34:23885--23899, 2021.

\bibitem{hosseini2017google}
Hossein Hosseini, Baicen Xiao, and Radha Poovendran.
\newblock Google's cloud vision api is not robust to noise.
\newblock In {\em 2017 16th IEEE International Conference on Machine Learning
  and Applications (ICMLA)}, pages 101--105. IEEE, 2017.

\bibitem{hendrycks2019benchmarking}
Dan Hendrycks and Thomas Dietterich.
\newblock Benchmarking neural network robustness to common corruptions and
  perturbations.
\newblock In {\em International Conference on Learning Representations}, 2019.

\bibitem{jang2021noise}
Hojin Jang, Devin McCormack, and Frank Tong.
\newblock Noise-trained deep neural networks effectively predict human vision
  and its neural responses to challenging images.
\newblock {\em PLOS Biology}, 19(12):e3001418, 2021.

\bibitem{hendrycks2020augmix}
Dan Hendrycks, Norman Mu, Ekin~D Cubuk, Barret Zoph, Justin Gilmer, and Balaji
  Lakshminarayanan.
\newblock Augmix: A simple data processing method to improve robustness and
  uncertainty.
\newblock In {\em International Conference on Learning Representations}, 2020.

\bibitem{hendrycks2021many}
Dan Hendrycks, Steven Basart, Norman Mu, Saurav Kadavath, Frank Wang, Evan
  Dorundo, Rahul Desai, Tyler Zhu, Samyak Parajuli, Mike Guo, Dawn Song, Jacob
  Steinhardt, and Justin Gilmer.
\newblock The many faces of robustness: A critical analysis of
  out-of-distribution generalization.
\newblock In {\em Proceedings of the IEEE/CVF International Conference on
  Computer Vision (ICCV)}, pages 8340--8349, October 2021.

\bibitem{yun2019cutmix}
Sangdoo Yun, Dongyoon Han, Seong~Joon Oh, Sanghyuk Chun, Junsuk Choe, and
  Youngjoon Yoo.
\newblock Cutmix: Regularization strategy to train strong classifiers with
  localizable features.
\newblock In {\em Proceedings of the IEEE/CVF International Conference on
  Computer Vision}, pages 6023--6032, 2019.

\bibitem{zhang2018mixup}
Hongyi Zhang, Moustapha Cisse, Yann~N Dauphin, and David Lopez-Paz.
\newblock mixup: Beyond empirical risk minimization.
\newblock In {\em International Conference on Learning Representations}, 2018.

\bibitem{huang2022robustness}
Tianjian Huang, Shaunak~Ashish Halbe, Chinnadhurai Sankar, Pooyan Amini, Satwik
  Kottur, Alborz Geramifard, Meisam Razaviyayn, and Ahmad Beirami.
\newblock Robustness through data augmentation loss consistency.
\newblock {\em Transactions on Machine Learning Research}, 2022.

\bibitem{kim2019learning}
Byungju Kim, Hyunwoo Kim, Kyungsu Kim, Sungjin Kim, and Junmo Kim.
\newblock Learning not to learn: Training deep neural networks with biased
  data.
\newblock In {\em Proceedings of the IEEE/CVF Conference on Computer Vision and
  Pattern Recognition}, pages 9012--9020, 2019.

\bibitem{sinha2021consistency}
Samarth Sinha and Adji~Bousso Dieng.
\newblock Consistency regularization for variational auto-encoders.
\newblock {\em Advances in Neural Information Processing Systems},
  34:12943--12954, 2021.

\bibitem{von2021self}
Julius Von~K{\"u}gelgen, Yash Sharma, Luigi Gresele, Wieland Brendel, Bernhard
  Sch{\"o}lkopf, Michel Besserve, and Francesco Locatello.
\newblock Self-supervised learning with data augmentations provably isolates
  content from style.
\newblock {\em Advances in Neural Information Processing Systems},
  34:16451--16467, 2021.

\bibitem{bahdanau2019systematic}
Dzmitry Bahdanau, Shikhar Murty, Michael Noukhovitch, Thien~Huu Nguyen, Harm
  de~Vries, and Aaron Courville.
\newblock Systematic generalization: What is required and can it be learned?
\newblock In {\em International Conference on Learning Representations}, 2019.

\bibitem{chalmers1990fodor}
David Chalmers.
\newblock Why {Fodor} and {Pylyshyn} were wrong: The simplest refutation.
\newblock In {\em Proceedings of the twelfth Annual Conference of the Cognitive
  Science Society}, pages 340--347, 1990.

\bibitem{fodor1988connectionism}
Jerry~A Fodor and Zenon~W Pylyshyn.
\newblock Connectionism and cognitive architecture: A critical analysis.
\newblock {\em Cognition}, 28(1-2):3--71, 1988.

\bibitem{goyal2022inductive}
Anirudh Goyal and Yoshua Bengio.
\newblock Inductive biases for deep learning of higher-level cognition.
\newblock {\em Proceedings of the Royal Society A}, 478(2266):20210068, 2022.

\bibitem{lake2018generalization}
Brenden Lake and Marco Baroni.
\newblock Generalization without systematicity: On the compositional skills of
  sequence-to-sequence recurrent networks.
\newblock In {\em International Conference on Machine Learning}, pages
  2873--2882. PMLR, 2018.

\bibitem{lake2017building}
Brenden~M Lake, Tomer~D Ullman, Joshua~B Tenenbaum, and Samuel~J Gershman.
\newblock Building machines that learn and think like people.
\newblock {\em Behavioral and Brain Sciences}, 40, 2017.

\bibitem{mendez2022review}
Jorge~A Mendez and Eric Eaton.
\newblock How to reuse and compose knowledge for a lifetime of tasks: A survey
  on continual learning and functional composition.
\newblock {\em arXiv preprint arXiv:2207.07730}, 2022.

\bibitem{peters2017elements}
Jonas Peters, Dominik Janzing, and Bernhard Sch{\"o}lkopf.
\newblock {\em Elements of causal inference: foundations and learning
  algorithms}.
\newblock The MIT Press, 2017.

\bibitem{parascandolo2018learning}
Giambattista Parascandolo, Niki Kilbertus, Mateo Rojas-Carulla, and Bernhard
  Sch{\"o}lkopf.
\newblock Learning independent causal mechanisms.
\newblock In {\em International Conference on Machine Learning}, pages
  4036--4044. PMLR, 2018.

\bibitem{gulrajani2021search}
Ishaan Gulrajani and David Lopez-Paz.
\newblock In search of lost domain generalization.
\newblock In {\em International Conference on Learning Representations}, 2022.

\bibitem{chen2020simclr}
Ting Chen, Simon Kornblith, Mohammad Norouzi, and Geoffrey Hinton.
\newblock A simple framework for contrastive learning of visual
  representations.
\newblock In {\em International Conference on Machine Learning}, pages
  1597--1607. PMLR, 2020.

\bibitem{gutmann2010nce}
Michael Gutmann and Aapo Hyvärinen.
\newblock Noise-contrastive estimation: A new estimation principle for
  unnormalized statistical models.
\newblock In {\em Proceedings of the Thirteenth International Conference on
  Artificial Intelligence and Statistics}, volume~9, pages 297--304. PMLR,
  2010.

\bibitem{hadsell2006dimensionality}
Raia Hadsell, Sumit Chopra, and Yann LeCun.
\newblock Dimensionality reduction by learning an invariant mapping.
\newblock In {\em 2006 IEEE Computer Society Conference on Computer Vision and
  Pattern Recognition (CVPR'06)}, volume~2, pages 1735--1742. IEEE, 2006.

\bibitem{ahmed2021systematic}
Faruk Ahmed, Yoshua Bengio, Harm Van~Seijen, and Aaron Courville.
\newblock Systematic generalisation with group invariant predictions.
\newblock In {\em International Conference on Learning Representations}, 2021.

\bibitem{albuquerque2019generalizing}
Isabela Albuquerque, Jo{\~a}o Monteiro, Mohammad Darvishi, Tiago~H Falk, and
  Ioannis Mitliagkas.
\newblock Generalizing to unseen domains via distribution matching.
\newblock {\em arXiv preprint arXiv:1911.00804}, 2019.

\bibitem{arjovsky2019irm}
Martin Arjovsky, L{\'e}on Bottou, Ishaan Gulrajani, and David Lopez-Paz.
\newblock Invariant risk minimization.
\newblock {\em arXiv preprint arXiv:1907.02893}, 2019.

\bibitem{dou2019domain}
Qi~Dou, Daniel Coelho~de Castro, Konstantinos Kamnitsas, and Ben Glocker.
\newblock Domain generalization via model-agnostic learning of semantic
  features.
\newblock {\em Advances in Neural Information Processing Systems}, 32, 2019.

\bibitem{li2018deep}
Ya~Li, Xinmei Tian, Mingming Gong, Yajing Liu, Tongliang Liu, Kun Zhang, and
  Dacheng Tao.
\newblock Deep domain generalization via conditional invariant adversarial
  networks.
\newblock In {\em Proceedings of the European Conference on Computer Vision
  (ECCV)}, pages 624--639, 2018.

\bibitem{ghifary2015domain}
Muhammad Ghifary, W~Bastiaan Kleijn, Mengjie Zhang, and David Balduzzi.
\newblock Domain generalization for object recognition with multi-task
  autoencoders.
\newblock In {\em Proceedings of the IEEE International Conference on Computer
  Vision}, pages 2551--2559, 2015.

\bibitem{kim2021selfreg}
Daehee Kim, Youngjun Yoo, Seunghyun Park, Jinkyu Kim, and Jaekoo Lee.
\newblock Selfreg: Self-supervised contrastive regularization for domain
  generalization.
\newblock In {\em Proceedings of the IEEE/CVF International Conference on
  Computer Vision}, pages 9619--9628, 2021.

\bibitem{li2018domain}
Haoliang Li, Sinno~Jialin Pan, Shiqi Wang, and Alex~C Kot.
\newblock Domain generalization with adversarial feature learning.
\newblock In {\em Proceedings of the IEEE Conference on Computer Vision and
  Pattern Recognition}, pages 5400--5409, 2018.

\bibitem{motiian2017unified}
Saeid Motiian, Marco Piccirilli, Donald~A Adjeroh, and Gianfranco Doretto.
\newblock Unified deep supervised domain adaptation and generalization.
\newblock In {\em Proceedings of the IEEE International Conference on Computer
  Vision}, pages 5715--5725, 2017.

\bibitem{sakai2022three}
Akira Sakai, Taro Sunagawa, Spandan Madan, Kanata Suzuki, Takashi Katoh,
  Hiromichi Kobashi, Hanspeter Pfister, Pawan Sinha, Xavier Boix, and Tomotake
  Sasaki.
\newblock Three approaches to facilitate invariant neurons and generalization
  to out-of-distribution orientations and illuminations.
\newblock {\em Neural Networks}, 155:119--143, 2022.

\bibitem{creager2021environment}
Elliot Creager, J{\"o}rn-Henrik Jacobsen, and Richard Zemel.
\newblock Environment inference for invariant learning.
\newblock In {\em International Conference on Machine Learning}, pages
  2189--2200. PMLR, 2021.

\bibitem{pfeiffer2023modular}
Jonas Pfeiffer, Sebastian Ruder, Ivan Vuli{\'c}, and Edoardo~Maria Ponti.
\newblock Modular deep learning.
\newblock {\em arXiv preprint arXiv:2302.11529}, 2023.

\bibitem{krueger2021out}
David Krueger, Ethan Caballero, Joern-Henrik Jacobsen, Amy Zhang, Jonathan
  Binas, Dinghuai Zhang, Remi Le~Priol, and Aaron Courville.
\newblock Out-of-distribution generalization via risk extrapolation (rex).
\newblock In {\em International Conference on Machine Learning}, pages
  5815--5826. PMLR, 2021.

\bibitem{eastwood2022probable}
Cian Eastwood, Alexander Robey, Shashank Singh, Julius von K\"{u}gelgen, Hamed
  Hassani, George~J. Pappas, and Bernhard Sch\"{o}lkopf.
\newblock Probable domain generalization via quantile risk minimization.
\newblock In {\em Advances in Neural Information Processing Systems},
  volume~35, pages 17340--17358, 2022.

\bibitem{ganin2016dann}
Yaroslav Ganin, Evgeniya Ustinova, Hana Ajakan, Pascal Germain, Hugo
  Larochelle, Fran{\c{c}}ois Laviolette, Mario Marchand, and Victor Lempitsky.
\newblock Domain-adversarial training of neural networks.
\newblock {\em Journal of Machine Learning Research}, 17(1):2096--2030, 2016.

\bibitem{tzeng2017adversarial}
Eric Tzeng, Judy Hoffman, Kate Saenko, and Trevor Darrell.
\newblock Adversarial discriminative domain adaptation.
\newblock In {\em Proceedings of the IEEE Conference on Computer Vision and
  Pattern Recognition}, pages 7167--7176, 2017.

\bibitem{sun2016deepcoral}
Baochen Sun and Kate Saenko.
\newblock Deep {CORAL}: Correlation alignment for deep domain adaptation.
\newblock In {\em European Conference on Computer Vision}, pages 443--450.
  Springer, 2016.

\bibitem{long2015learning}
Mingsheng Long, Yue Cao, Jianmin Wang, and Michael Jordan.
\newblock Learning transferable features with deep adaptation networks.
\newblock In {\em International Conference on Machine Learning}, pages 97--105,
  2015.

\bibitem{long2018cdan}
Mingsheng Long, Zhangjie Cao, Jianmin Wang, and Michael~I Jordan.
\newblock Conditional adversarial domain adaptation.
\newblock In {\em Advances in Neural Information Processing Systems}, 2018.

\bibitem{eastwood2022sourcefree}
Cian Eastwood, Ian Mason, Christopher K.~I. Williams, and Bernhard Sch\"olkopf.
\newblock Source-free adaptation to measurement shift via bottom-up feature
  restoration.
\newblock In {\em International Conference on Learning Representations}, 2022.

\bibitem{ben2007domain}
Shai Ben-David, John Blitzer, Koby Crammer, and Fernando Pereira.
\newblock Analysis of representations for domain adaptation.
\newblock In {\em Advances in Neural Information Processing Systems}, pages
  137--144, 2007.

\bibitem{ben2010theory}
Shai Ben-David, John Blitzer, Koby Crammer, Alex Kulesza, Fernando Pereira, and
  Jennifer~Wortman Vaughan.
\newblock A theory of learning from different domains.
\newblock {\em Machine Learning}, 79(1):151--175, 2010.

\bibitem{zhao2019learning}
Han Zhao, Remi~Tachet Des~Combes, Kun Zhang, and Geoffrey Gordon.
\newblock On learning invariant representations for domain adaptation.
\newblock In {\em International conference on machine learning}, pages
  7523--7532. PMLR, 2019.

\bibitem{johansson2019support}
Fredrik~D Johansson, David Sontag, and Rajesh Ranganath.
\newblock Support and invertibility in domain-invariant representations.
\newblock In {\em The 22nd International Conference on Artificial Intelligence
  and Statistics}, pages 527--536. PMLR, 2019.

\bibitem{rosenfeld2021risks}
Elan Rosenfeld, Pradeep Ravikumar, and Andrej Risteski.
\newblock The risks of invariant risk minimization.
\newblock In {\em International Conference on Learning Representations}, 2021.

\bibitem{shen2022connect}
Kendrick Shen, Robbie~M Jones, Ananya Kumar, Sang~Michael Xie, Jeff~Z HaoChen,
  Tengyu Ma, and Percy Liang.
\newblock Connect, not collapse: Explaining contrastive learning for
  unsupervised domain adaptation.
\newblock In {\em International Conference on Machine Learning}, pages
  19847--19878. PMLR, 2022.

\bibitem{akuzawa2020adversarial}
Kei Akuzawa, Yusuke Iwasawa, and Yutaka Matsuo.
\newblock Adversarial invariant feature learning with accuracy constraint for
  domain generalization.
\newblock In {\em Machine Learning and Knowledge Discovery in Databases:
  European Conference, ECML PKDD 2019, W{\"u}rzburg, Germany, September 16--20,
  2019, Proceedings, Part II}, pages 315--331. Springer, 2020.

\bibitem{battaglia2018relational}
Peter~W Battaglia, Jessica~B Hamrick, Victor Bapst, Alvaro Sanchez-Gonzalez,
  Vinicius Zambaldi, Mateusz Malinowski, Andrea Tacchetti, David Raposo, Adam
  Santoro, Ryan Faulkner, et~al.
\newblock Relational inductive biases, deep learning, and graph networks.
\newblock {\em arXiv preprint arXiv:1806.01261}, 2018.

\bibitem{cohen2016group}
Taco Cohen and Max Welling.
\newblock Group equivariant convolutional networks.
\newblock In {\em International conference on machine learning}, pages
  2990--2999. PMLR, 2016.

\bibitem{weiler2018learning}
Maurice Weiler, Fred~A Hamprecht, and Martin Storath.
\newblock Learning steerable filters for rotation equivariant cnns.
\newblock In {\em Proceedings of the IEEE Conference on Computer Vision and
  Pattern Recognition}, pages 849--858, 2018.

\bibitem{worrall2017harmonic}
Daniel~E Worrall, Stephan~J Garbin, Daniyar Turmukhambetov, and Gabriel~J
  Brostow.
\newblock Harmonic networks: Deep translation and rotation equivariance.
\newblock In {\em Proceedings of the IEEE Conference on Computer Vision and
  Pattern Recognition}, pages 5028--5037, 2017.

\bibitem{cohen2019general}
Taco~S Cohen, Mario Geiger, and Maurice Weiler.
\newblock A general theory of equivariant cnns on homogeneous spaces.
\newblock {\em Advances in neural information processing systems}, 32, 2019.

\bibitem{scholkopf2021towards}
Bernhard Sch{\"o}lkopf, Francesco Locatello, Stefan Bauer, Nan~Rosemary Ke, Nal
  Kalchbrenner, Anirudh Goyal, and Yoshua Bengio.
\newblock Towards causal representation learning.
\newblock {\em Proceedings of the IEEE}, 109(5):612--634, 2021.

\bibitem{goyal2021recurrent}
Anirudh Goyal, Alex Lamb, Jordan Hoffmann, Shagun Sodhani, Sergey Levine,
  Yoshua Bengio, and Bernhard Sch{\"o}lkopf.
\newblock Recurrent independent mechanisms.
\newblock In {\em International Conference on Learning Representations}, 2021.

\bibitem{chen2016infogan}
Xi~Chen, Yan Duan, Rein Houthooft, John Schulman, Ilya Sutskever, and Pieter
  Abbeel.
\newblock Infogan: Interpretable representation learning by information
  maximizing generative adversarial nets.
\newblock {\em Advances in neural information processing systems}, 29, 2016.

\bibitem{higgins2017beta}
Irina Higgins, Loic Matthey, Arka Pal, Christopher Burgess, Xavier Glorot,
  Matthew Botvinick, Shakir Mohamed, and Alexander Lerchner.
\newblock beta-vae: Learning basic visual concepts with a constrained
  variational framework.
\newblock In {\em International conference on learning representations}, 2017.

\bibitem{kim2018disentangling}
Hyunjik Kim and Andriy Mnih.
\newblock Disentangling by factorising.
\newblock In {\em International Conference on Machine Learning}, pages
  2649--2658. PMLR, 2018.

\bibitem{roth2023disentanglement}
Karsten Roth, Mark Ibrahim, Zeynep Akata, Pascal Vincent, and Diane
  Bouchacourt.
\newblock Disentanglement of correlated factors via hausdorff factorized
  support.
\newblock In {\em International Conference on Learning Representations}, 2023.

\bibitem{eastwood2018framework}
Cian Eastwood and Christopher~KI Williams.
\newblock A framework for the quantitative evaluation of disentangled
  representations.
\newblock In {\em International Conference on Learning Representations}, 2018.

\bibitem{locatello2019challenging}
Francesco Locatello, Stefan Bauer, Mario Lucic, Gunnar Raetsch, Sylvain Gelly,
  Bernhard Sch{\"o}lkopf, and Olivier Bachem.
\newblock Challenging common assumptions in the unsupervised learning of
  disentangled representations.
\newblock In {\em International Conference on Machine Learning}, pages
  4114--4124. PMLR, 2019.

\bibitem{schott2021visual}
Lukas Schott, Julius Von~K{\"u}gelgen, Frederik Tr{\"a}uble, Peter Gehler,
  Chris Russell, Matthias Bethge, Bernhard Sch{\"o}lkopf, Francesco Locatello,
  and Wieland Brendel.
\newblock Visual representation learning does not generalize strongly within
  the same domain.
\newblock In {\em International Conference on Learning Representations}, 2022.

\bibitem{montero2021role}
Milton~Llera Montero, Casimir~JH Ludwig, Rui~Ponte Costa, Gaurav Malhotra, and
  Jeffrey Bowers.
\newblock The role of disentanglement in generalisation.
\newblock In {\em International Conference on Learning Representations}, 2021.

\bibitem{montero2022lost}
Milton Montero, Jeffrey Bowers, Rui Ponte~Costa, Casimir Ludwig, and Gaurav
  Malhotra.
\newblock Lost in latent space: Examining failures of disentangled models at
  combinatorial generalisation.
\newblock In {\em Advances in Neural Information Processing Systems},
  volume~35, pages 10136--10149, 2022.

\bibitem{jaderberg2015spatial}
Max Jaderberg, Karen Simonyan, Andrew Zisserman, et~al.
\newblock Spatial transformer networks.
\newblock {\em Advances in neural information processing systems}, 28, 2015.

\bibitem{andreas2016learning}
Jacob Andreas, Marcus Rohrbach, Trevor Darrell, and Dan Klein.
\newblock Learning to compose neural networks for question answering.
\newblock In {\em Proceedings of the 2016 Conference of the North {A}merican
  Chapter of the Association for Computational Linguistics: Human Language
  Technologies}, 2016.

\bibitem{andreas2016neural}
Jacob Andreas, Marcus Rohrbach, Trevor Darrell, and Dan Klein.
\newblock Neural module networks.
\newblock In {\em Proceedings of the IEEE conference on computer vision and
  pattern recognition}, pages 39--48, 2016.

\bibitem{damario2021modular}
Vanessa D'Amario, Tomotake Sasaki, and Xavier Boix.
\newblock How modular should neural module networks be for systematic
  generalization?
\newblock {\em Advances in Neural Information Processing Systems},
  34:23374--23385, 2021.

\bibitem{goyal2022coordination}
Anirudh Goyal, Aniket Didolkar, Alex Lamb, Kartikeya Badola, Nan~Rosemary Ke,
  Nasim Rahaman, Jonathan Binas, Charles Blundell, Michael Mozer, and Yoshua
  Bengio.
\newblock Coordination among neural modules through a shared global workspace.
\newblock In {\em International Conference on Learning Representations}, 2022.

\bibitem{mendez2021lifelong}
Jorge~A Mendez and Eric Eaton.
\newblock Lifelong learning of compositional structures.
\newblock In {\em International Conference on Learning Representations}, 2021.

\bibitem{madan2022ood}
Spandan Madan, Timothy Henry, Jamell Dozier, Helen Ho, Nishchal Bhandari,
  Tomotake Sasaki, Fr{\'e}do Durand, Hanspeter Pfister, and Xavier Boix.
\newblock When and how convolutional neural networks generalize to
  out-of-distribution category--viewpoint combinations.
\newblock {\em Nature Machine Intelligence}, 4(2):146--153, 2022.

\bibitem{carvalho2023composing}
Wilka Carvalho, Angelos Filos, Richard~L Lewis, Satinder Singh, et~al.
\newblock Composing task knowledge with modular successor feature
  approximators.
\newblock In {\em International Conference on Learning Representations}, 2023.

\bibitem{bahdanau2019closure}
Dzmitry Bahdanau, Harm de~Vries, Timothy~J O'Donnell, Shikhar Murty, Philippe
  Beaudoin, Yoshua Bengio, and Aaron Courville.
\newblock Closure: Assessing systematic generalization of clevr models.
\newblock {\em arXiv preprint arXiv:1912.05783}, 2019.

\bibitem{lake2015human}
Brenden~M Lake, Ruslan Salakhutdinov, and Joshua~B Tenenbaum.
\newblock Human-level concept learning through probabilistic program induction.
\newblock {\em Science}, 350(6266):1332--1338, 2015.

\bibitem{romaszko2017vision}
Lukasz Romaszko, Christopher~KI Williams, Pol Moreno, and Pushmeet Kohli.
\newblock Vision-as-inverse-graphics: Obtaining a rich 3d explanation of a
  scene from a single image.
\newblock In {\em Proceedings of the IEEE International Conference on Computer
  Vision Workshops}, pages 851--859, 2017.

\bibitem{krishna2017visual}
Ranjay Krishna, Yuke Zhu, Oliver Groth, Justin Johnson, Kenji Hata, Joshua
  Kravitz, Stephanie Chen, Yannis Kalantidis, Li-Jia Li, David~A Shamma, et~al.
\newblock Visual genome: Connecting language and vision using crowdsourced
  dense image annotations.
\newblock {\em International journal of computer vision}, 123:32--73, 2017.

\bibitem{furrer2020compositional}
Daniel Furrer, Marc van Zee, Nathan Scales, and Nathanael Sch{\"a}rli.
\newblock Compositional generalization in semantic parsing: Pre-training vs.
  specialized architectures.
\newblock {\em arXiv preprint arXiv:2007.08970}, 2020.

\bibitem{hu2017learning}
Ronghang Hu, Jacob Andreas, Marcus Rohrbach, Trevor Darrell, and Kate Saenko.
\newblock Learning to reason: End-to-end module networks for visual question
  answering.
\newblock In {\em Proceedings of the IEEE International Conference on Computer
  Vision}, pages 804--813, 2017.

\bibitem{johnson2017clevr}
Justin Johnson, Bharath Hariharan, Laurens Van Der~Maaten, Li~Fei-Fei,
  C~Lawrence~Zitnick, and Ross Girshick.
\newblock Clevr: A diagnostic dataset for compositional language and elementary
  visual reasoning.
\newblock In {\em Proceedings of the IEEE Conference on Computer Vision and
  Pattern Recognition}, pages 2901--2910, 2017.

\bibitem{livska2018memorize}
Adam Li{\v{s}}ka, Germ{\'a}n Kruszewski, and Marco Baroni.
\newblock Memorize or generalize? searching for a compositional rnn in a
  haystack.
\newblock In {\em AEGAP Workshop, ICML}, 2018.

\bibitem{qiu2022improving}
Linlu Qiu, Peter Shaw, Panupong Pasupat, Pawel Nowak, Tal Linzen, Fei Sha, and
  Kristina Toutanova.
\newblock Improving compositional generalization with latent structure and data
  augmentation.
\newblock In {\em Proceedings of the 2022 Conference of the North American
  Chapter of the Association for Computational Linguistics: Human Language
  Technologies}, 2022.

\bibitem{xie2022coat}
Sirui Xie, Ari~S Morcos, Song-Chun Zhu, and Ramakrishna Vedantam.
\newblock {COAT}: Measuring object compositionality in emergent
  representations.
\newblock In {\em International Conference on Machine Learning}, 2022.

\bibitem{schmidhuber1990towards}
J\"urgen Schmidhuber.
\newblock Towards compositional learning in dynamic networks, technical report.
\newblock 1990.

\bibitem{keysers2020measuring}
Daniel Keysers, Nathanael Sch{\"a}rli, Nathan Scales, Hylke Buisman, Daniel
  Furrer, Sergii Kashubin, Nikola Momchev, Danila Sinopalnikov, Lukasz
  Stafiniak, Tibor Tihon, et~al.
\newblock Measuring compositional generalization: A comprehensive method on
  realistic data.
\newblock In {\em International Conference on Learning Representations}, 2020.

\bibitem{van2004lack}
Frank van~der Velde, Gwendid~T van der Voort van~der Kleij, and Marc de~Kamps.
\newblock Lack of combinatorial productivity in language processing with simple
  recurrent networks.
\newblock {\em Connection Science}, 16(1):21--46, 2004.

\bibitem{kaplan2020scaling}
Jared Kaplan, Sam McCandlish, Tom Henighan, Tom~B Brown, Benjamin Chess, Rewon
  Child, Scott Gray, Alec Radford, Jeffrey Wu, and Dario Amodei.
\newblock Scaling laws for neural language models.
\newblock {\em arXiv preprint arXiv:2001.08361}, 2020.

\bibitem{radford2021learning}
Alec Radford, Jong~Wook Kim, Chris Hallacy, Aditya Ramesh, Gabriel Goh,
  Sandhini Agarwal, Girish Sastry, Amanda Askell, Pamela Mishkin, Jack Clark,
  et~al.
\newblock Learning transferable visual models from natural language
  supervision.
\newblock In {\em International Conference on Machine Learning}, pages
  8748--8763. PMLR, 2021.

\bibitem{geiger2020perspective}
Mario Geiger, Leonardo Petrini, and Matthieu Wyart.
\newblock Perspective: A phase diagram for deep learning unifying jamming,
  feature learning and lazy training.
\newblock {\em arXiv preprint arXiv:2012.15110}, 2020.

\bibitem{wainwright2019high}
Martin~J. Wainwright.
\newblock {\em High-Dimensional Statistics: A Non-Asymptotic Viewpoint}.
\newblock Cambridge University Press, 2019.

\bibitem{taleb2020statistical}
Nassim~Nicholas Taleb.
\newblock Statistical consequences of fat tails: Real world preasymptotics,
  epistemology, and applications.
\newblock {\em arXiv preprint arXiv:2001.10488}, 2020.

\bibitem{cooper2021out}
Avi Cooper, Xavier Boix, Daniel Harari, Spandan Madan, Hanspeter Pfister,
  Tomotake Sasaki, and Pawan Sinha.
\newblock To which out-of-distribution object orientations are dnns capable of
  generalizing?
\newblock {\em arXiv preprint arXiv:2109.13445}, 2021.

\bibitem{baidya2021combining}
Avinash Baidya, Joel Dapello, James~J DiCarlo, and Tiago Marques.
\newblock Combining different v1 brain model variants to improve robustness to
  image corruptions in cnns.
\newblock In {\em SVRHM at NeurIPS 2021 Workshops}, 2021.

\bibitem{zhou2022domain}
Kaiyang Zhou, Ziwei Liu, Yu~Qiao, Tao Xiang, and Chen~Change Loy.
\newblock Domain generalization: A survey.
\newblock {\em IEEE Transactions on Pattern Analysis and Machine Intelligence},
  2022.

\bibitem{mahowald2023dissociating}
Kyle Mahowald, Anna~A. Ivanova, Idan~A. Blank, Nancy Kanwisher, Joshua~B.
  Tenenbaum, and Evelina Fedorenko.
\newblock Dissociating language and thought in large language models: a
  cognitive perspective.
\newblock {\em arXiv preprint arXiv:2301.06627}, 2023.

\bibitem{rebuffi17adapters}
S-A Rebuffi, H.~Bilen, and A.~Vedaldi.
\newblock Learning multiple visual domains with residual adapters.
\newblock In {\em Advances in Neural Information Processing Systems}, 2017.

\bibitem{rebuffi2018efficient}
Sylvestre-Alvise Rebuffi, Hakan Bilen, and Andrea Vedaldi.
\newblock Efficient parametrization of multi-domain deep neural networks.
\newblock In {\em CVPR}, 2018.

\bibitem{eastwood2022unitlevel}
Cian Eastwood, Ian Mason, and Christopher K.~I. Williams.
\newblock Unit-level surprise in neural networks.
\newblock In {\em Proceedings of ``I (Still) Can't Believe It's Not Better'' at
  NeurIPS 2021 Workshops}, volume 163 of {\em Proceedings of Machine Learning
  Research}. PMLR, 2022.

\bibitem{royer2020flexible}
Am{\'e}lie Royer and Christoph Lampert.
\newblock A flexible selection scheme for minimum-effort transfer learning.
\newblock In {\em Proceedings of the IEEE/CVF Winter Conference on Applications
  of Computer Vision}, pages 2191--2200, 2020.

\bibitem{lee2023surgical}
Yoonho Lee, Annie~S Chen, Fahim Tajwar, Ananya Kumar, Huaxiu Yao, Percy Liang,
  and Chelsea Finn.
\newblock Surgical fine-tuning improves adaptation to distribution shifts.
\newblock In {\em International Conference on Learning Representations}, 2023.

\bibitem{bengio2019meta}
Yoshua Bengio, Tristan Deleu, Nasim Rahaman, Rosemary Ke, S{\'e}bastien
  Lachapelle, Olexa Bilaniuk, Anirudh Goyal, and Christopher Pal.
\newblock A meta-transfer objective for learning to disentangle causal
  mechanisms.
\newblock {\em arXiv preprint arXiv:1901.10912}, 2019.

\bibitem{le2021analysis}
R{\'e}mi Le~Priol, Reza Babanezhad, Yoshua Bengio, and Simon Lacoste-Julien.
\newblock An analysis of the adaptation speed of causal models.
\newblock In {\em International Conference on Artificial Intelligence and
  Statistics}, pages 775--783. PMLR, 2021.

\bibitem{cohen2017emnist}
Gregory Cohen, Saeed Afshar, Jonathan Tapson, and Andre Van~Schaik.
\newblock {EMNIST}: Extending {MNIST} to handwritten letters.
\newblock In {\em 2017 International Joint Conference on Neural Networks
  (IJCNN)}, pages 2921--2926. IEEE, 2017.

\bibitem{krizhevsky2009cifar}
Alex Krizhevsky, Geoffrey Hinton, et~al.
\newblock Learning multiple layers of features from tiny images.
\newblock 2009.

\bibitem{ng2014facescrub}
Hong-Wei Ng and Stefan Winkler.
\newblock A data-driven approach to cleaning large face datasets.
\newblock In {\em 2014 IEEE International Conference on Image Processing
  (ICIP)}, pages 343--347. IEEE, 2014.

\bibitem{vogelsang2018potential}
Lukas Vogelsang, Sharon Gilad-Gutnick, Evan Ehrenberg, Albert Yonas, Sidney
  Diamond, Richard Held, and Pawan Sinha.
\newblock Potential downside of high initial visual acuity.
\newblock {\em Proceedings of the National Academy of Sciences},
  115(44):11333--11338, 2018.

\bibitem{lecun1998gradient}
Yann LeCun, L{\'e}on Bottou, Yoshua Bengio, and Patrick Haffner.
\newblock Gradient-based learning applied to document recognition.
\newblock {\em Proceedings of the IEEE}, 86(11):2278--2324, 1998.

\bibitem{he2016resnet}
Kaiming He, Xiangyu Zhang, Shaoqing Ren, and Jian Sun.
\newblock Deep residual learning for image recognition.
\newblock In {\em Proceedings of the IEEE Conference on Computer Vision and
  Pattern Recognition}, pages 770--778, 2016.

\bibitem{szegedy2016inception}
Christian Szegedy, Vincent Vanhoucke, Sergey Ioffe, Jon Shlens, and Zbigniew
  Wojna.
\newblock Rethinking the inception architecture for computer vision.
\newblock In {\em Proceedings of the IEEE Conference on Computer Vision and
  Pattern Recognition}, pages 2818--2826, 2016.

\bibitem{yamada2022transformer}
Moyuru Yamada, Vanessa D'Amario, Kentaro Takemoto, Xavier Boix, and Tomotake
  Sasaki.
\newblock {Transformer Module Networks} for systematic generalization in visual
  question answering.
\newblock Technical Report CBMM Memo No.~121, Ver.2, Center for Brains, Minds
  and Machines, 2023.

\bibitem{anselmi2016unsupervised}
Fabio Anselmi, Joel~Z Leibo, Lorenzo Rosasco, Jim Mutch, Andrea Tacchetti, and
  Tomaso Poggio.
\newblock Unsupervised learning of invariant representations.
\newblock {\em Theoretical Computer Science}, 633:112--121, 2016.

\bibitem{poggio2016visual}
Tomaso~A Poggio and Fabio Anselmi.
\newblock {\em Visual cortex and deep networks: learning invariant
  representations}.
\newblock MIT press, 2016.

\bibitem{kong2022partial}
Lingjing Kong, Shaoan Xie, Weiran Yao, Yujia Zheng, Guangyi Chen, Petar
  Stojanov, Victor Akinwande, and Kun Zhang.
\newblock Partial disentanglement for domain adaptation.
\newblock In {\em International Conference on Machine Learning}, pages
  11455--11472. PMLR, 2022.

\bibitem{sun2023invariance}
Qingyao Sun, Kevin Murphy, Sayna Ebrahimi, and Alexander D'Amour.
\newblock Beyond invariance: Test-time label-shift adaptation for distributions
  with "spurious" correlations.
\newblock {\em arXiv preprint arXiv:2211.15646}, 2023.

\bibitem{sperandio2015size}
Irene Sperandio and Philippe~A. Chouinard.
\newblock The mechanisms of size constancy.
\newblock {\em Multisensory Research}, 28(3-4):253 -- 283, 2015.

\bibitem{kohler1970gestalt}
Wolfgang K{\"o}hler.
\newblock {\em Gestalt psychology: An introduction to new concepts in modern
  psychology}, volume~18.
\newblock WW Norton \& Company, 1970.

\bibitem{sarkar2023deephys}
Anirban Sarkar, Matthew Groth, Ian Mason, Tomotake Sasaki, and Xavier Boix.
\newblock Deephys: Deep electrophysiology, debugging neural networks under
  distribution shifts.
\newblock {\em arXiv preprint arXiv:2303.11912}, 2023.

\bibitem{olah2020zoom}
Chris Olah, Nick Cammarata, Ludwig Schubert, Gabriel Goh, Michael Petrov, and
  Shan Carter.
\newblock Zoom in: An introduction to circuits.
\newblock {\em Distill}, 2020.
\newblock https://distill.pub/2020/circuits/zoom-in.

\bibitem{paszke2019pytorch}
Adam Paszke, Sam Gross, Francisco Massa, Adam Lerer, James Bradbury, Gregory
  Chanan, Trevor Killeen, Zeming Lin, Natalia Gimelshein, Luca Antiga, et~al.
\newblock Pytorch: An imperative style, high-performance deep learning library.
\newblock {\em Advances in Neural Information Processing Systems}, 32, 2019.

\end{thebibliography}

\newpage
\normalsize
\appendix 
\addcontentsline{toc}{section}{Appendix} % Add the appendix text to the document TOC
\part{Appendix} % Start the appendix part
\parttoc 
\newpage

\section{Choice of Corruptions}
\label{app:corruption-choice}

The choice of corruptions used in our compositional robustness task is quite subtle. We want to ensure a good mixture of different types of corruptions and the compositions they form, but without creating a compositional space that is so big that it becomes prohibitively expensive to evaluate. Due to the exponential increase in the number of possible compositions as the number of elemental corruptions increases, and in order to reduce computational costs, we make the following concessions: (i) we keep the total number of elemental corruptions low whilst ensuring a good mixture of elemental corruptions; (ii) we include compositions constructed from every combination of elemental corruptions but sample the possible permutations (orderings) of elemental corruptions that make up a composition (see Appendix~\ref{app:sampling}); (iii) we do not consider the 3D projection problem (see Appendix~\ref{app:3d-projection}). 

\begin{figure*}[b]
    \centering
    \includegraphics[trim={0 17cm 0 0}, width=\textwidth]{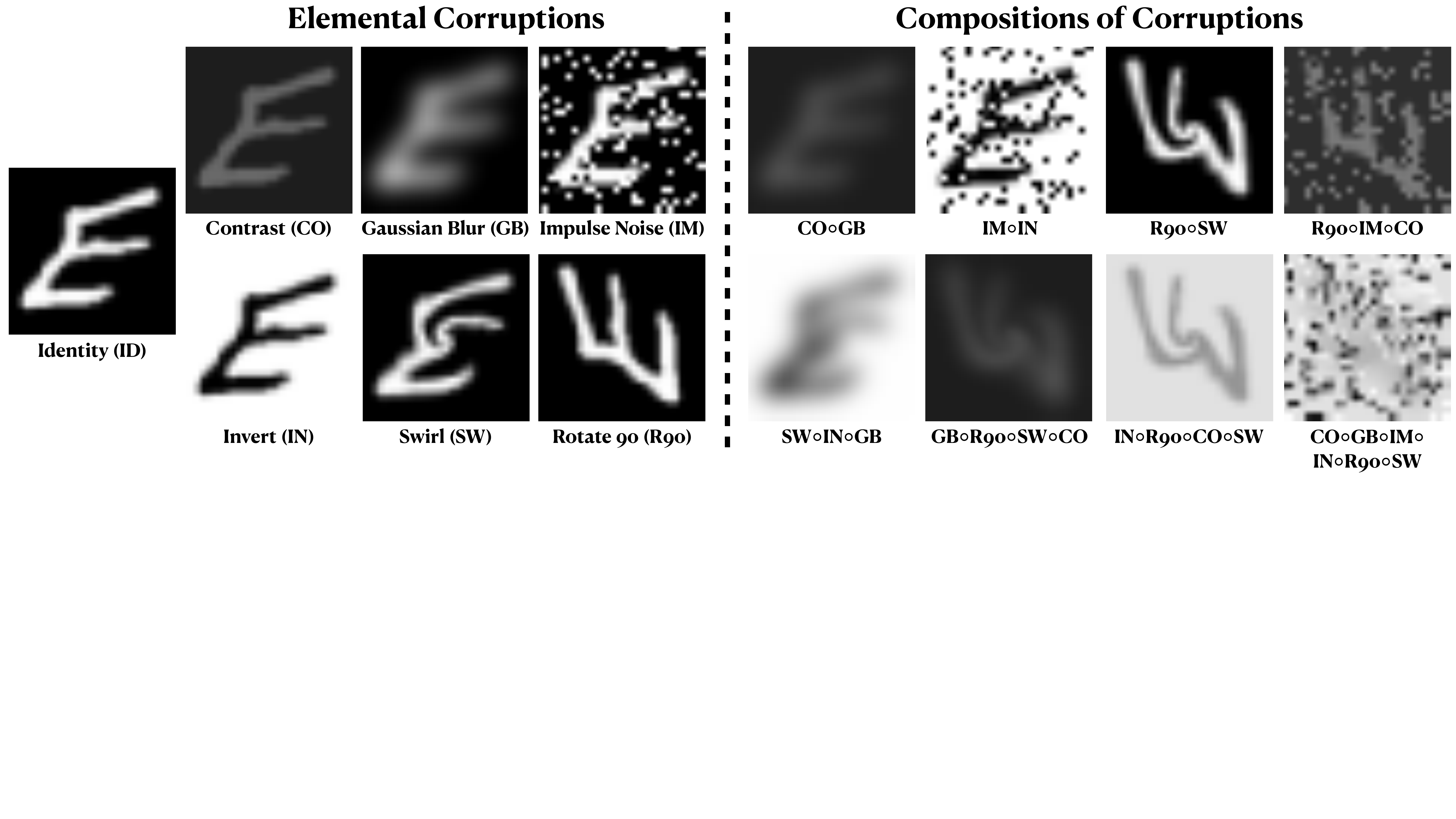}
    \caption{The compositional robustness task for \emnist.}
    \label{fig:benchmark-emnist}
\end{figure*}

\begin{figure*}[tb]
    \centering
    \includegraphics[trim={0 17cm 0 0}, width=\textwidth]{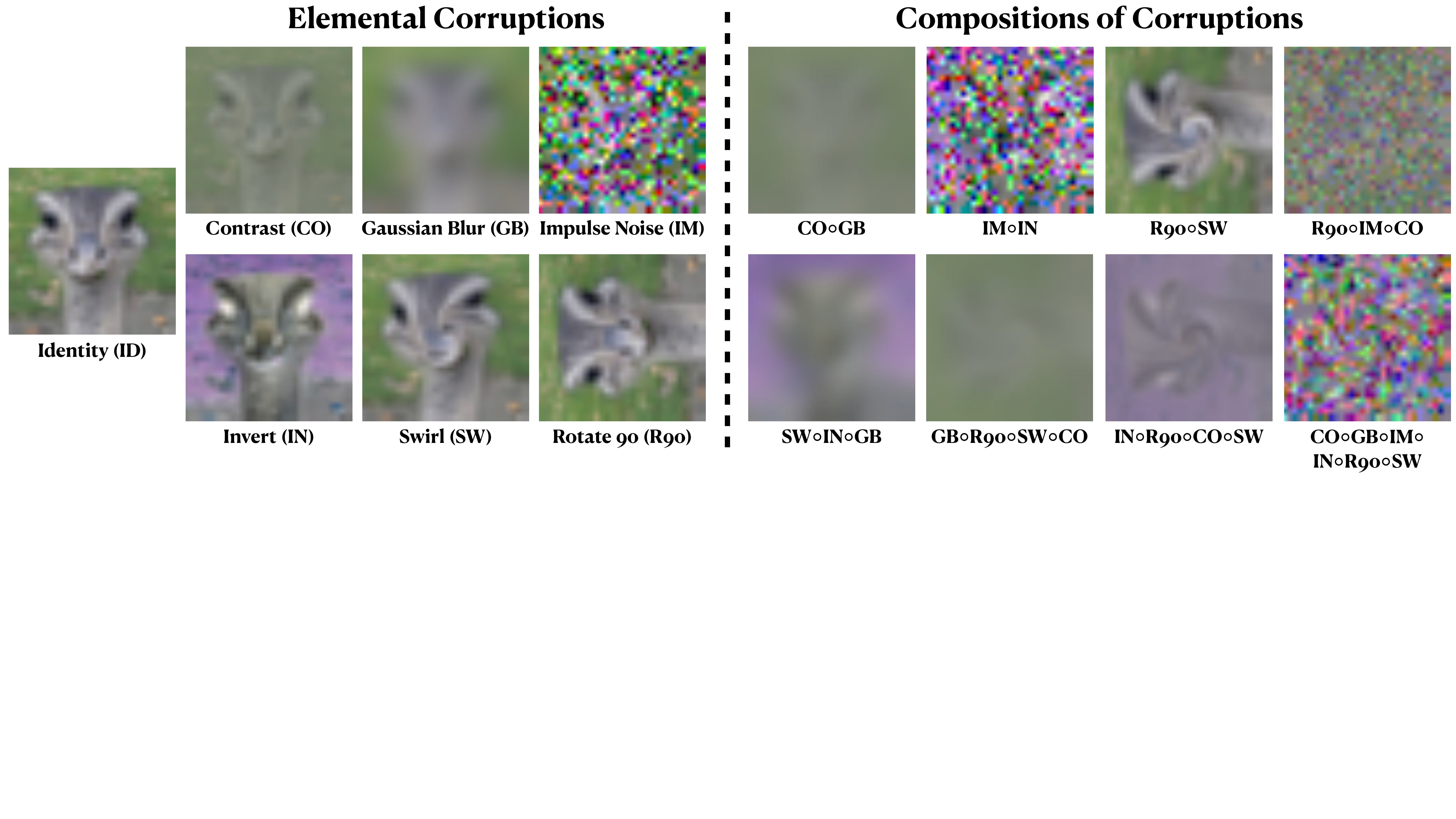}
    \caption{The compositional robustness task for \cifar.}
    \label{fig:benchmark-cifar}
\end{figure*}

As discussed in Section~\ref{sec:setup}, along with the \emph{Identity (ID)} data we consider the corruptions, \emph{Contrast (CO)}, \emph{Gaussian Blur (GB)}, \emph{Impulse Noise (IM)}, \emph{Invert (IN)}, \emph{Rotate 90$\degree$ (R90)} and \emph{Swirl (SW)}, which can be seen for \emnist\ and \cifar\ in Figures~\ref{fig:benchmark-emnist} and  \ref{fig:benchmark-cifar} respectively. We consider two different behaviors that corrupting functions may exhibit and select this set of corruptions to get a mixture of behaviors. Firstly, corruptions can be local or long-ranged, where images under local corruptions (such as \emph{Invert}) can be transformed to the \emph{Identity} image by applying a patch-wise operation. On the other hand, long-ranged corruptions (such as \emph{Rotate 90$\degree$}) require a holistic understanding of the image. Secondly, corruptions can be lossless or lossy, where lossless corruptions lose no information so can be perfectly inverted and lossy corruptions may lose information due to randomness or the application of non-invertible corrupting functions. Finally, our implementation allows for corruptions to be applied with differing severity, for example by adding more or fewer random pixels for \emph{Impulse Noise} or by increasing or decreasing the Gaussian filter size when creating \emph{Gaussian Blur}. In our experiments we keep the severity fixed as varying the severity would again increase the size of the compositional space.

\subsection{Sampling and Commutativity}
\label{app:sampling}

Our set of elemental corruptions allows us to consider compositions made up of up to six corruptions at once (we do not allow for repeated application of elemental corruptions). As not all elemental corruptions are commutative under composition (e.g. \emph{IM}$\circ$\emph{GB} $\neq$ \emph{GB}$\circ$\emph{IM}), we must take into account the possible orderings of elemental corruptions when constructing compositions. When taking into account possible orderings there are  $^6\!P_2 = 30$ possible orderings of two corruptions but  $^6\!P_6 = 720$ possible ordering of six corruptions, where $^n\!P_r = n! / (n-r)!$, counts the number of possible permutations. As we don't want results to be dominated by compositions of larger numbers of corruptions and to reduce the number of compositional test domains, we \emph{sample} the possible orderings.

For compositions of two corruptions, we consider all possible orderings giving $^6\!P_2 = 30$ compositions. For compositions of more than two corruptions we aim to get as close to $^6\!P_2 = 30$ test domains as possible whilst maintaining a balance of the possible unique combinations of elemental corruptions. This means we first calculate the number of unique combinations as  $^n\!C_r = n! / r!(n-r)!$, where $n$ is the total number of elemental corruptions and $r$ is the number of elemental corruptions in the compositions we are considering. We then sample the same number of possible orderings of each unique combination until we get as close as possible to $30$ domains. As an example, for compositions of three corruptions $^6\!C_3 = 20$, so we have twenty unique combinations of three elemental corruptions. For each unique combination we sample two possible orderings, giving forty test domains. For compositions of four corruptions we have fifteen unique combinations so we again sample two possible orderings,  for compositions of five corruptions we have six unique combinations so we sample five orderings and for compositions of six corruptions there is only one unique combinations so we sample thirty different orderings.

% We perform this sampling both to reduce computational load and to ensure the results are not dominated by compositions of larger numbers of corruptions which tend to be harder to solve.

    % - For singles there is no order to consider. We also train on clean data. This is $7$ to test on.
    % - For pairs we consider all pairs and all orders. Whilst some are commutative (rotate 90, invert) other are not (blur and noise, noise and contrast). This is $6P2 = 30$.
    % - To keep the evaluation space down for combinations of 3 or more we take every combination but sample the possible orders (otherwise we have $\sim 2000$ permutations and evaluation takes too long). To do the sampling we take each combination and sample the order until we have a total close to $30$, but we make sure the combinations themselves are equally represented. So e.g. for compositions of 4 there are 15 possible combinations, so each combination has 2 orders sampled from the 24 possible..

\subsection{The 3D Projection Problem}
\label{app:3d-projection}

A final point of interest when choosing which elemental corruptions to consider is the problem of 3D projection. There are certain corruptions that occur in natural data that are inherently 3-dimensional, yet we only see the results as a projection onto a 2-dimensional image plane. This fact introduces complexity in the way corruptions can be applied and composed if we are aiming to create a system with vision that is as robust as humans. 

\begin{figure*}[tb]
    \centering
    \includegraphics[trim={8cm 12cm 15cm 3cm},clip,width=\textwidth]{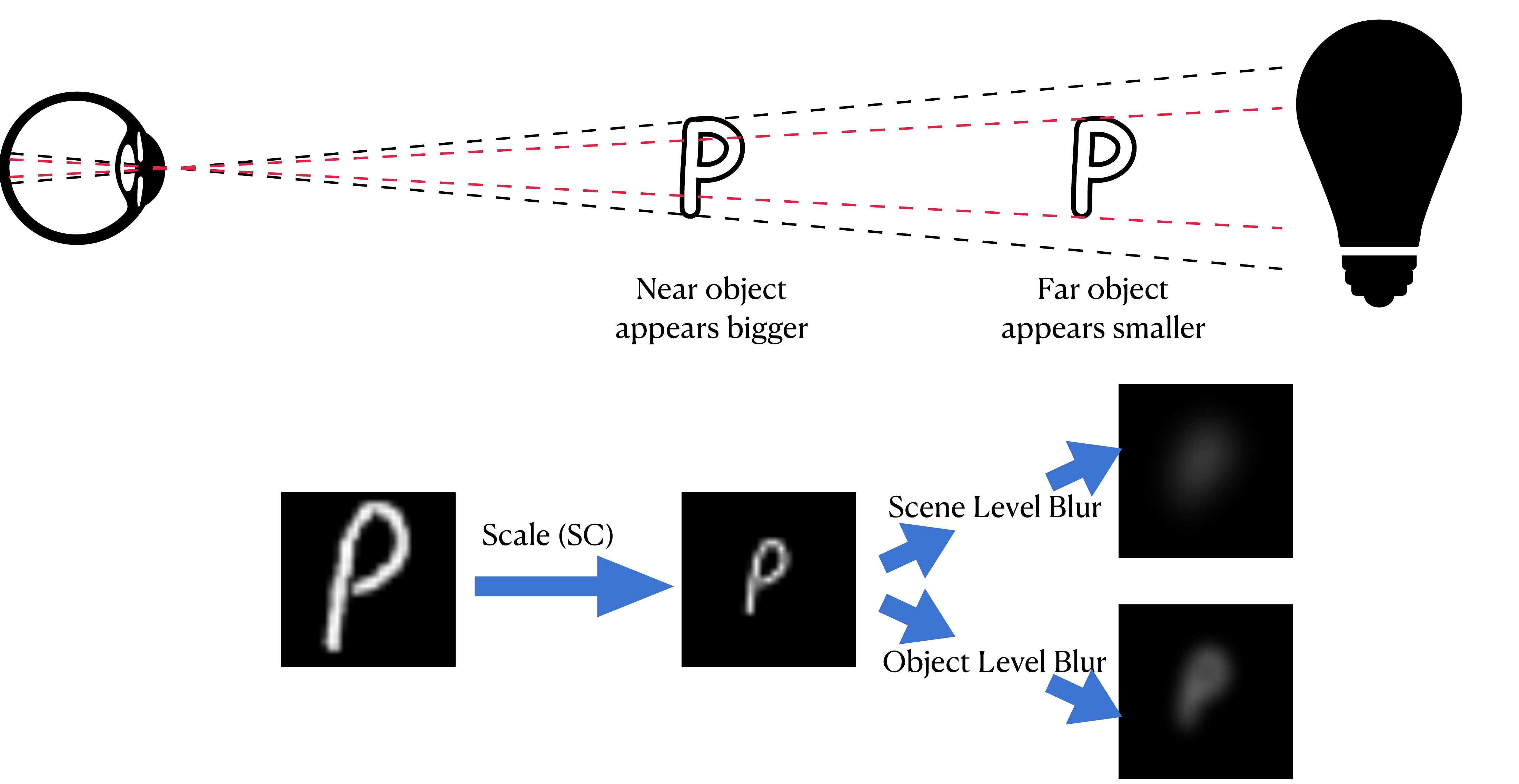}
    \caption{The 3D projection problem. Corruptions of 2D images can represent 3D processes, for example scaling an image represents moving the object further away from the viewer (top). When composing corruptions, this can lead to different orderings of corruptions representing different 3D processes (bottom).}
    \label{fig:eye-schematic}
\end{figure*}

To see the problem, consider the corruption \emph{Scale (SC)}, where we create a zoomed out version of a base image (see Figure~\ref{fig:eye-schematic}). Imagine that we then also consider the composition of \emph{Scale} with \emph{Gaussian Blur}. \emph{SC}$\circ$\emph{GB} creates a very different image to \emph{GB}$\circ$\emph{SC}, but more importantly these represent fundamentally different processes in the 3D world. If scaling is applied before blurring this corresponds to the case where there is a fixed amount of blur in the scene (e.g. because of an eye condition) and the object we care about is moved further away from the viewer. On the other hand if blurring is applied before scaling this corresponds to the case where the object itself is blurry (e.g. because of damage around the edges). This process is depicted in Figure~\ref{fig:eye-schematic}.

The point of this discussion is to demonstrate that applying a corruption at the scene level can be fundamentally different from applying a corruption at the object level. Whilst this can be taken into account (e.g. by changing the order of \emph{Gaussian Blur} and \emph{Scale}), we aim to avoid this situation by only considering corruptions where changing the ordering under composition does not change the composition from a scene level process to an object level process (or vice versa). This makes our task more practical as we can apply it to any image classification dataset. Since we may not even consciously perceive the effect of scaling accurately \citep{sperandio2015size, kohler1970gestalt}, future work may find that different processes in 3D space should be handled in different ways or at different levels of abstraction. 

\newpage

\section{Module Implementation and Interpretability}
\label{app:modules}

\begin{figure*}[b]
    \centering
    \includegraphics[trim={3cm 7cm 4cm 9cm},clip,width=\textwidth]{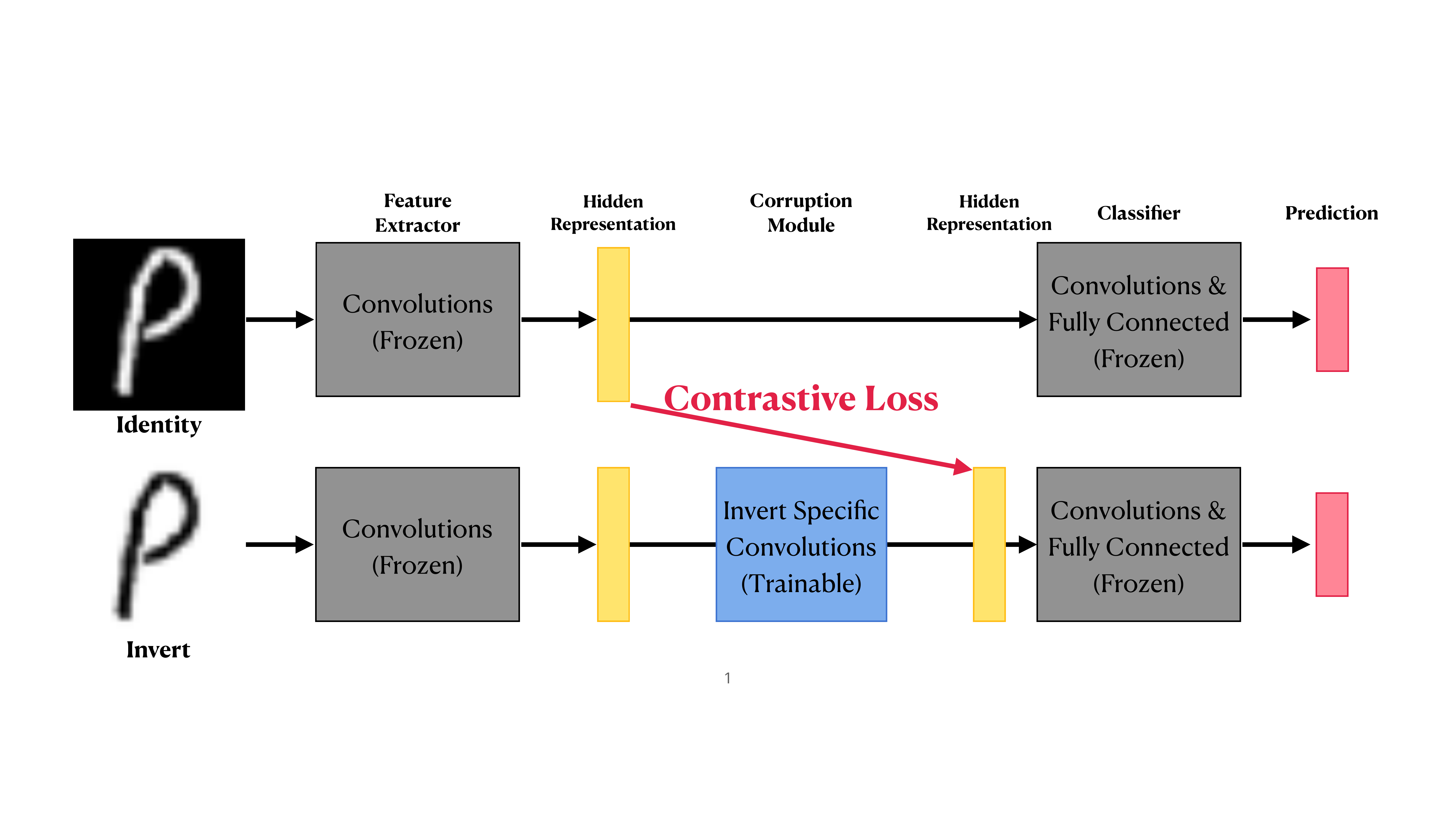}
    \caption{Module training diagram. After pre-training on \emph{Identity} data, shared network parameters are frozen (gray boxes) and a module (blue box) is trained to align the representation of the corrupted \emph{Invert} image with the representation of the \emph{Identity} image using the contrastive loss. In this figure, apart from those of the module, all parameters are identical between the top and bottom networks.}
    \label{fig:modules}
\end{figure*}

Figure~\ref{fig:modules} shows the training process for a module trained on the \emph{Invert} corruption. First a network is trained on \emph{Identity} data to learn parameters $\bm{\theta}_{\textit{shared}}$. These weights are then frozen (gray boxes in Figure~\ref{fig:modules}) and a module is trained to `undo' the \emph{Invert} corruption in latent space (blue box in Figure~\ref{fig:modules}). To train the module, the contrastive loss is used to align representations of \emph{Identity} data before the module is applied with representations of \emph{Invert} data after the module is applied. As described in the main text, we also use the cross entropy loss to ensure classification accuracy is maintained.

\begin{figure*}[tb]
    \centering
    \includegraphics[trim={5cm 0cm 16.5cm 1cm},clip,width=\textwidth]{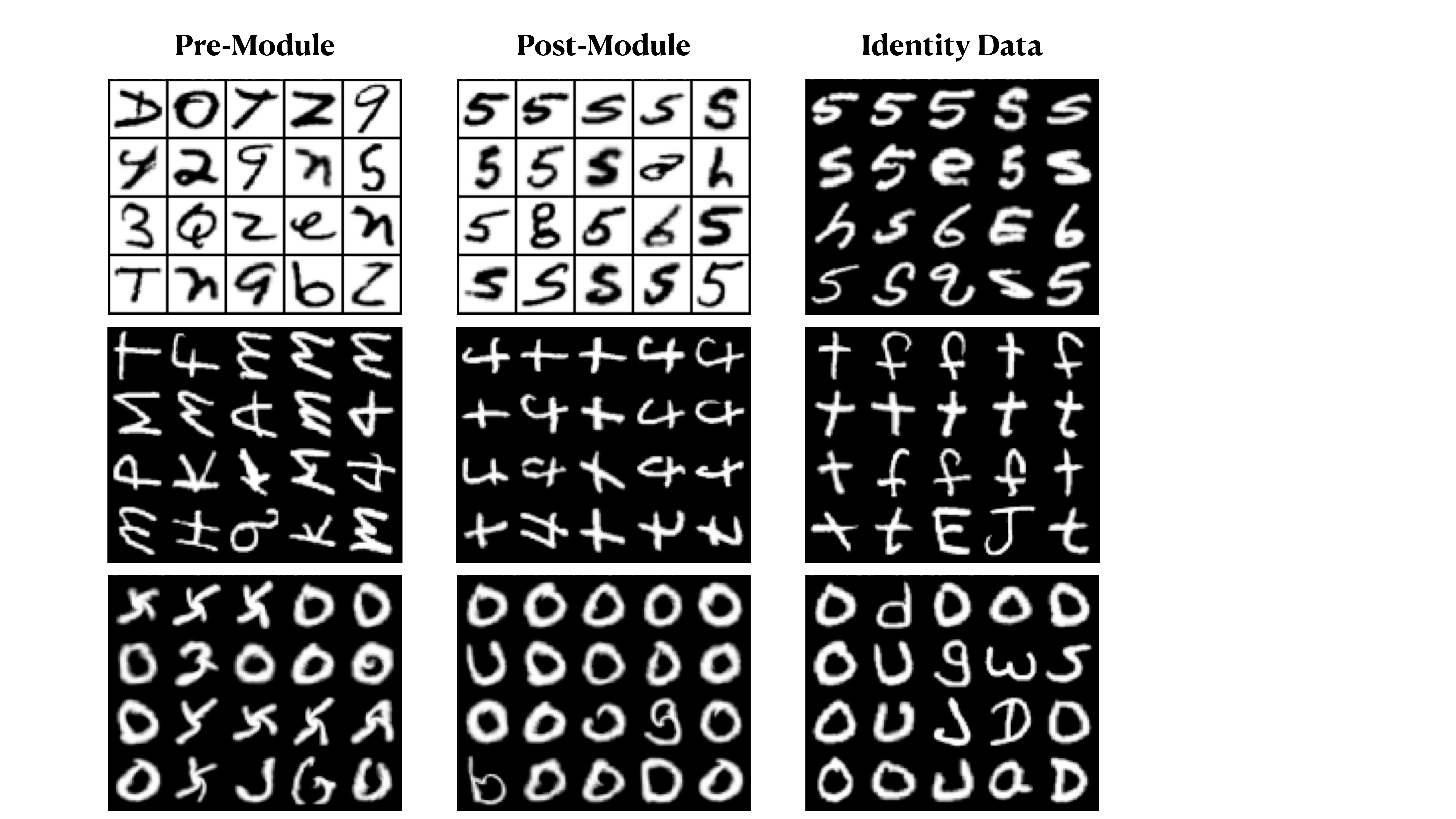}
    \caption{Training modules to undo elemental corruptions. Images in this grid represent some of the images that maximally activate a neuron. The first column shows a neuron before the module is applied, this is equivalent to the images that the neuron is tuned to for a network trained only on \emph{Identity} data. The second column shows the same neuron after the module is applied. The third column shows how the \emph{Identity} data activates this neuron. By comparing across columns we can see that modules learns to align the hidden representations so that neurons fire for similar class instances between \emph{Identity} and corrupted data. Top to bottom the corruptions in the rows are, \emph{Invert}, \emph{Rotate 90$\degree$} and \emph{Swirl}.}
    \label{fig:deephys}
\end{figure*}

Using interpretability tool Deephys \citep{sarkar2023deephys}, we visualize the effect of modules trained in this way in Figure~\ref{fig:deephys} . We find neurons which are initially activated by very different class instances when comparing \emph{Identity} data with corrupted data, but after applying the module, neurons fire for similar class instances between the \emph{Identity} and corrupted data.

\newpage

\section{Elemental and Composition Invariance Scores}
\label{app:invariance-scores}

This section first gives the full formalization for our elemental invariance score and composition invariance score following Madan et al. \citep{madan2022ood}. We then give a worked example with and exemplar activation grid to further detail the invariance scores. 

For every neuron in the penultimate layer of a network (after applying modules if applicable) we calculate the mean activation per domain-category pair over all test data. The activations are normalized by the maximum firing of the neuron over all domains with any dead neurons (with maximum firing less than $10^{-6}$) discarded. For a specific test domain we select only the domain-category pairs where the domain is either the test domain itself or one of the elemental corruptions used to create the composition for the test domain. Using \emph{CO}$\circ$\emph{GB} as an example for \cifar, this would leave us with a separate grid of size $4 \times 10$ for every neuron, where the rows are the corruptions \emph{ID}, \emph{CO}, \emph{GB} and \emph{CO}$\circ$\emph{GB} and the columns are the $10$ categories of \cifar\ (an example for one neuron is shown in Table~\ref{tab:activation-grid}). % Since we always use ReLU activations each entry in the activation grid represents the activation for that neuron as a percentage of the maximum firing rate of the neuron.

This domain specific activation grid for a single neuron is then normalized again so that all values lie between $0$ and $1$ by subtracting the minimum value in the grid from every cell and dividing by the difference between the maximum and minimum values. We notate the activation values by $a_{i,j}$, with $i$ referencing the domain and $j$ the category. Additionally we take the number of elemental corruption domains in the grid to be indexed $1, \dots, E$ and the composition to have index $E+1$, that is, $i \in \{1, \dots, E+1\}$. Taking the view that neurons can be interpreted as feature detectors \citep{olah2020zoom, sarkar2023deephys}, we select the preferred category, $j^*$, on the training domains as the category for which the neuron maximally activates, $j^* = {\operatorname{argmax}}_j \sum_{i=1}^E a_{i,j}$. We then calculate the \emph{elemental invariance score}, $I_e$ as the maximum difference in activations amongst the elemental corruptions, with the idea that this score should be high when all elemental corruptions activate the neuron in a similar way. We additionally calculate the \emph{composition invariance score}, $I_c$, which measures how similarly the neuron activates on the composition compared to the closest elemental corruption. 

\begin{equation}
    I_e = 1 - (\max_i a_{i,j^*} - \min_i a_{i,j^*}), \qquad I_c = 1 - \min \{|a_{i,j^*} - a_{E+1, j*}|\}_{i=1}^E.
\end{equation}

These scores always lie between $0$ and $1$, with higher numbers representing more invariant representations. We calculate these scores for every neuron in the penultimate layer of the network and report the median scores over all (non-dead) neurons in our results.

% Note that this is very similar to just the L1 distance. If we did average pairwise L1 distance this is same as the difference of average firing rates. By doing per-category and this normalisation etc. we get something a bit more granular -> expand this idea a little?
\subsection{Worked example}
% This is the actual table for neuron 4, \emph{CO}$\circ$\emph{GB}, seed 13579111, although I am unsure of the row ordering. Elemental invariance score: 0.9554243570496728. Composition invariance score: 0.6821811252656317.
\begin{table}[tb]
\caption{An exemplar activation grid}
\label{tab:activation-grid}
\begin{tabular}{@{}lllllllllll@{}}
\toprule
      & Cat. 1  & Cat. 2  & Cat. 3  & Cat. 4  & Cat. 5  & Cat. 6  & Cat. 7  & Cat. 8  & Cat. 9  & Cat. 10 \\ \midrule
\emph{CO}                 & 0.002 & 0.007 & 0.038 & 0.089 & 0.039 & 0.794 & 0.998 & 0.015 & 0.022 & 0.005 \\
\emph{GB}                 & 0.011 & 0.021 & 0.070 & 0.144 & 0.061 & 0.733 & 0.955$\dagger$ & 0.043 & 0.029 & 0.020 \\
\emph{ID}                 & 0.020 & 0.004 & 0.051 & 0.090 & 0.039 & 0.791 & 1.000$*$ & 0.016 & 0.018 & 0.000 \\ \midrule
\emph{CO}$\circ$\emph{GB} & 0.035 & 0.102 & 0.109 & 0.126 & 0.087 & 0.415 & 0.638$\ddagger$ & 0.078 & 0.116 & 0.138 \\ \bottomrule
\end{tabular}
\end{table}

Table~\ref{tab:activation-grid} shows an exemplar activation grid for a single neuron for the test domain containing the composition \emph{CO}$\circ$\emph{GB} on \cifar. We see the $4$ rows consist of the composition alongside the elemental corruptions that are relevant for \emph{CO}$\circ$\emph{GB}, and the $10$ columns for each of the $10$ categories of \cifar, creating domain-category pairs. 

To calculate the invariance scores for this example we first find the preferred category as $j^* = {\operatorname{argmax}}_j \sum_{i=1}^E a_{i,j}$, which indicates that this neuron activates maximally for category $7$. The elemental invariance score is the worst case difference amongst the elemental corruption activations for this category (the maximum is marked $*$, and the minimum is marked $\dagger$).

\begin{align*}
    I_e &= 1 - (\max_i a_{i,j^*} - \min_i a_{i,j^*}) \\
    &= 1 - (1.000 - 0.955) \\
    &= 0.955
\end{align*}

The composition invariance score finds the activation amongst the elemental corruptions that is closest to the composition's activation (marked $\ddagger$) for the preferred category.

\begin{align*}
    I_c &= 1 - \min \{|a_{i,j^*} - a_{E+1, j*}|\}_{i=1}^E \\
    &= 1 - \min \{|0.998 - 0.638|, |0.995 - 0.638|, |1.000 - 0.638| \} \\
    &= 0.683
\end{align*}

For this particular neuron, we would deduce that the elemental corruptions have relatively invariant activations whereas the activations are less invariant when we include the composition.

\newpage
\section{Network Architecture Details}
\label{app:implementation}
This appendix gives the specific architecture of the simple convolutional network used for \emnist\ experiments in Table~\ref{table:arch-details}. For \cifar\ we use ResNet18 and for \facescrub\ Inception-v3. In both cases we use the official PyTorch \citep{paszke2019pytorch} implementations of the architectures. Rather than giving a lengthy description of the possible architectures for modules between every layer of these networks we refer the reader to the associated code repository (file lib/networks.py). The architectures of the auto-encoders used in Section~\ref{sec:critical-analysis} can also be found in this file.

\begin{table}[h!]
\caption{The network architecture used for \emnist\ experiments. For convolutions, the weights-shape is: \textit{number of input channels} $\times$ \textit{number of output channels} $\times$ \textit{filter height} $\times$ \textit{filter width}.}
\label{table:arch-details}
\begin{tabular}{llccccc}
\toprule
\textbf{Block }      & \textbf{Weights-Shape} & \textbf{Stride} & \textbf{Padding} & \textbf{Activation} & \textbf{Dropout Prob.} &  \\ \toprule
Convolution   & $3 \times 64 \times 5 \times 5$   & $2$ & $2$ & ReLU & $0.1$ &  \\
\midrule
Convolution   & $64 \times 128 \times 5 \times 5$ & $2$ & $2$ & ReLU & $0.3$ &  \\
\midrule
Convolution   & $128 \times 256 \times 5\times 5$ & $2$ & $2$ & ReLU & $0.5$ &  \\
\midrule
Convolution   & $256 \times 256 \times 5\times 5$ & $2$ & $2$ & ReLU & $0.5$ &  \\
\midrule
Linear      & $1024 \times 512$ & N/A & N/A & ReLU & $0.5$ &  \\
\midrule
Linear &    $512 \times \text{Number of Classes}$       & N/A & N/A & Softmax & $0$  & \\ \bottomrule
\end{tabular}
\end{table}

\section{Invariance Scores for All Datasets}
\label{app:invariance}

% \subsection{Invariance Summary Plots}

\begin{figure}[tb]
    \centering
    \begin{subfigure}[t]{0.9\textwidth}
        \centering
        \includegraphics[trim={0 60cm 0 0},clip,width=\linewidth]{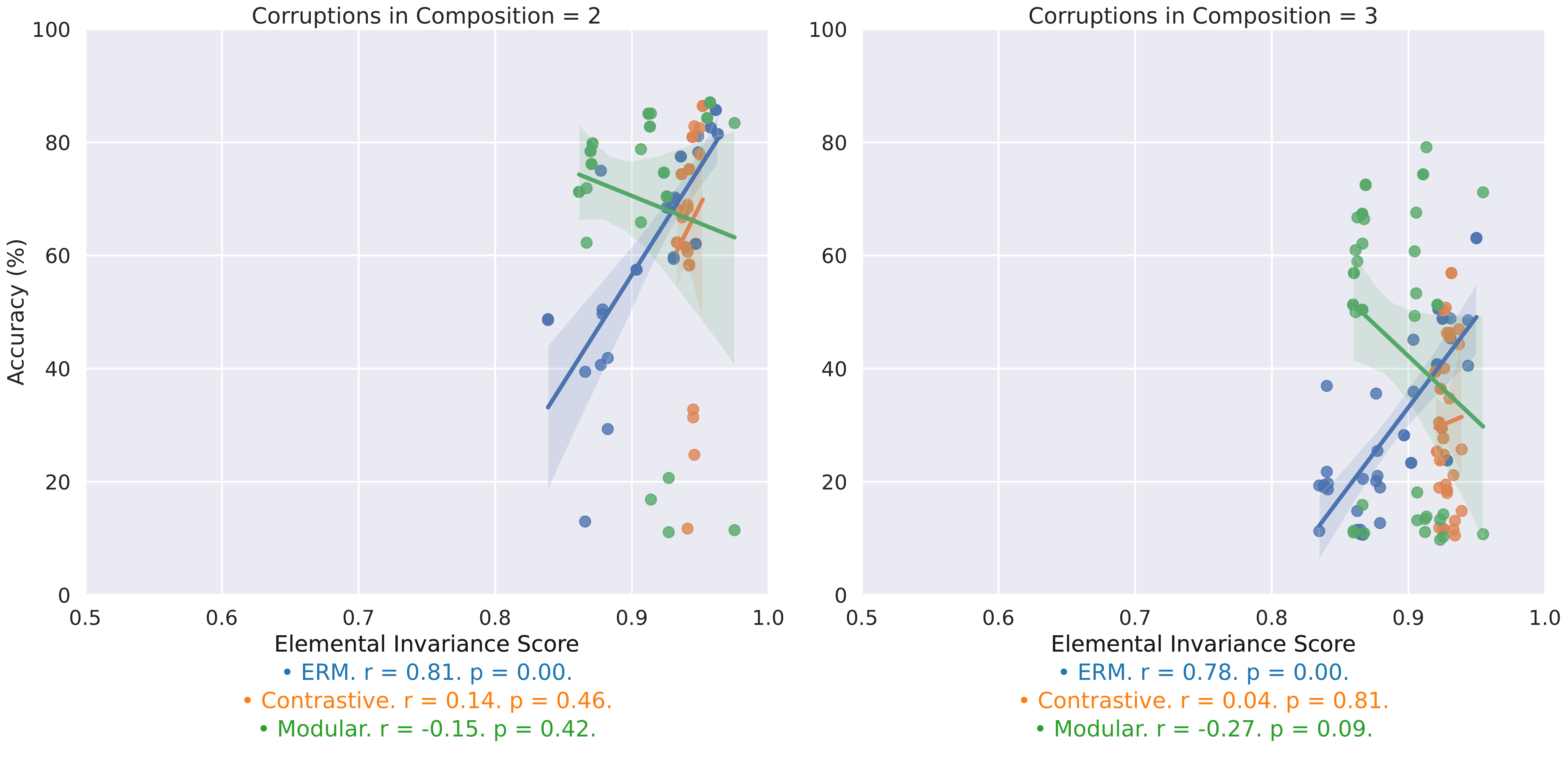} \\
    \end{subfigure}
    \begin{subfigure}[t]{0.9\textwidth}
        \centering
        \includegraphics[trim={0 60cm 0 0},clip,width=\linewidth]{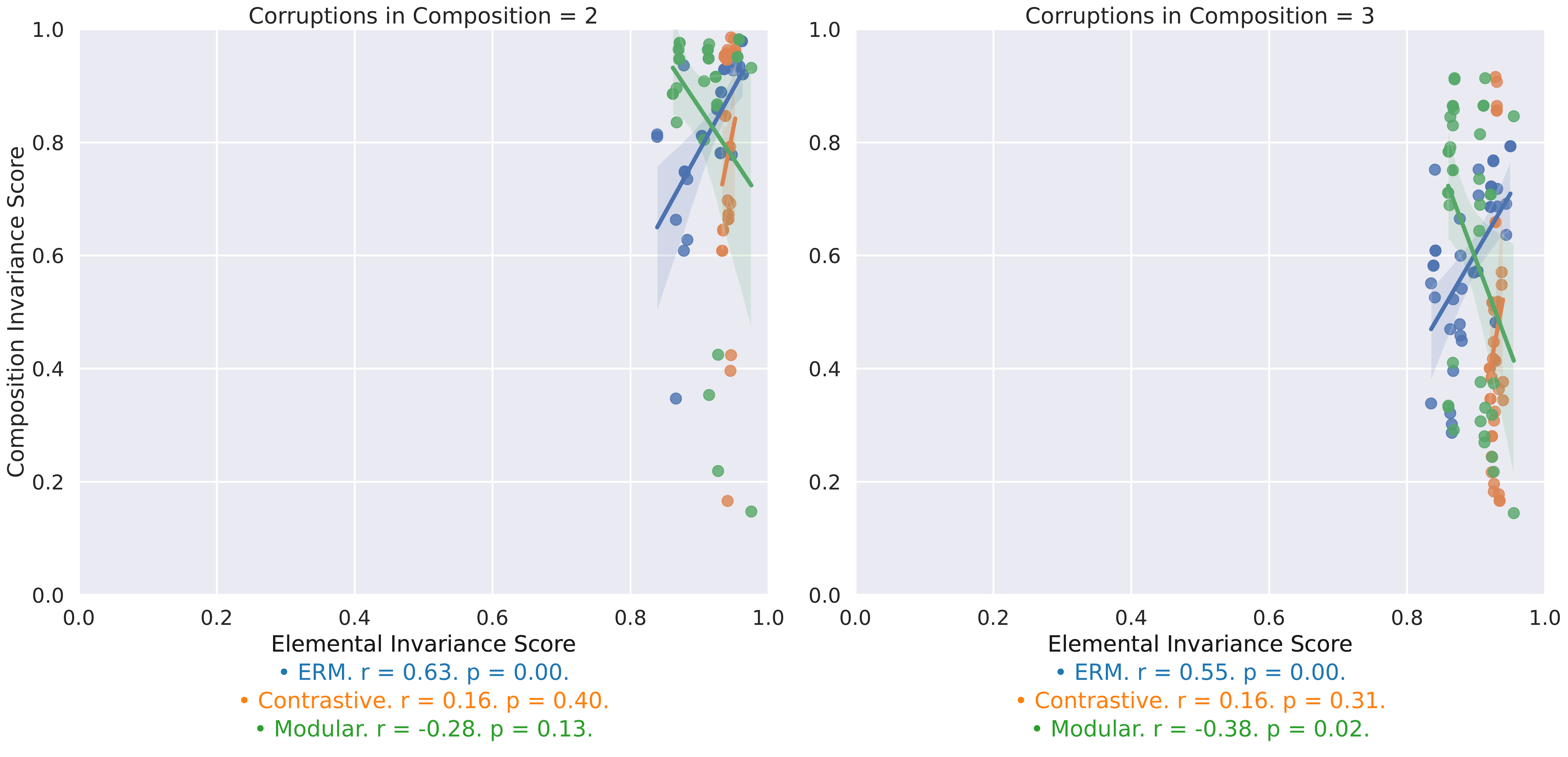} \\
    \end{subfigure}
    \begin{subfigure}[t]{0.9\textwidth}
        \centering
        \includegraphics[trim={0 60cm 0 0},clip,width=\linewidth]{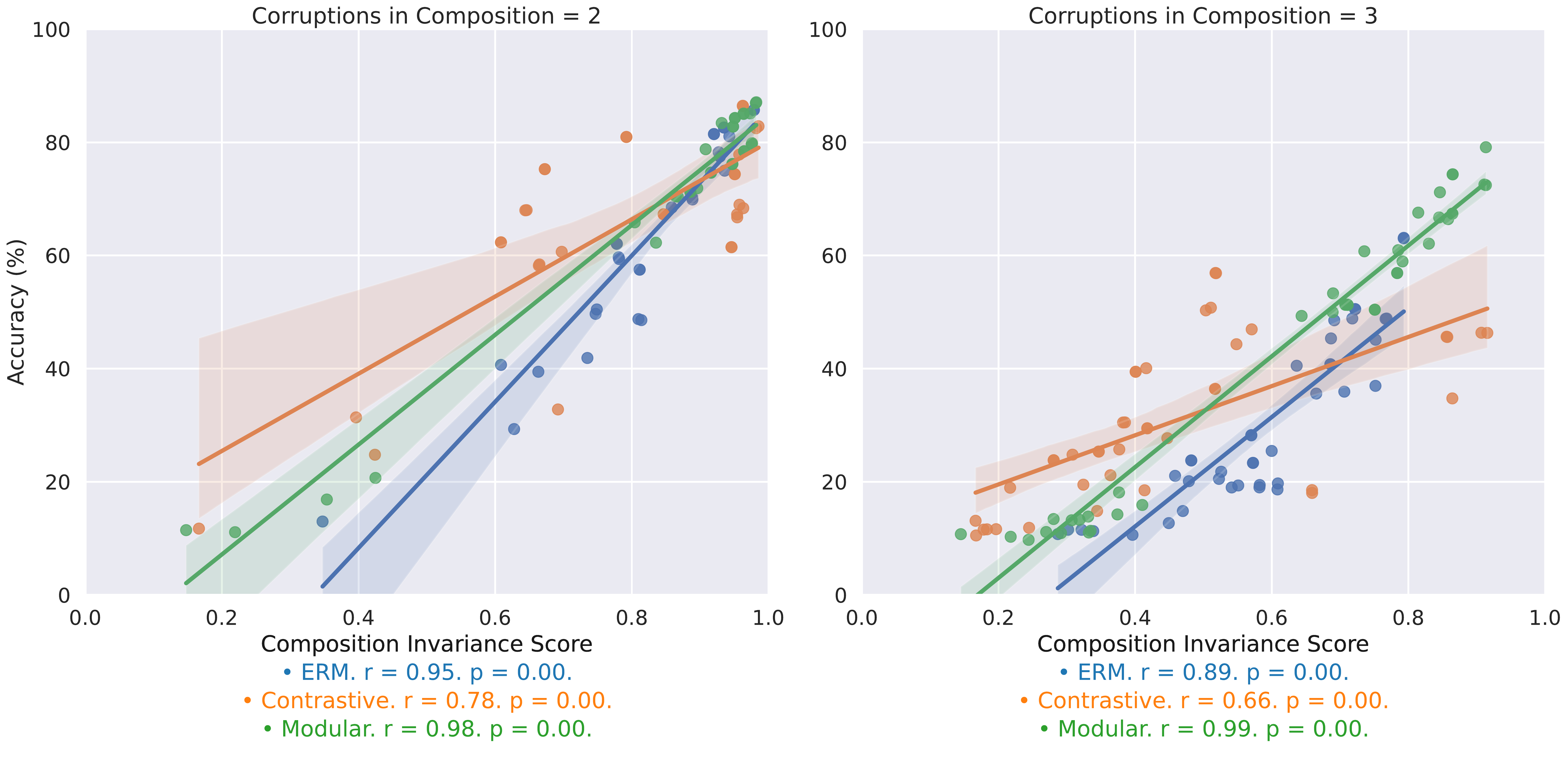} \\
    \end{subfigure}
    \vspace{-1.5mm}
    \caption{Correlating invariance scores with compositional robustness for \cifar. These plots plot the same relationships as in Figure~\ref{fig:invariance-emnist}.}
    \label{fig:invariance-cifar}
    \vspace{-6.5mm}
\end{figure}

\begin{figure}[tb]
    \centering
    \begin{subfigure}[t]{0.9\textwidth}
        \centering
        \includegraphics[trim={0 60cm 0 0},clip,width=\linewidth]{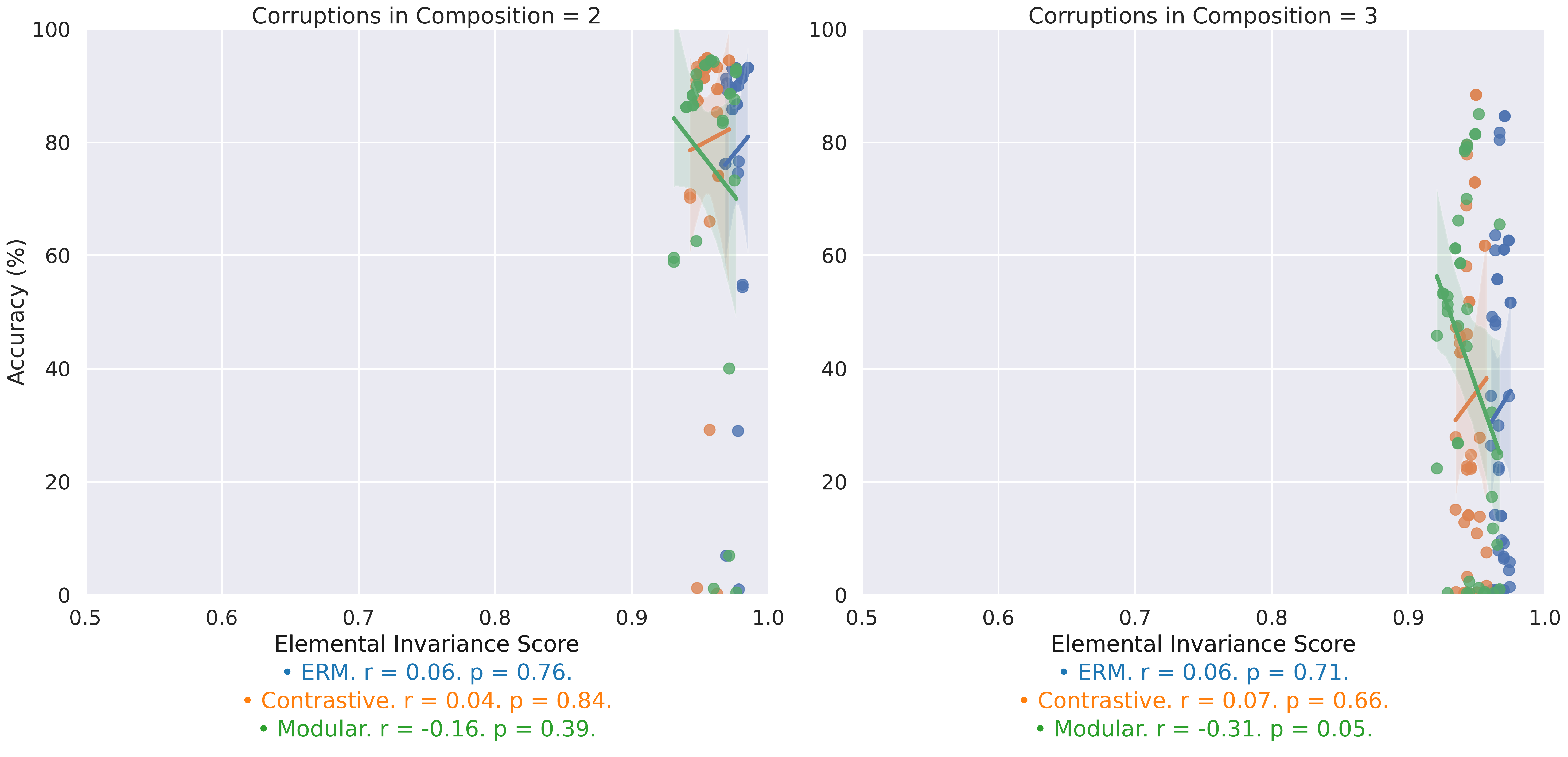} \\
    \end{subfigure}
    \begin{subfigure}[t]{0.9\textwidth}
        \centering
        \includegraphics[trim={0 60cm 0 0},clip,width=\linewidth]{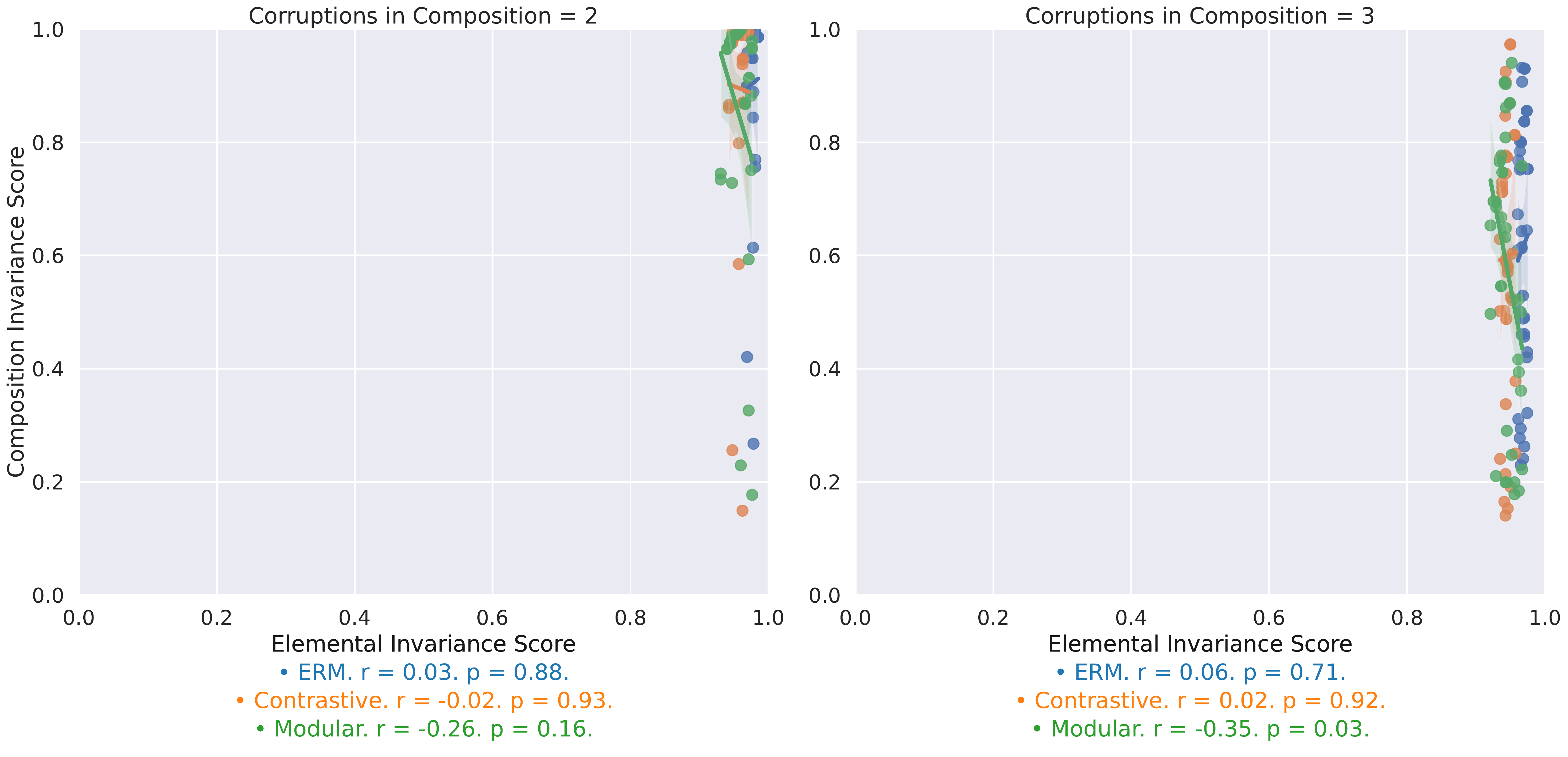} \\
    \end{subfigure}
    \begin{subfigure}[t]{0.9\textwidth}
        \centering
        \includegraphics[trim={0 60cm 0 0},clip,width=\linewidth]{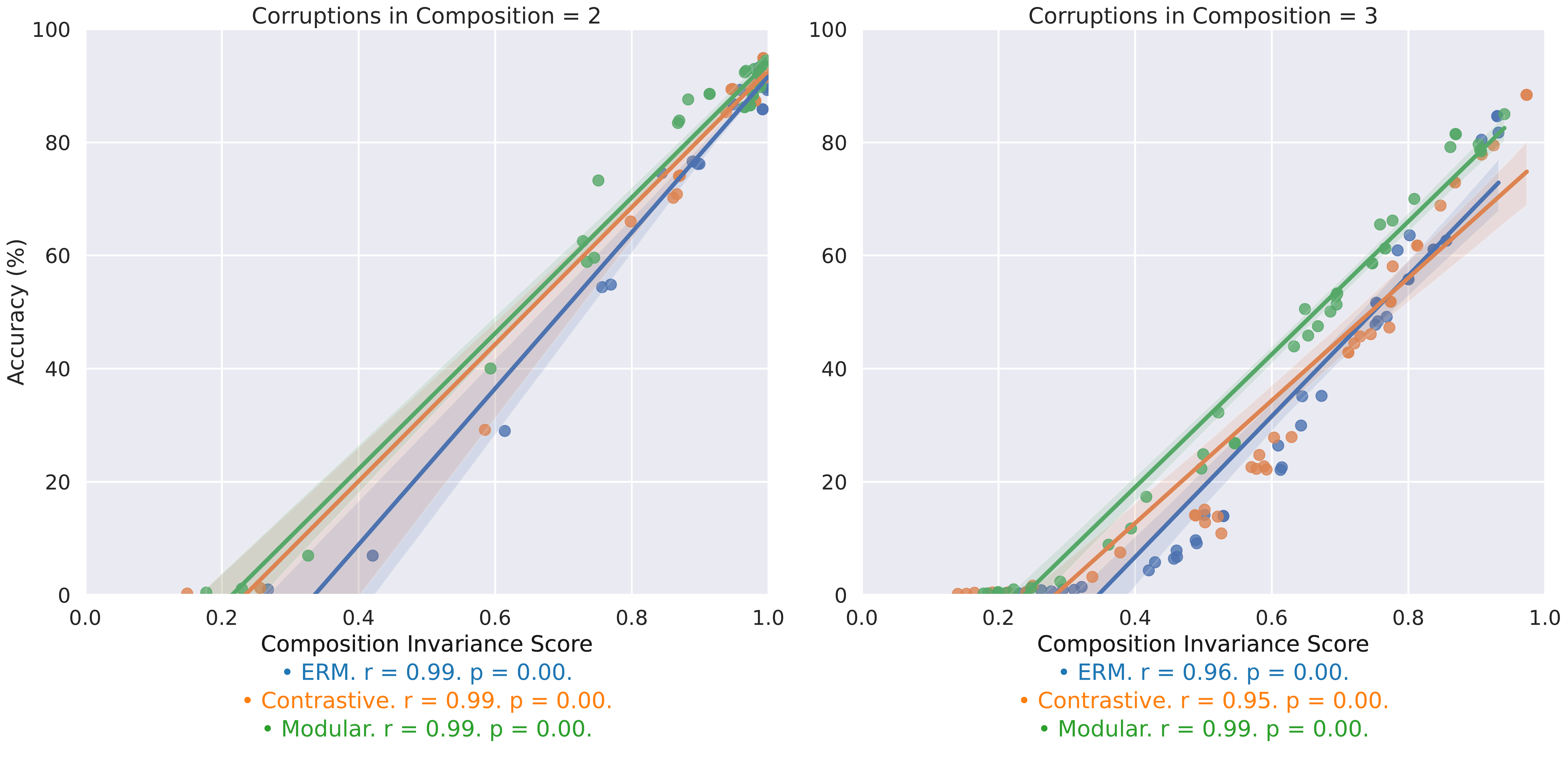} \\
    \end{subfigure}
    \vspace{-1.5mm}
    \caption{Correlating invariance scores with compositional robustness for \facescrub. These plots plot the same relationships as in Figure~\ref{fig:invariance-emnist}.}
    \label{fig:invariance-facescrub}
    \vspace{-6.5mm}
\end{figure}

Appendices~\ref{app:invariance}, \ref{app:seeds} and \ref{app:heat-maps} show a large number of plots over the following pages. This appendix contains further plots correlating invariance scores with compositional robustness. To begin we show the invariance summary plots for \cifar\ (Figure~\ref{fig:invariance-cifar}) and \facescrub\ (Figure~\ref{fig:invariance-facescrub}). These plots are the equivalent of Figure~\ref{fig:invariance-emnist} for \emnist\ from the main text. 

% \subsection{Invariance Plots for All Compositions}

\begin{figure}[tb]
    \centering
    \includegraphics[trim={0 0 0 0},clip,width=0.9\textwidth]{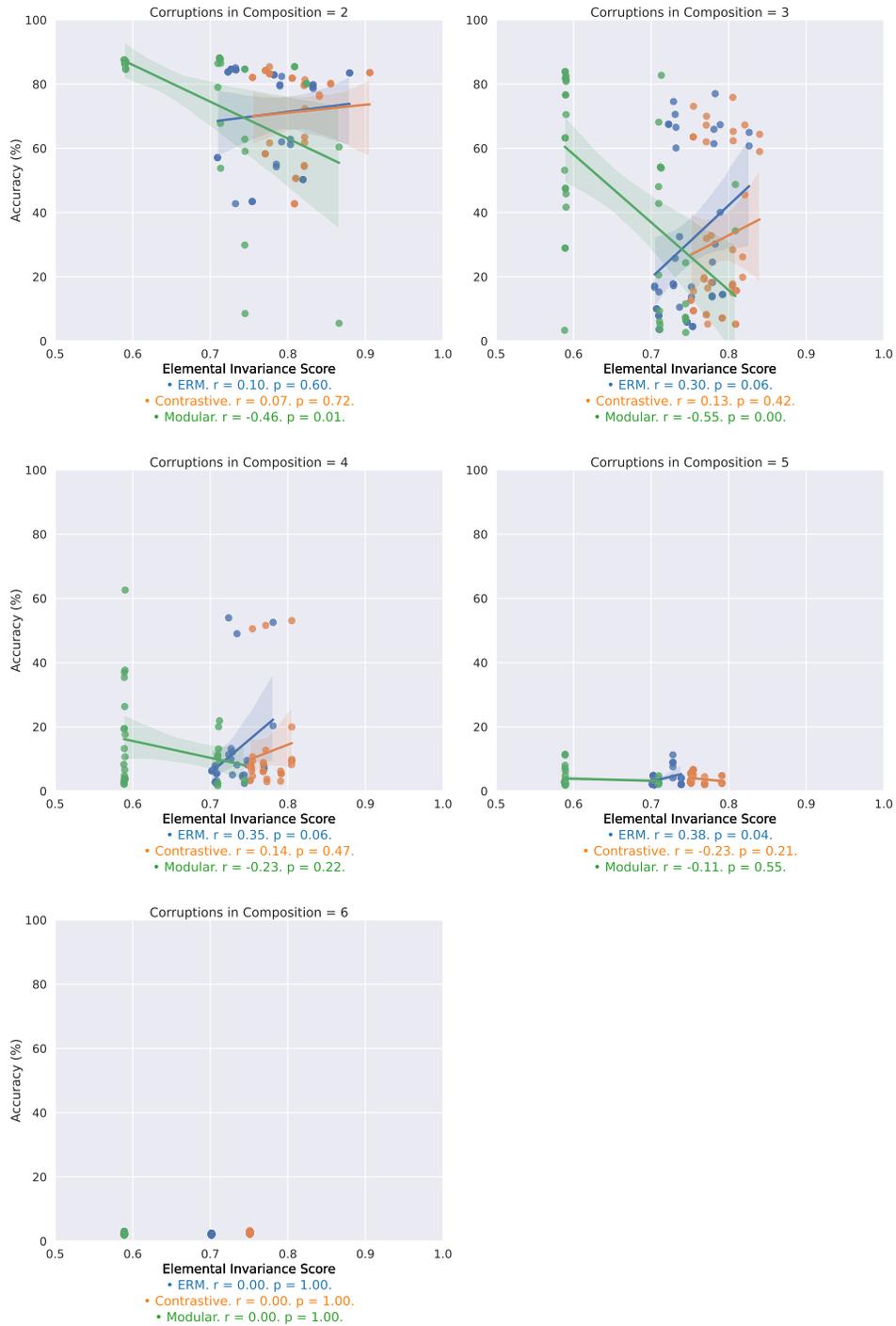}
    \caption{Correlating the elemental invariance score with compositional robustness for \emnist. These plots expand the first row of Figure~\ref{fig:invariance-emnist} to show all compositional test domains.}
    \label{fig:elem-acc-emnist}
    \vspace{-6.5mm}
\end{figure}

\begin{figure}[tb]
    \centering
    \includegraphics[trim={0 0 0 0},clip,width=0.9\textwidth]{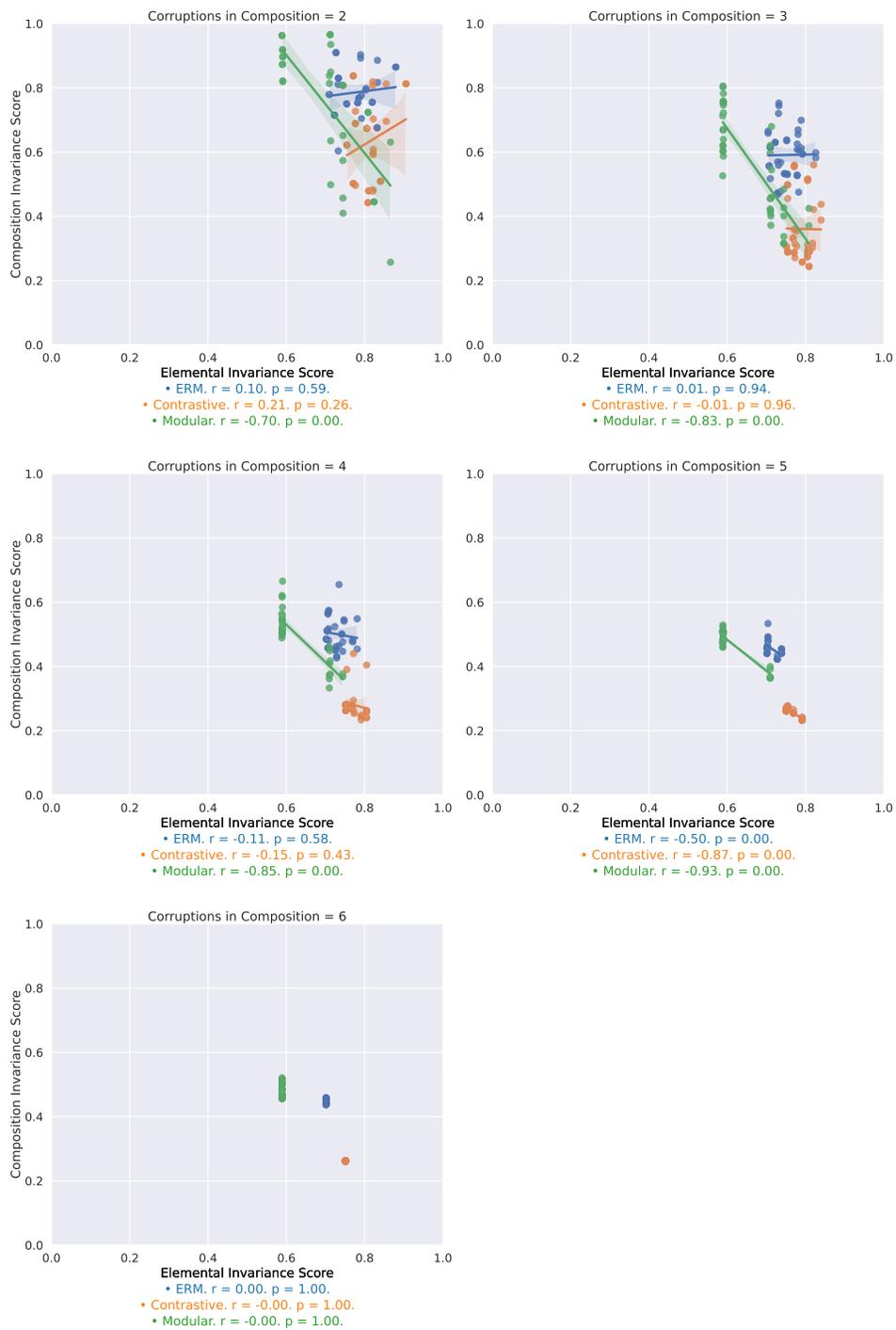}
    \caption{Correlating the elemental invariance score with the composition invariance score for \emnist. These plots expand the second row of Figure~\ref{fig:invariance-emnist} to show all compositional test domains.}
    \label{fig:elem-comp-emnist}
    \vspace{-6.5mm}
\end{figure}

\begin{figure}[tb]
    \centering
    \includegraphics[trim={0 0 0 0},clip,width=0.9\textwidth]{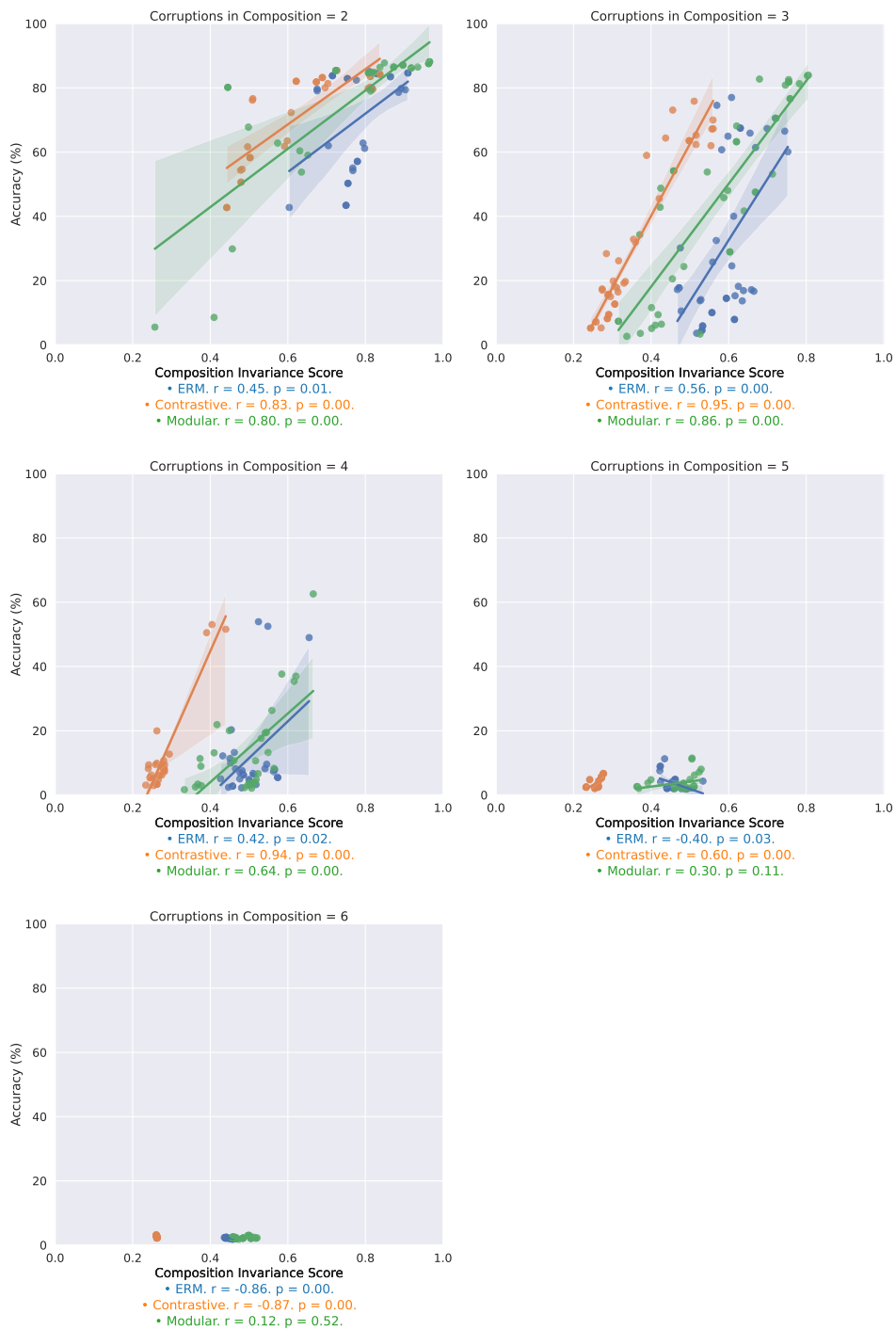}
    \caption{Correlating the composition invariance score with compositional robustness for \emnist. These plots expand the third row of Figure~\ref{fig:invariance-emnist} to show all compositional test domains.}
    \label{fig:comp-acc-emnist}
    \vspace{-6.5mm}
\end{figure}

\begin{figure}[tb]
    \centering
    \includegraphics[trim={0 0 0 0},clip,width=0.9\textwidth]{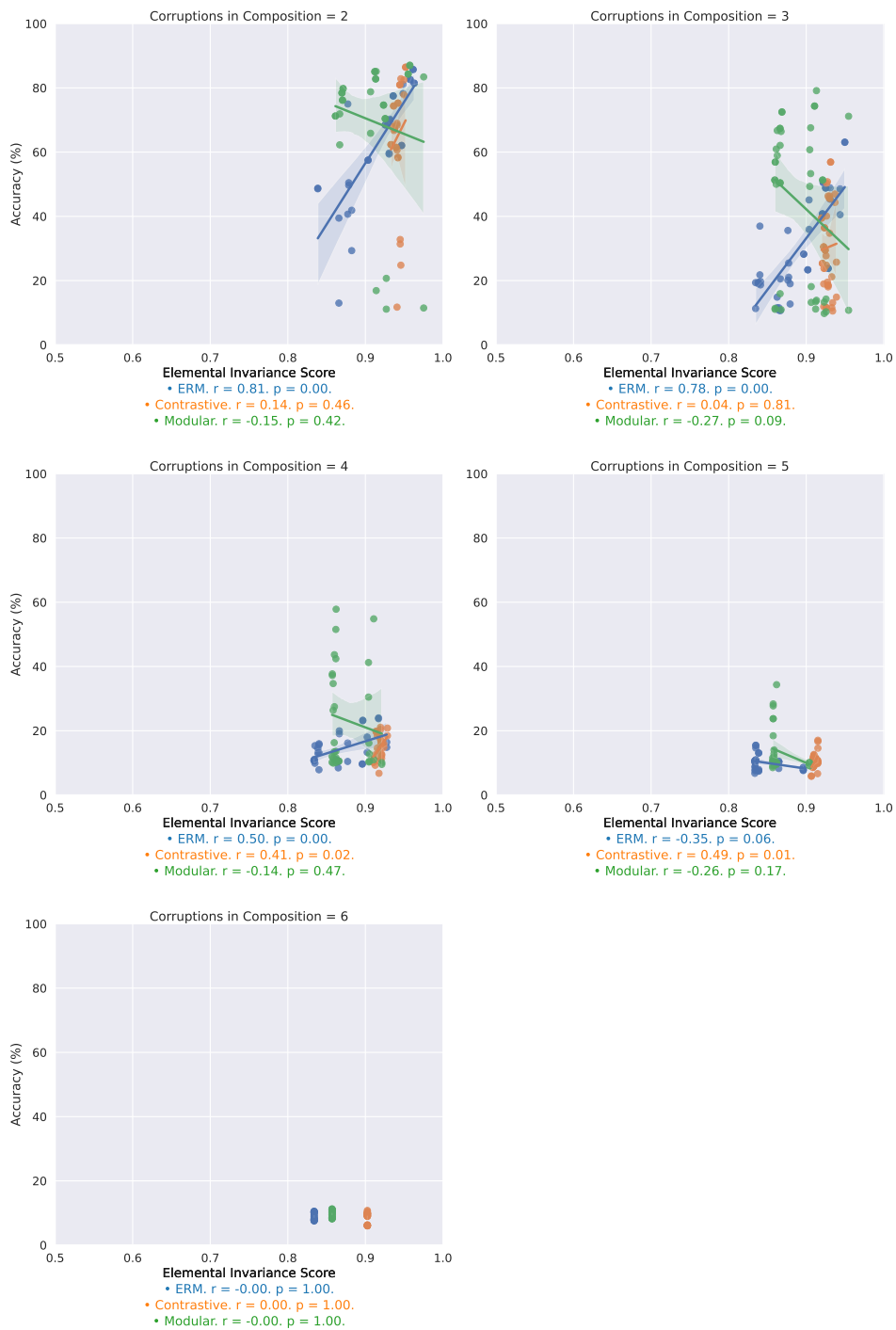}
    \caption{Correlating the elemental invariance score with compositional robustness for \cifar. These plots expand the first row of Figure~\ref{fig:invariance-cifar} to show all compositional test domains.}
    \label{fig:elem-acc-cifar}
    \vspace{-6.5mm}
\end{figure}

\begin{figure}[tb]
    \centering
    \includegraphics[trim={0 0 0 0},clip,width=0.9\textwidth]{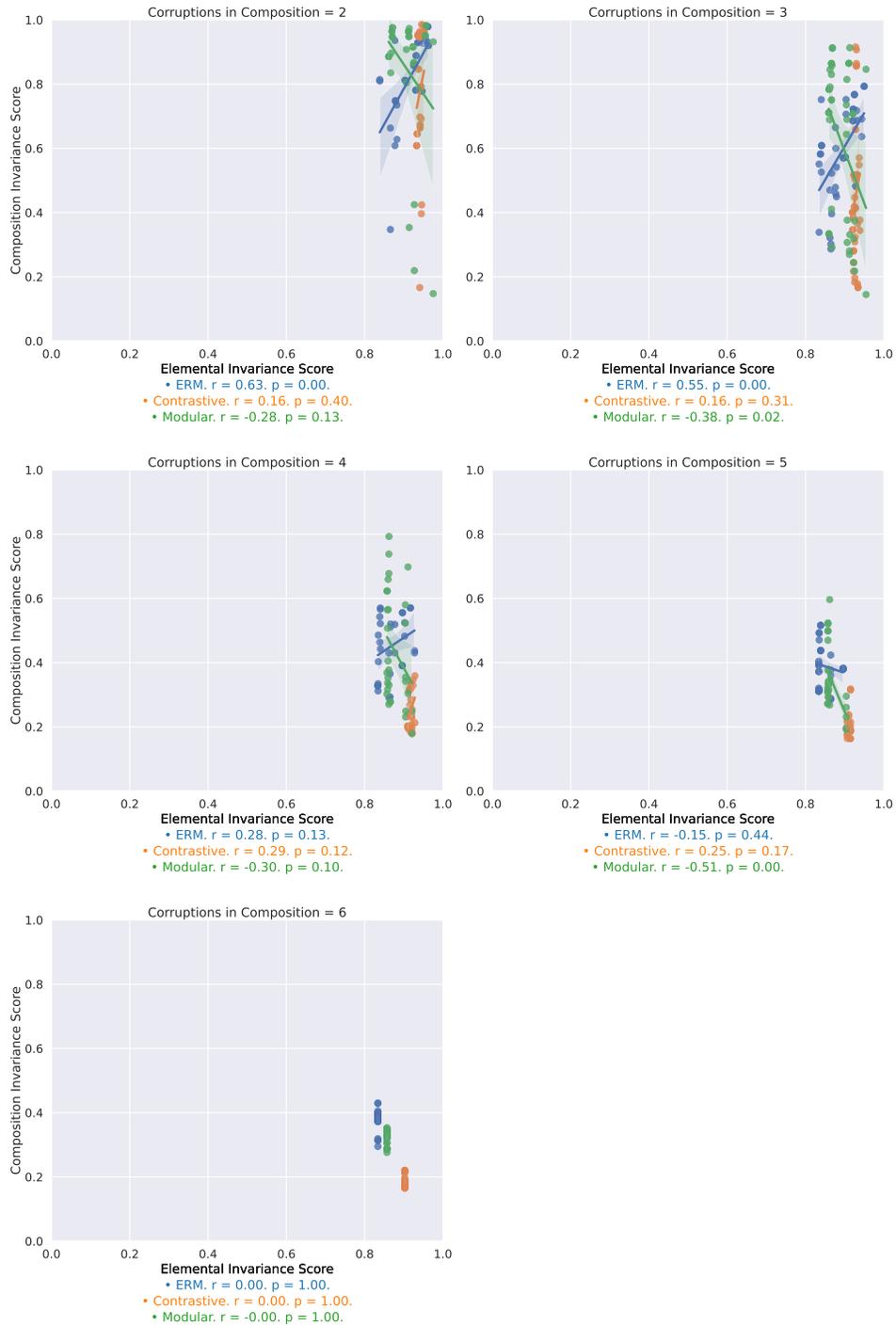}
    \caption{Correlating the elemental invariance score with the composition invariance score for \cifar. These plots expand the second row of Figure~\ref{fig:invariance-cifar} to show all compositional test domains.}
    \label{fig:elem-comp-cifar}
    \vspace{-6.5mm}
\end{figure}

\begin{figure}[tb]
    \centering
    \includegraphics[trim={0 0 0 0},clip,width=0.9\textwidth]{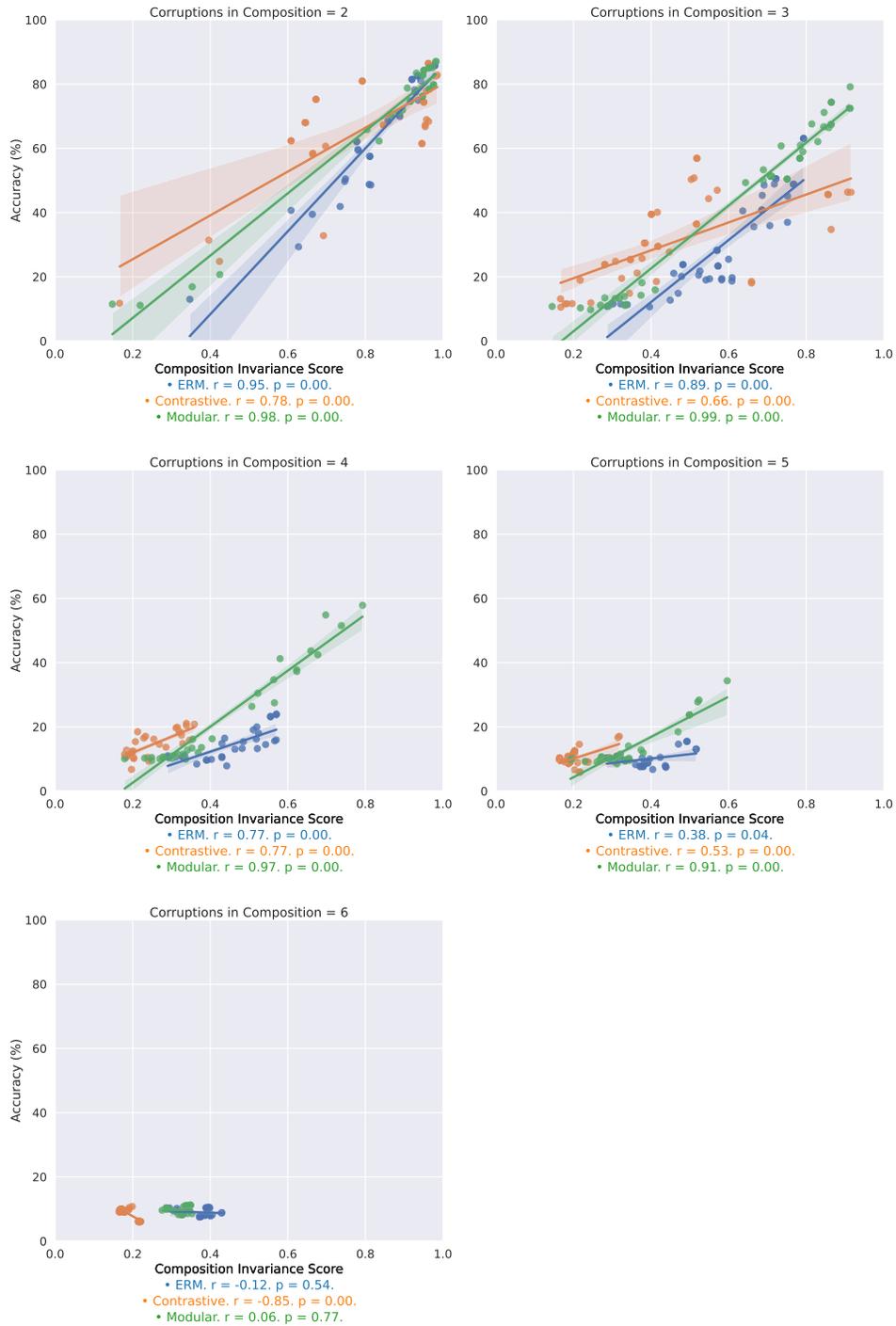}
    \caption{Correlating the composition invariance score with compositional robustness for \cifar. These plots expand the third row of Figure~\ref{fig:invariance-cifar} to show all compositional test domains.}
    \label{fig:comp-acc-cifar}
    \vspace{-6.5mm}
\end{figure}

\begin{figure}[tb]
    \centering
    \includegraphics[trim={0 0 0 0},clip,width=0.9\textwidth]{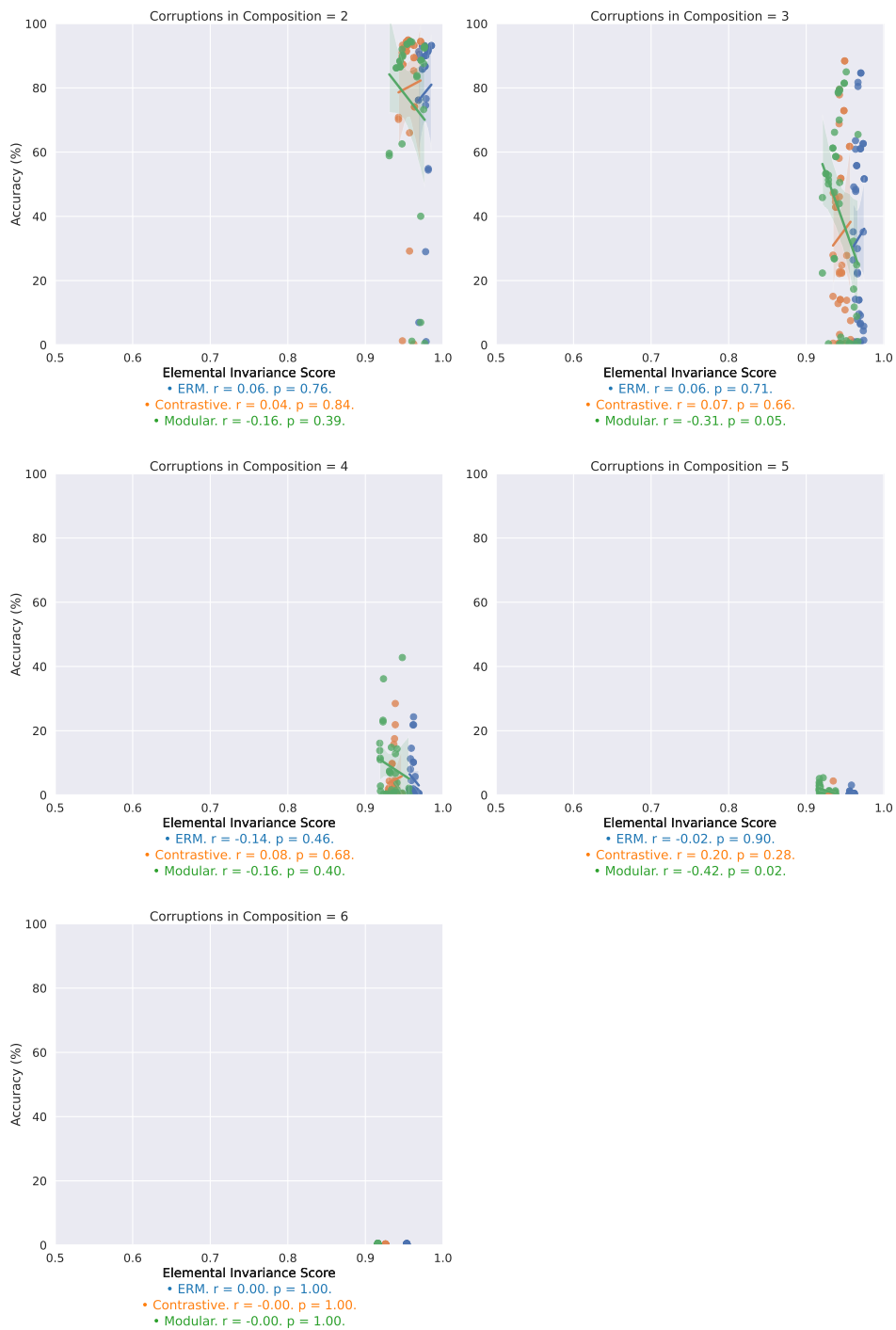}
    \caption{Correlating the elemental invariance score with compositional robustness for \facescrub. These plots expand the first row of Figure~\ref{fig:invariance-facescrub} to show all compositional test domains.}
    \label{fig:elem-acc-facescrub}
    \vspace{-6.5mm}
\end{figure}

\begin{figure}[tb]
    \centering
    \includegraphics[trim={0 0 0 0},clip,width=0.9\textwidth]{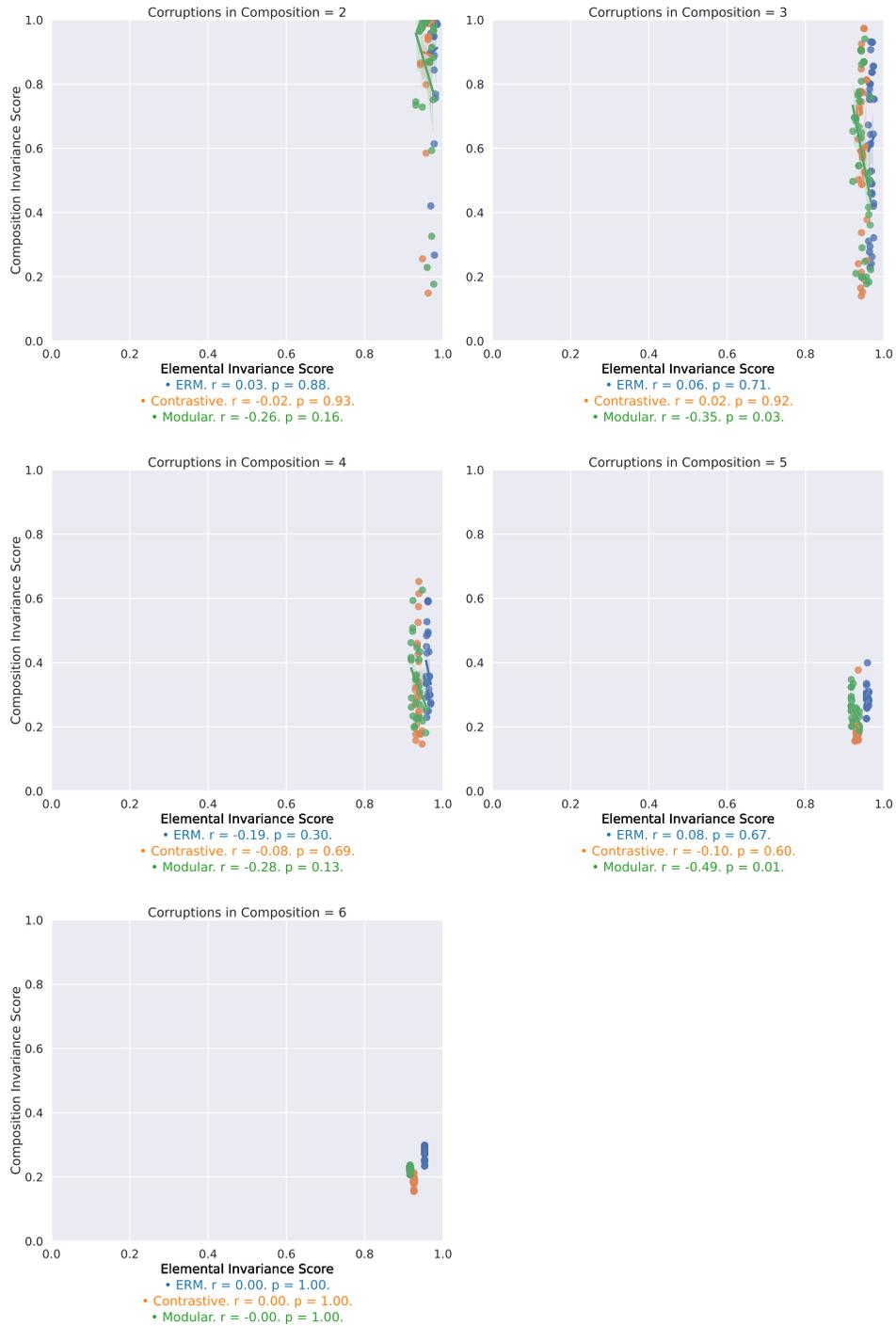}
    \caption{Correlating the elemental invariance score with the composition invariance score for \facescrub. These plots expand the second row of Figure~\ref{fig:invariance-facescrub} to show all compositional test domains.}
    \label{fig:elem-comp-facescrub}
    \vspace{-6.5mm}
\end{figure}

\begin{figure}[tb]
    \centering
    \includegraphics[trim={0 0 0 0},clip,width=0.9\textwidth]{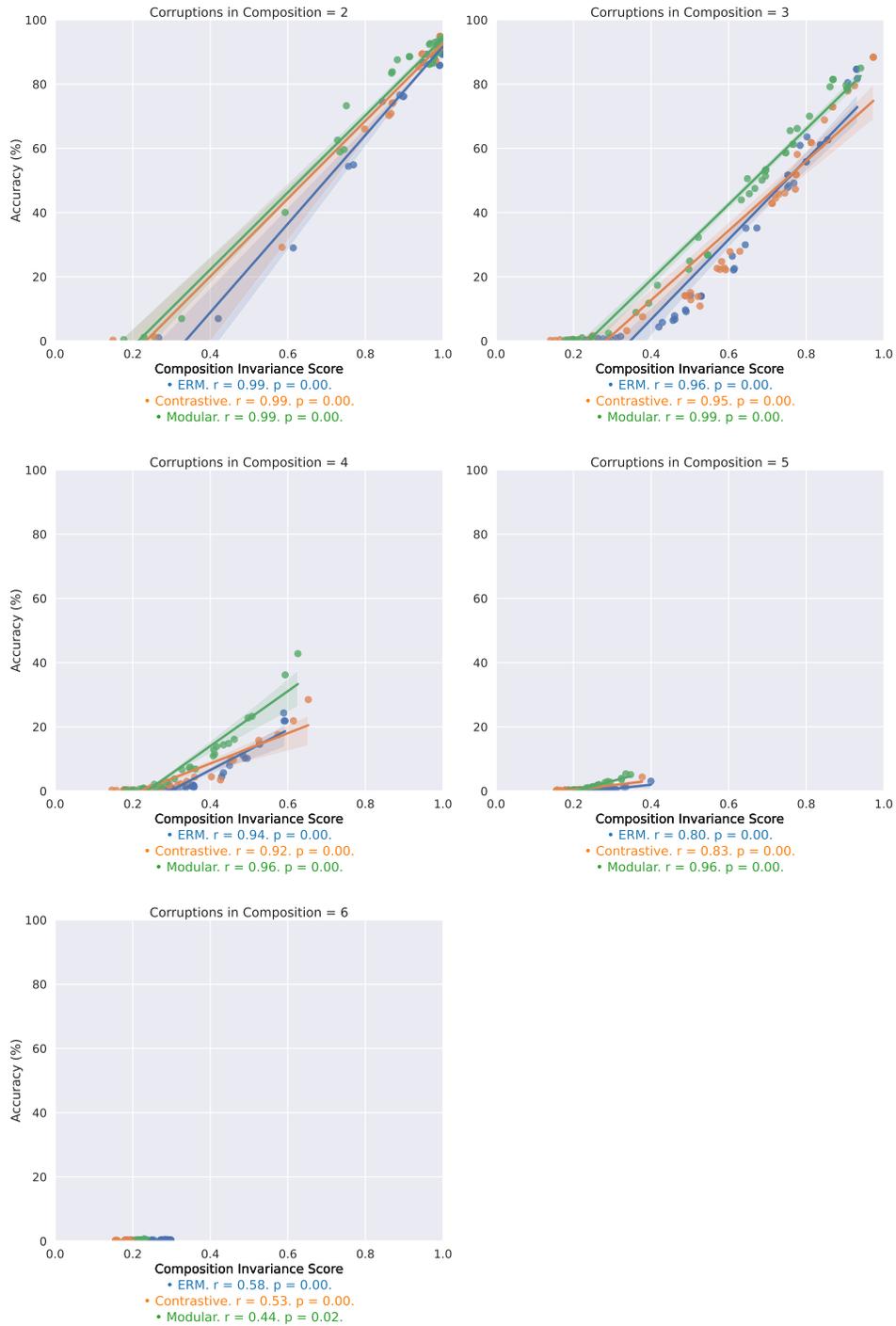}
    \caption{Correlating the composition invariance score with compositional robustness for \facescrub. These plots expand the third row of Figure~\ref{fig:invariance-facescrub} to show all compositional test domains.}
    \label{fig:comp-acc-facescrub}
    \vspace{-6.5mm}
\end{figure}

Following this, in Figures~\ref{fig:elem-acc-emnist}-\ref{fig:comp-acc-facescrub}, we show the invariance summary plots (Figures~\ref{fig:invariance-emnist}, \ref{fig:invariance-cifar}, \ref{fig:invariance-facescrub}) expanded over all compositional test domains. That is, these plots include the plots for compositions containing more than three corruptions. For compositions of more than three corruptions accuracy is often low, making it challenging to uncover meaningful trends.

% \subsection{Invariance Plots for All Compositions}

% EMNIST

% CIFAR

%FACESCRUB

\section{Variance Over Seeds}
\label{app:seeds}

\begin{figure}[tb]
    \centering
    % \vspace{-6mm}
    \begin{subfigure}[t]{0.32\textwidth}
        \centering
        \includegraphics[width=\linewidth]{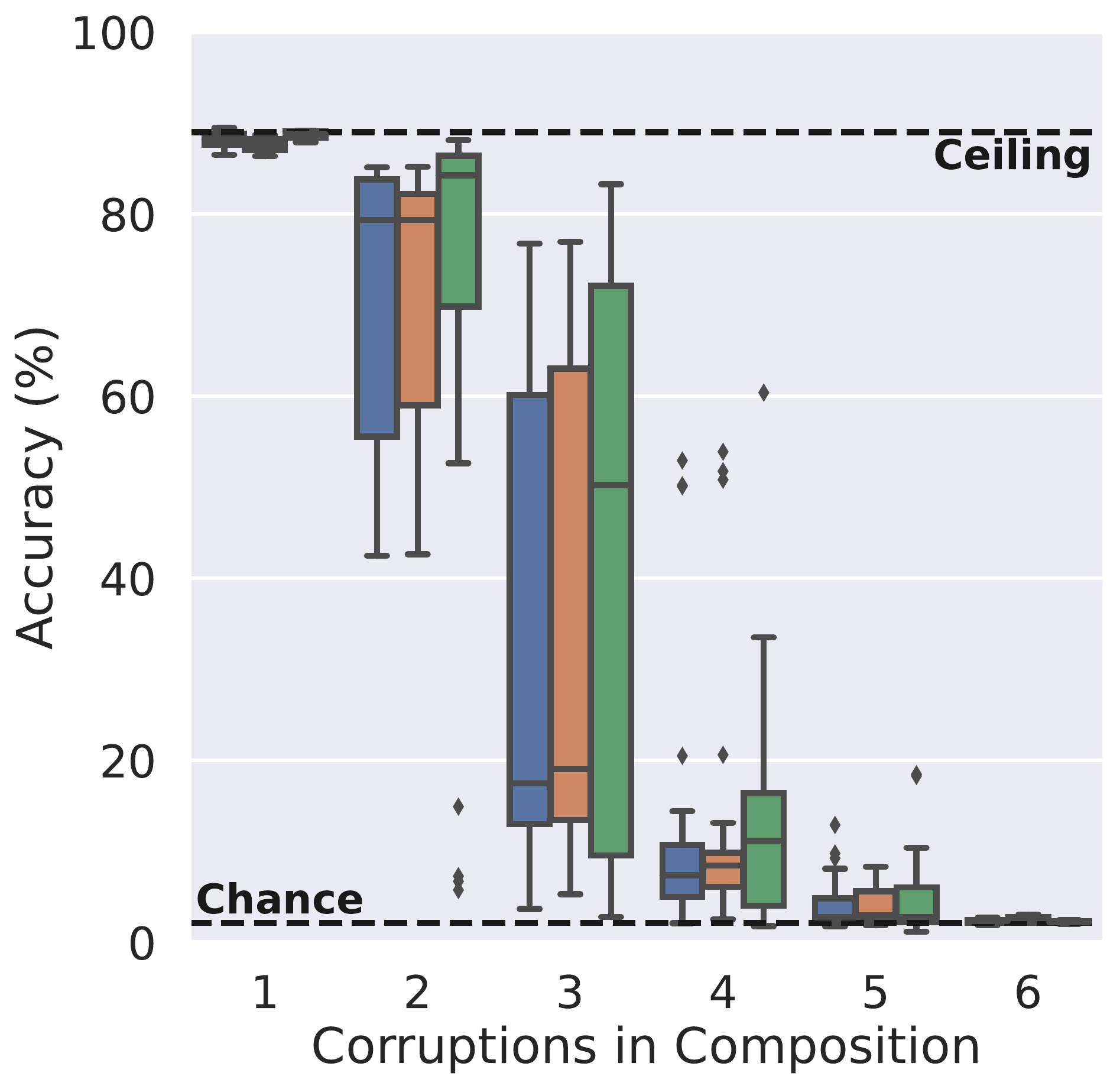} \\
        \caption{EMNIST}
        \label{fig:seed2-comp-em}
    \end{subfigure}%
    \hfill
    \begin{subfigure}[t]{0.32\textwidth}
        \centering
        \includegraphics[width=\linewidth]{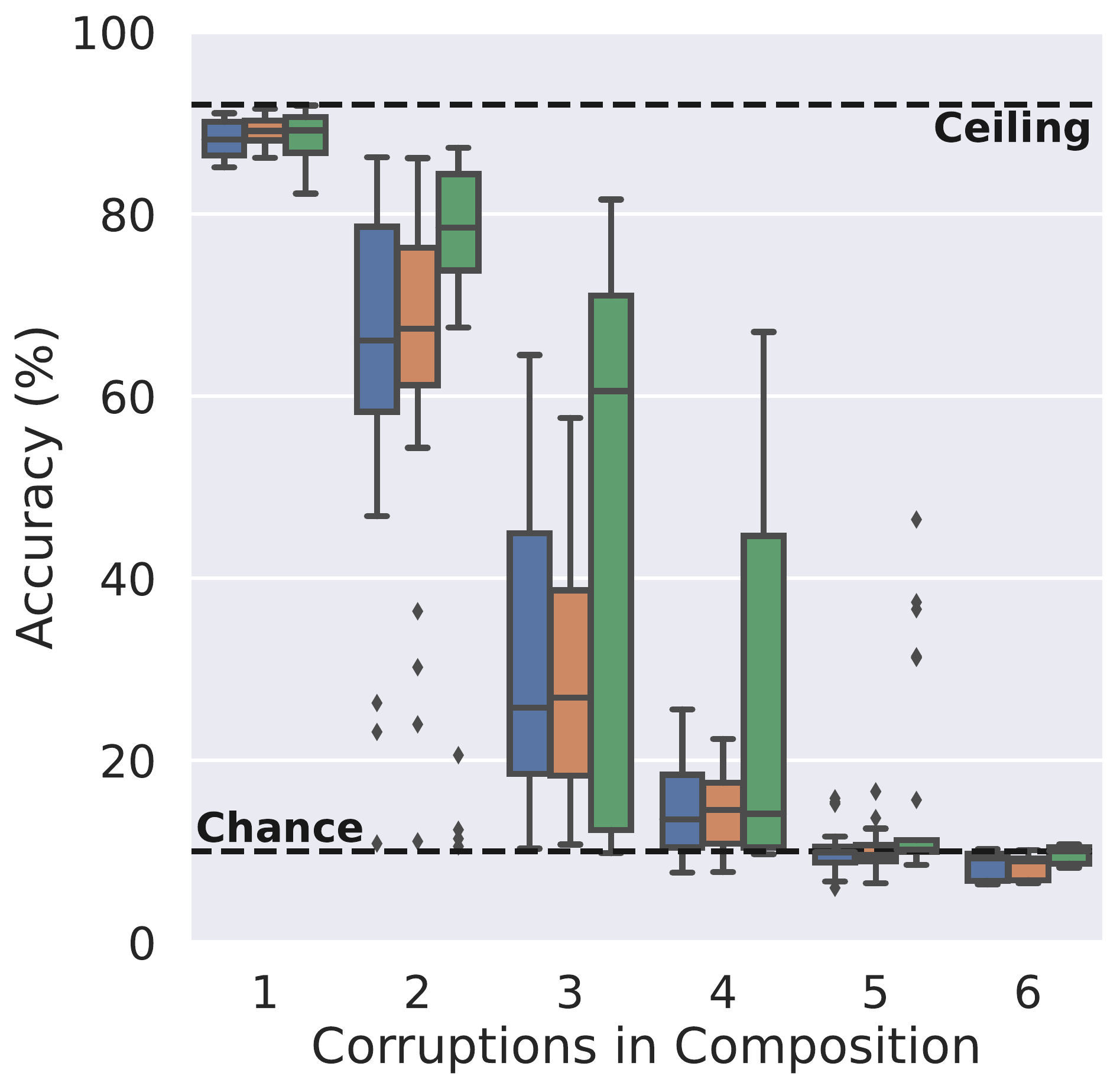} \\
        \caption{CIFAR-10}
        \label{fig:seed2-comp-cf}
    \end{subfigure}%
    \hfill
    \begin{subfigure}[t]{0.32\textwidth}
        \centering
        \includegraphics[width=\linewidth]{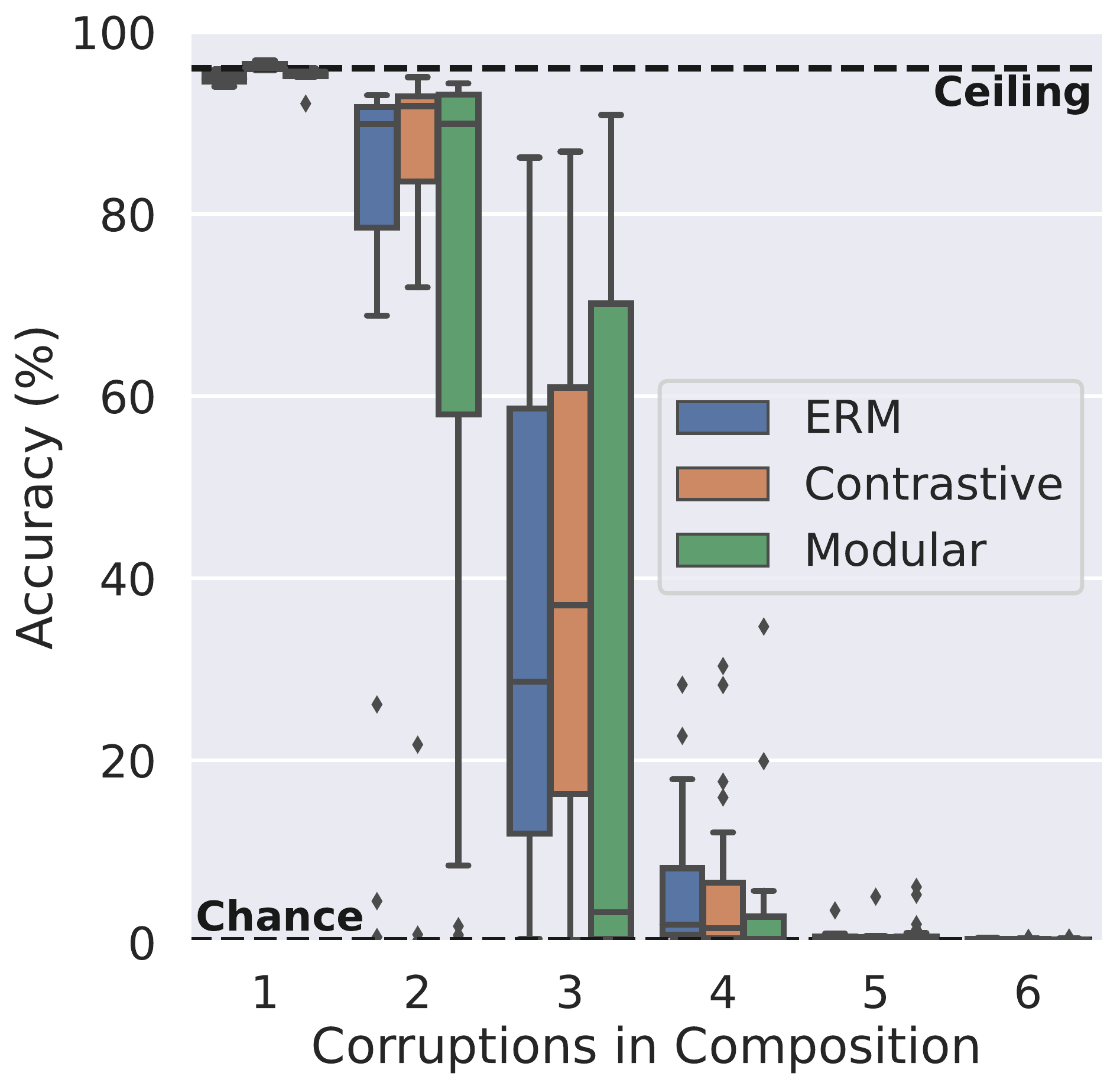} \\
        \caption{FACESCRUB}
        \label{fig:seed2-comp-fs}
    \end{subfigure}
    \vspace{-1.5mm}
    \caption{Evaluating compositional robustness on different datasets (second random seed). This figure is the same as Figure~\ref{fig:comp} with a different random seeding.}
    \label{fig:seed2-comp}
    \vspace{-4.5mm}
\end{figure}

\begin{figure}[tb]
    \centering
    \begin{subfigure}[t]{0.9\textwidth}
        \centering
        \includegraphics[trim={0 60cm 0 0},clip,width=\linewidth]{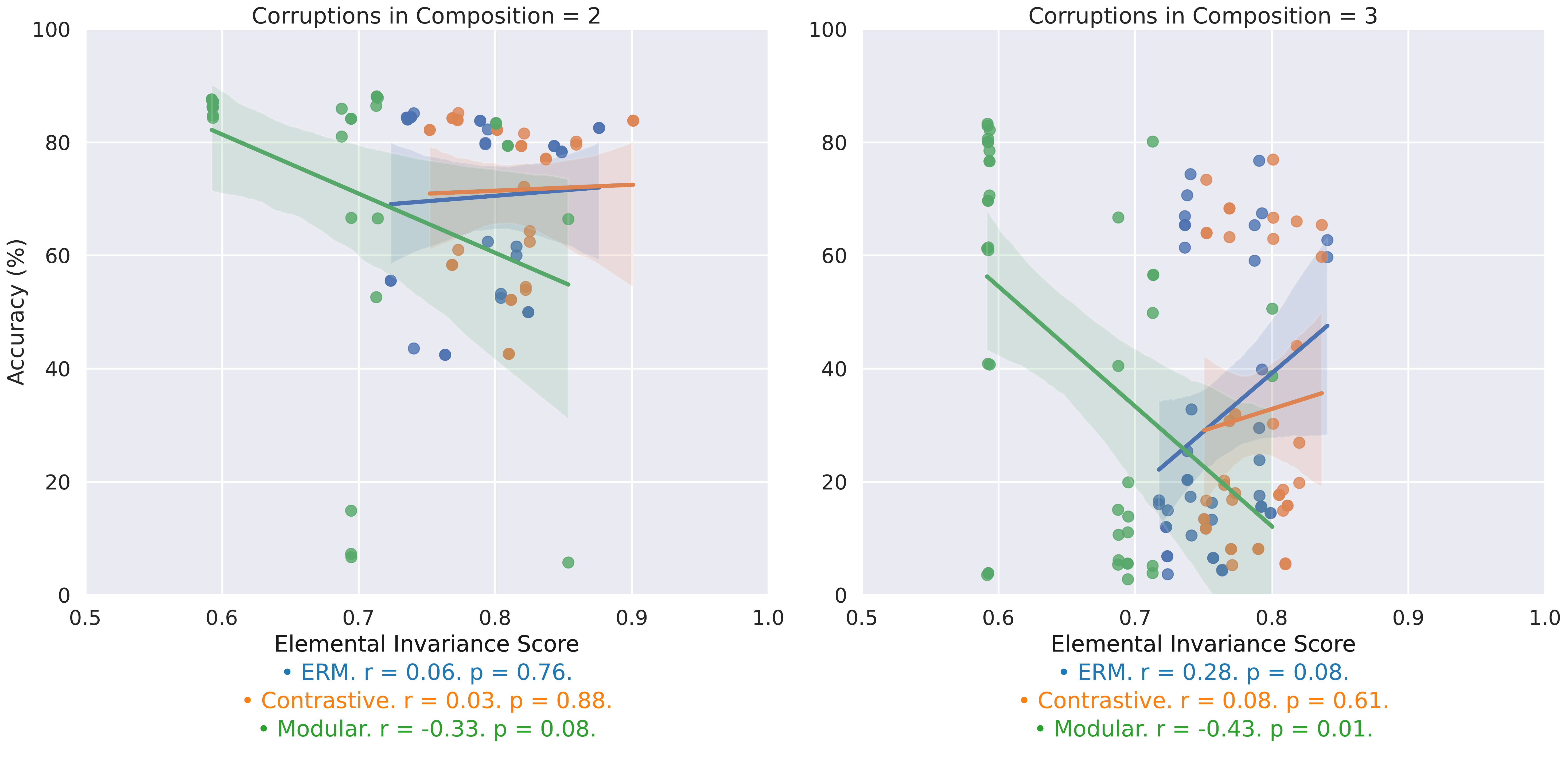} \\
    \end{subfigure}
    \begin{subfigure}[t]{0.9\textwidth}
        \centering
        \includegraphics[trim={0 60cm 0 0},clip,width=\linewidth]{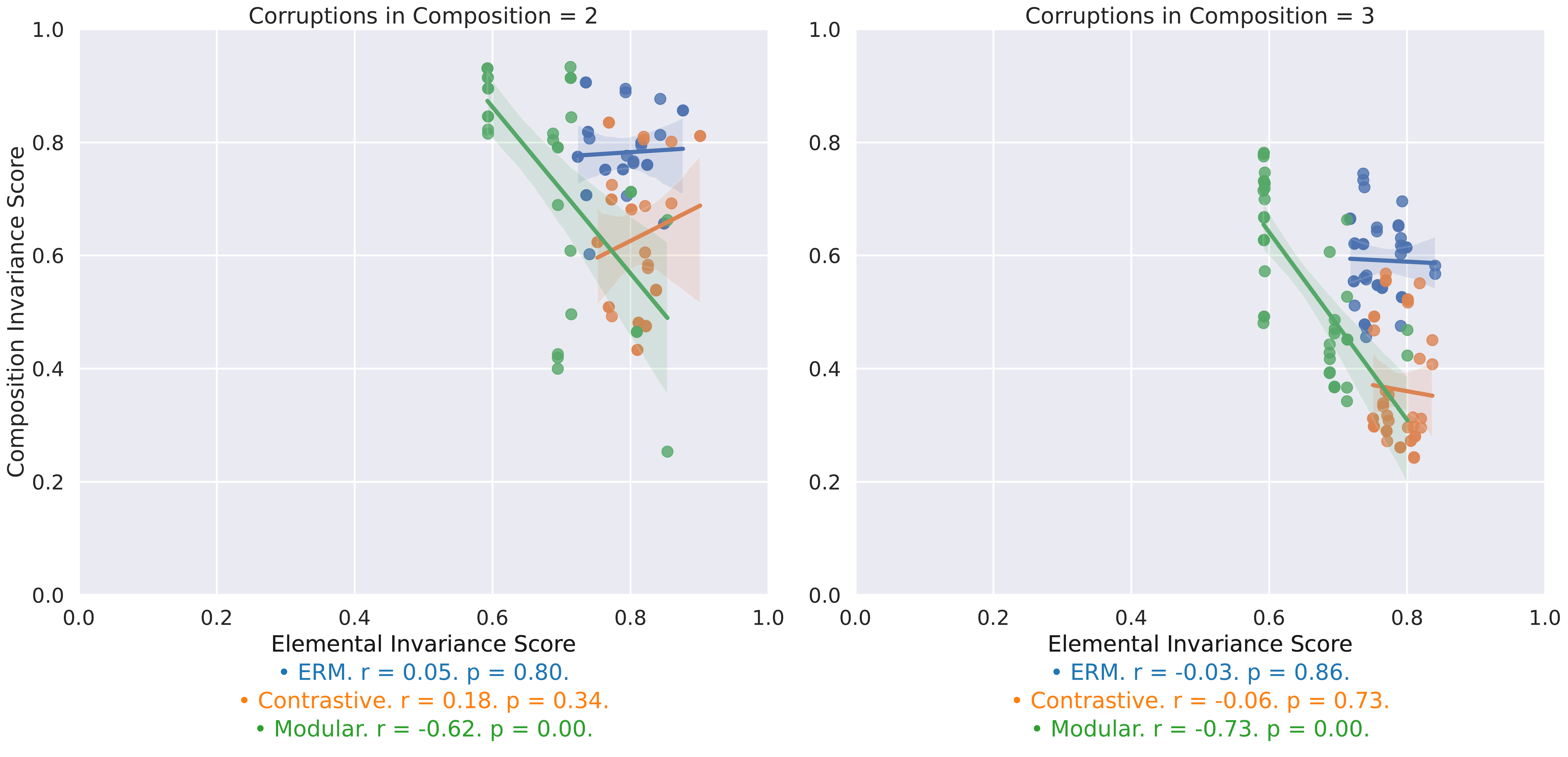} \\
    \end{subfigure}
    \begin{subfigure}[t]{0.9\textwidth}
        \centering
        \includegraphics[trim={0 60cm 0 0},clip,width=\linewidth]{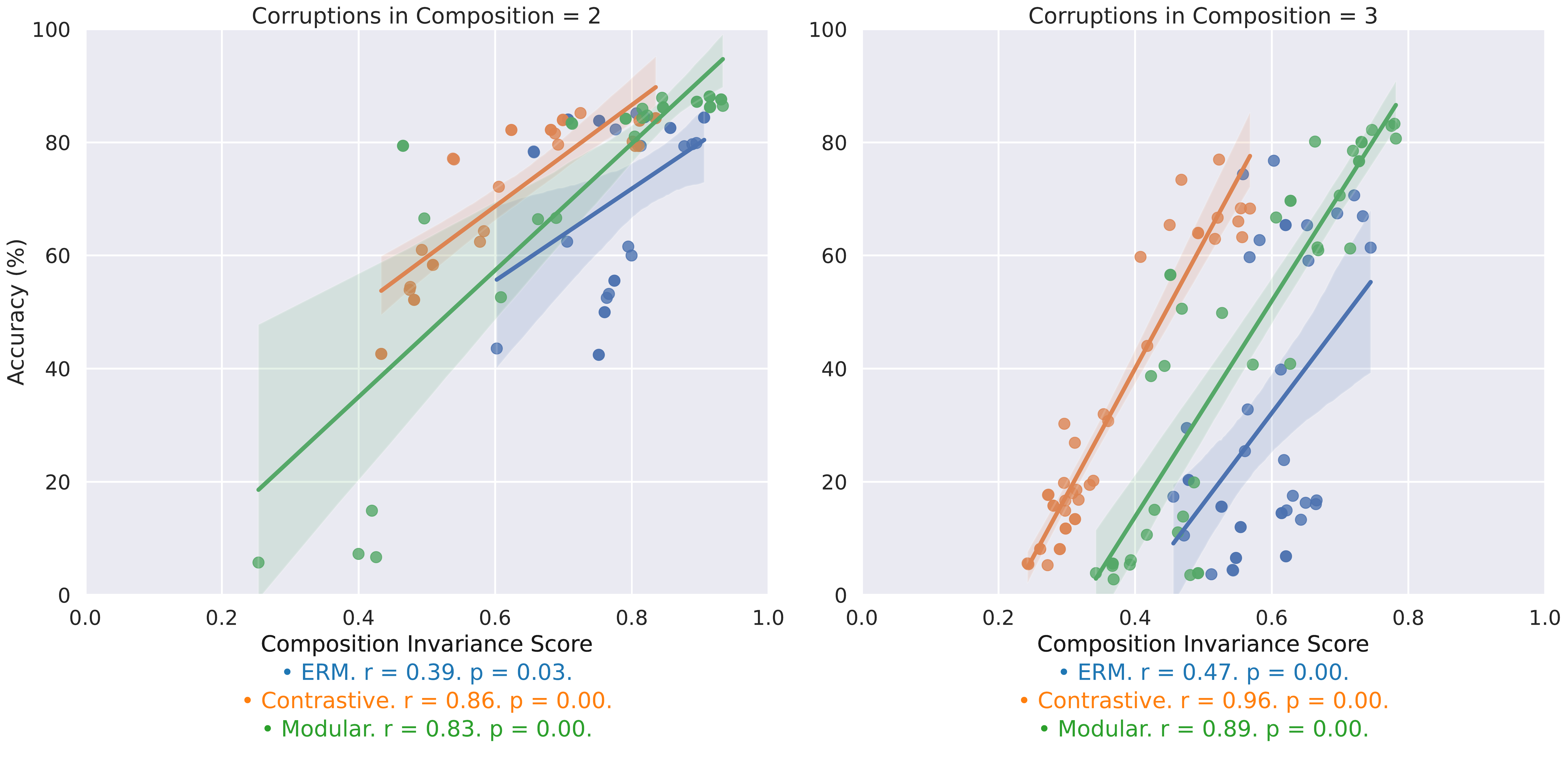} \\
    \end{subfigure}
    \vspace{-1.5mm}
    \caption{Correlating invariance scores with compositional robustness for \emnist\ (second random seed). This is the same as Figure~\ref{fig:invariance-emnist} with a different random seeding.}
    \label{fig:seed2-invariance-emnist}
    \vspace{-6.5mm}
\end{figure}

\begin{figure}[tb]
    \centering
    \begin{subfigure}[t]{0.9\textwidth}
        \centering
        \includegraphics[trim={0 60cm 0 0},clip,width=\linewidth]{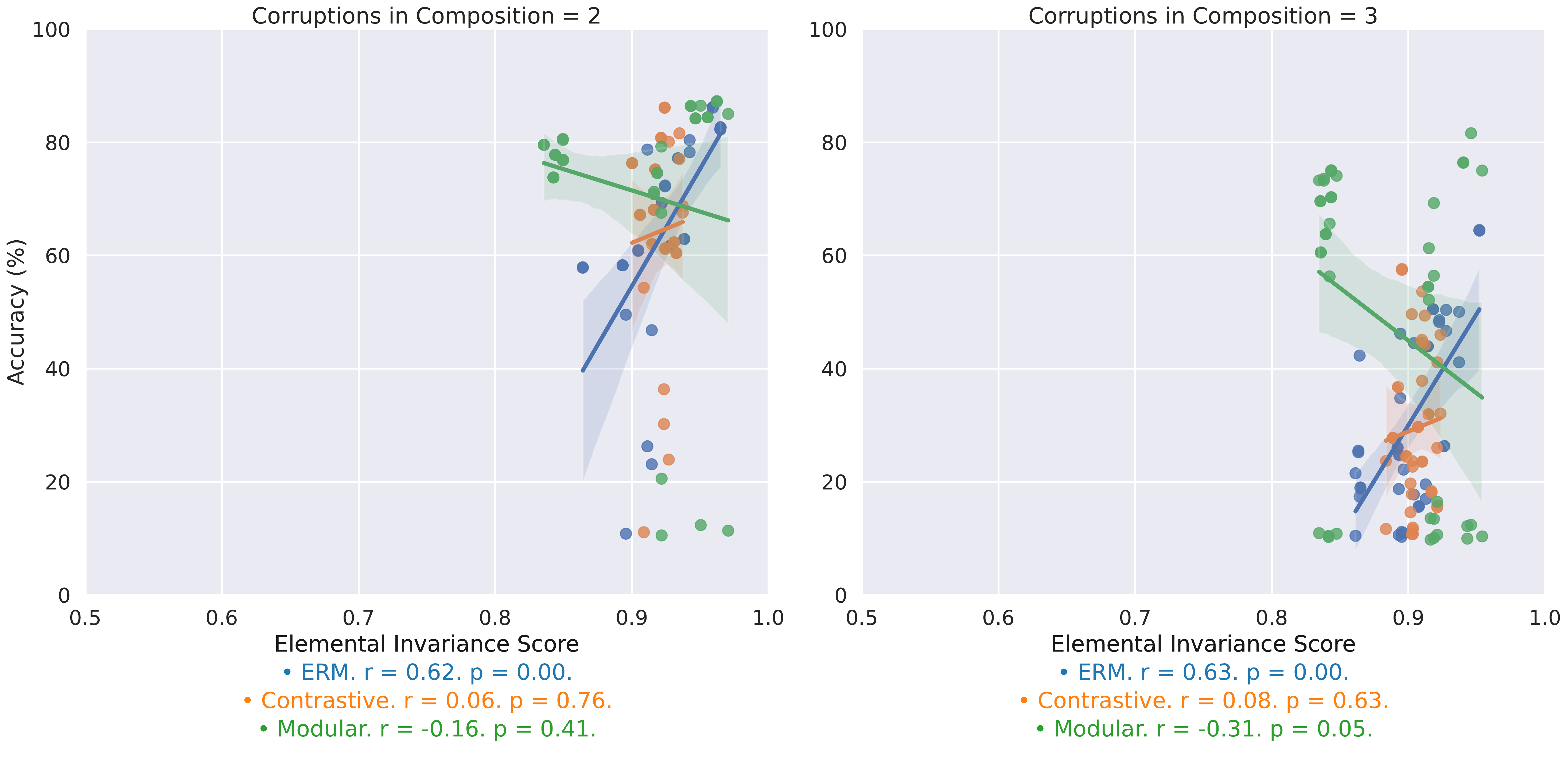} \\
    \end{subfigure}
    \begin{subfigure}[t]{0.9\textwidth}
        \centering
        \includegraphics[trim={0 60cm 0 0},clip,width=\linewidth]{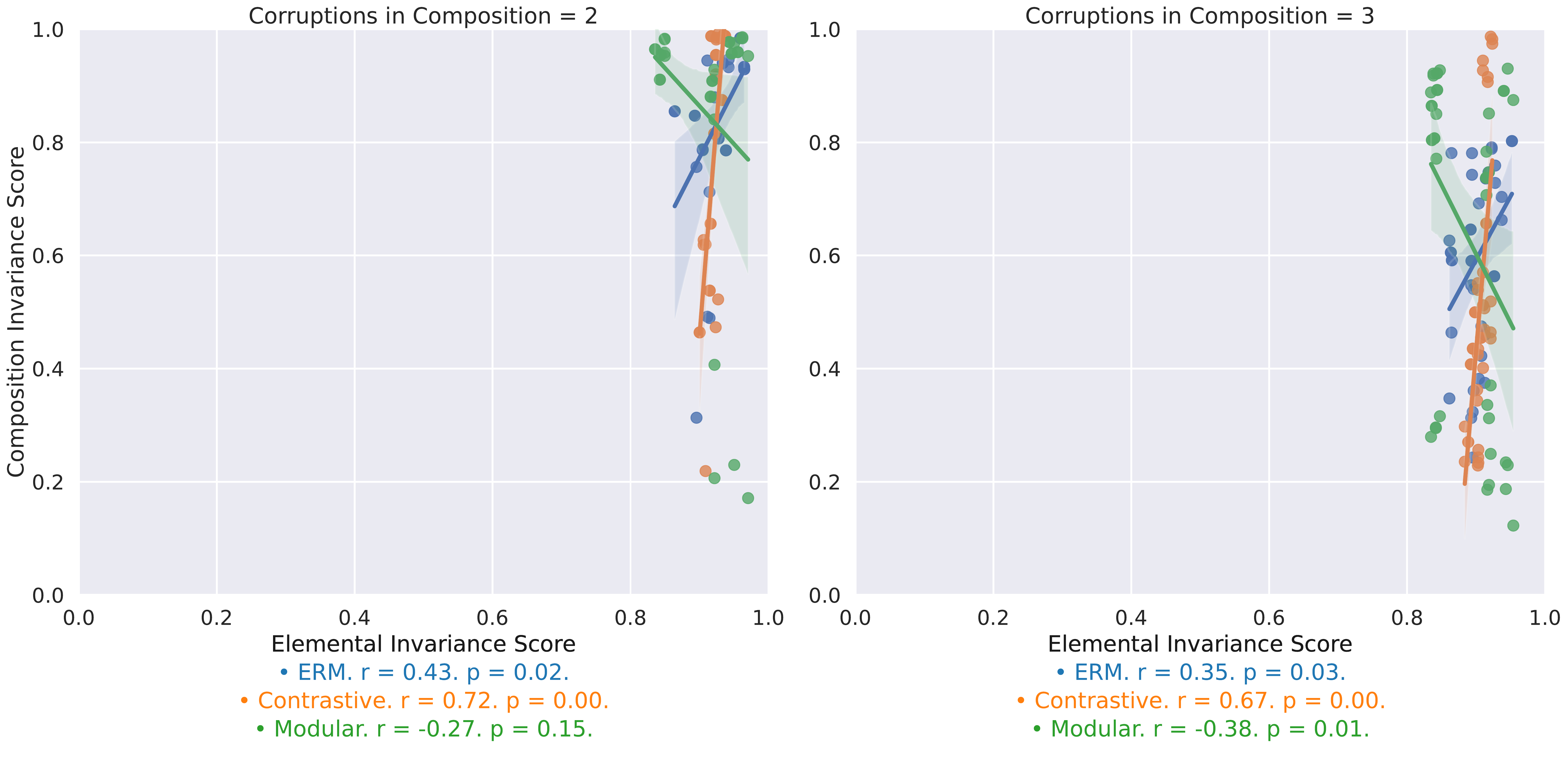} \\
    \end{subfigure}
    \begin{subfigure}[t]{0.9\textwidth}
        \centering
        \includegraphics[trim={0 60cm 0 0},clip,width=\linewidth]{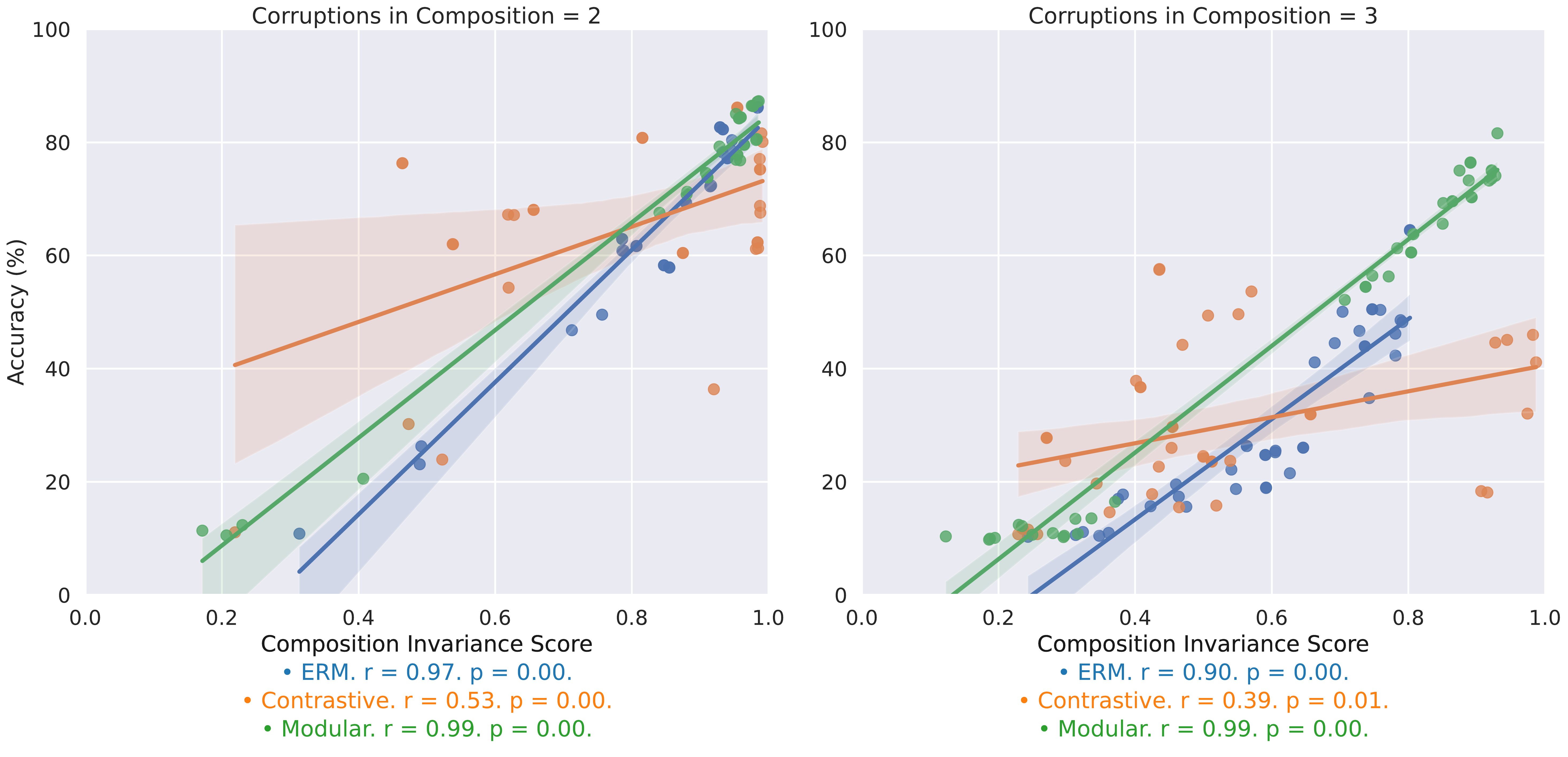} \\
    \end{subfigure}
    \vspace{-1.5mm}
    \caption{Correlating invariance scores with compositional robustness for \cifar\ (second random seed). This is the same as Figure~\ref{fig:invariance-cifar} with a different random seeding.}
    \label{fig:seed2-invariance-cifar}
    \vspace{-6.5mm}
\end{figure}

\begin{figure}[tb]
    \centering
    \begin{subfigure}[t]{0.9\textwidth}
        \centering
        \includegraphics[trim={0 60cm 0 0},clip,width=\linewidth]{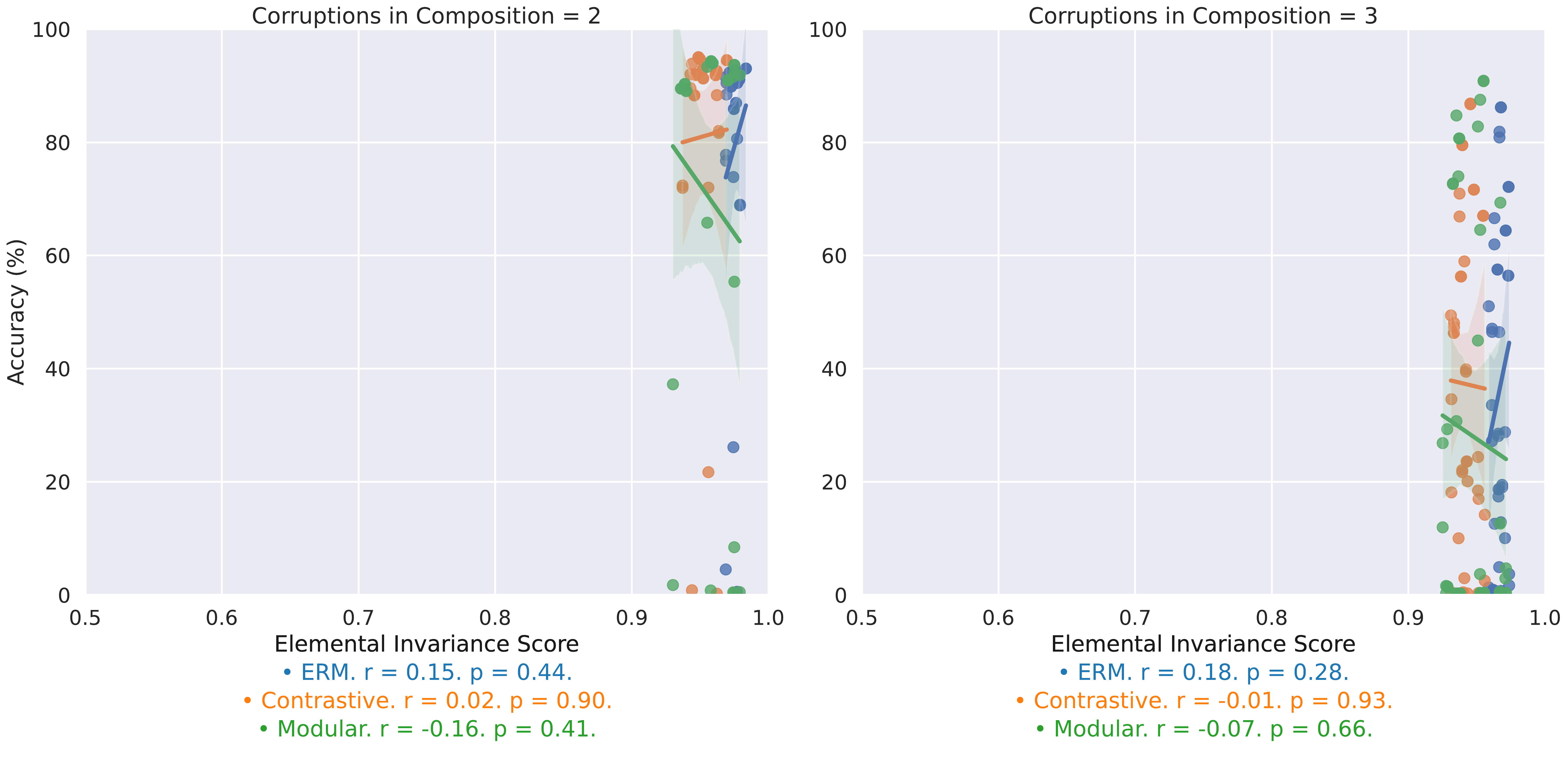} \\
    \end{subfigure}
    \begin{subfigure}[t]{0.9\textwidth}
        \centering
        \includegraphics[trim={0 60cm 0 0},clip,width=\linewidth]{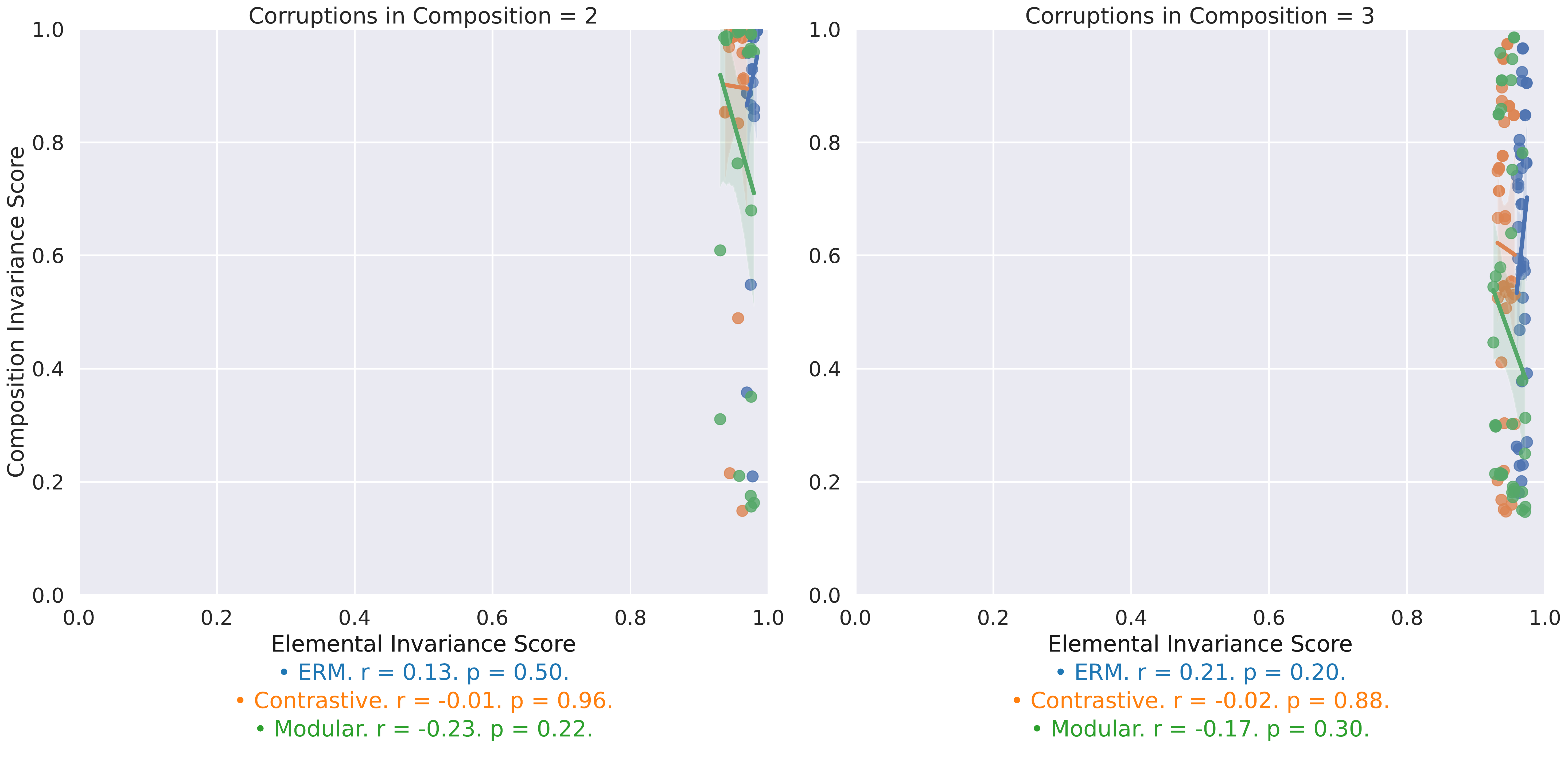} \\
    \end{subfigure}
    \begin{subfigure}[t]{0.9\textwidth}
        \centering
        \includegraphics[trim={0 60cm 0 0},clip,width=\linewidth]{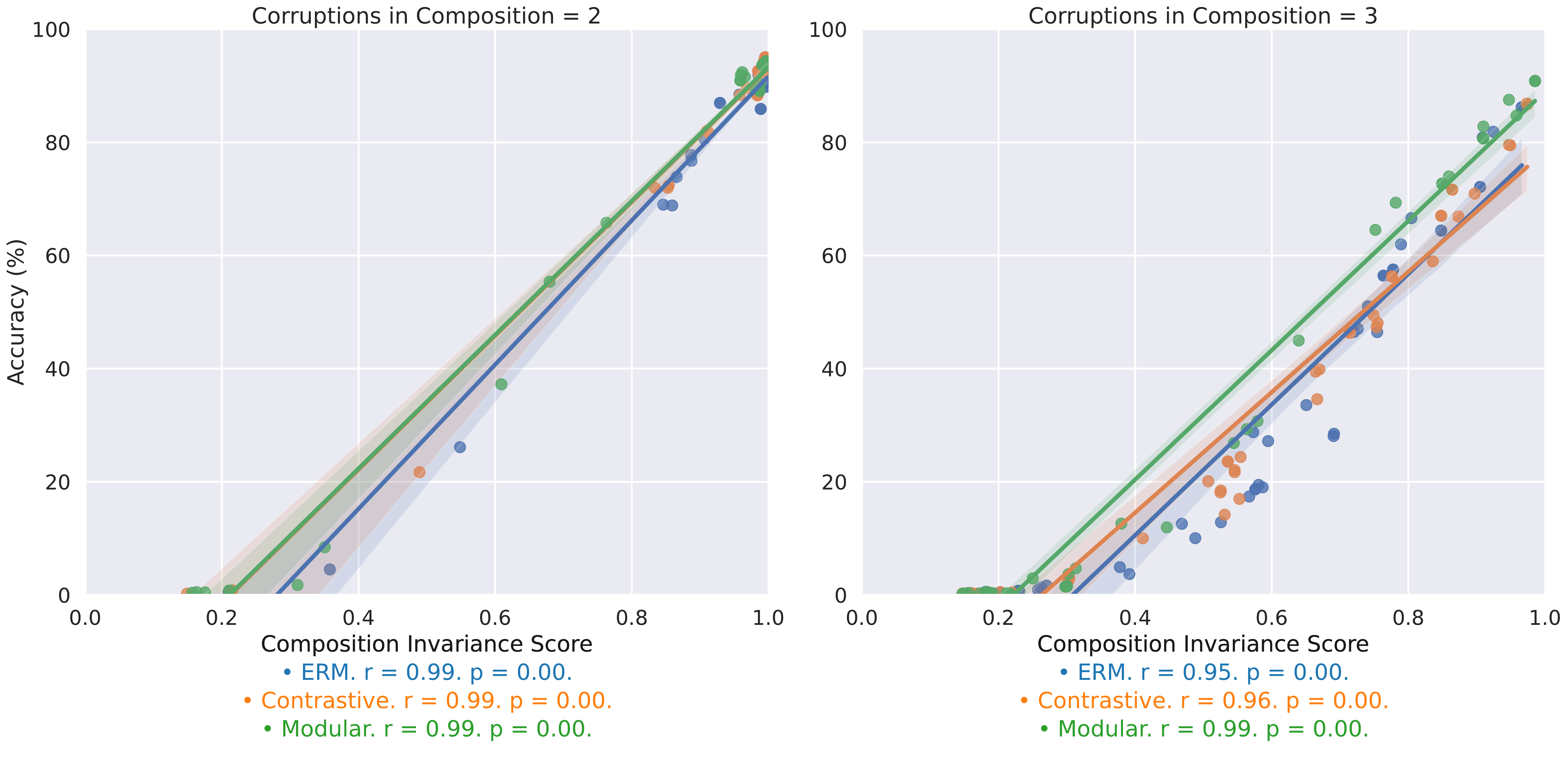} \\
    \end{subfigure}
    \vspace{-1.5mm}
    \caption{Correlating invariance scores with compositional robustness for \facescrub\ (second random seed). This is the same as Figure~\ref{fig:invariance-facescrub} with a different random seeding.}
    \label{fig:seed2-invariance-facescrub}
    \vspace{-6.5mm}
\end{figure}

\begin{figure}[tb]
    \centering
    % \vspace{-6mm}
    \begin{subfigure}[t]{0.32\textwidth}
        \centering
        \includegraphics[width=\linewidth]{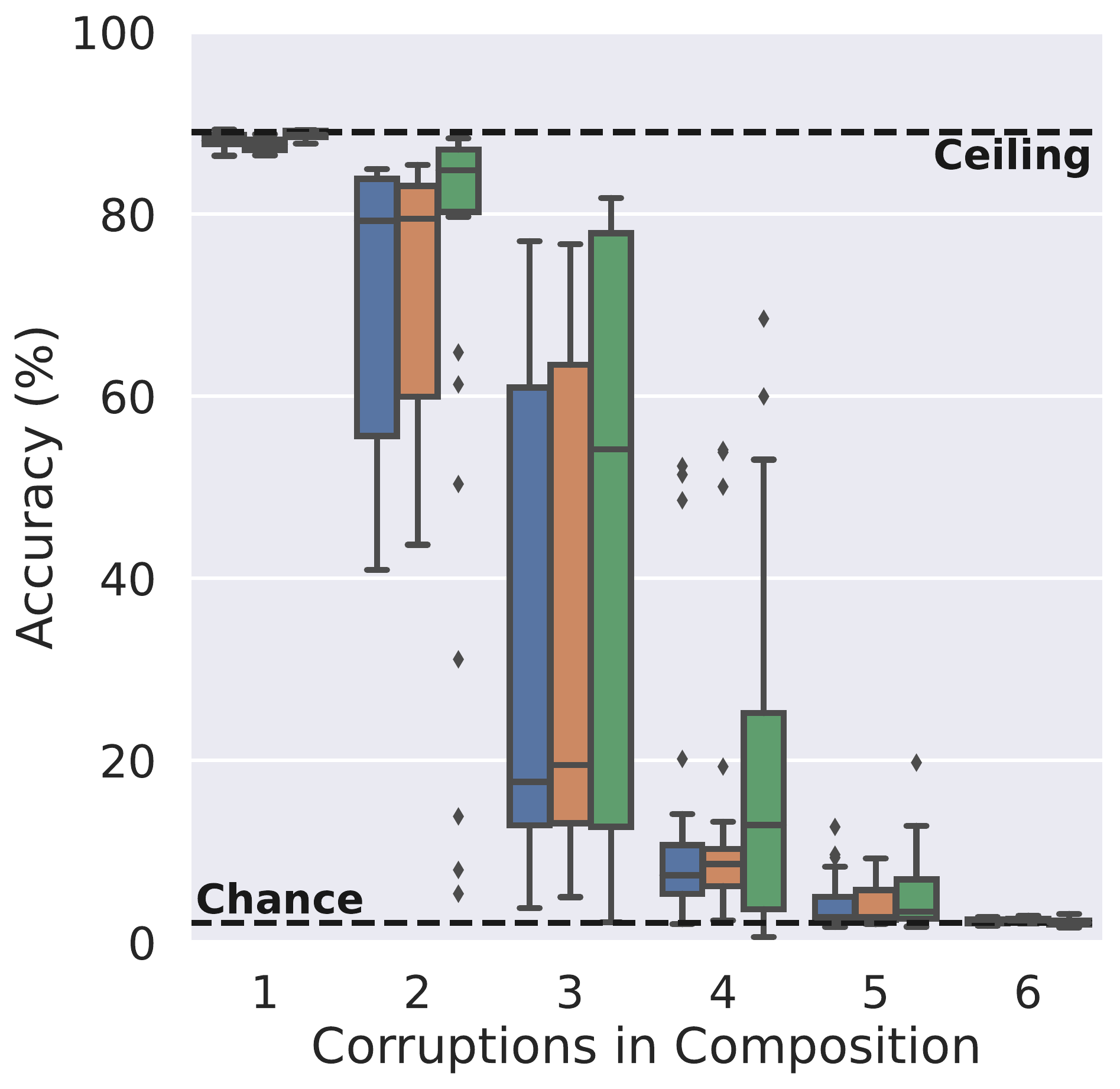} \\
        \caption{EMNIST}
        \label{fig:seed3-comp-em}
    \end{subfigure}%
    \hfill
    \begin{subfigure}[t]{0.32\textwidth}
        \centering
        \includegraphics[width=\linewidth]{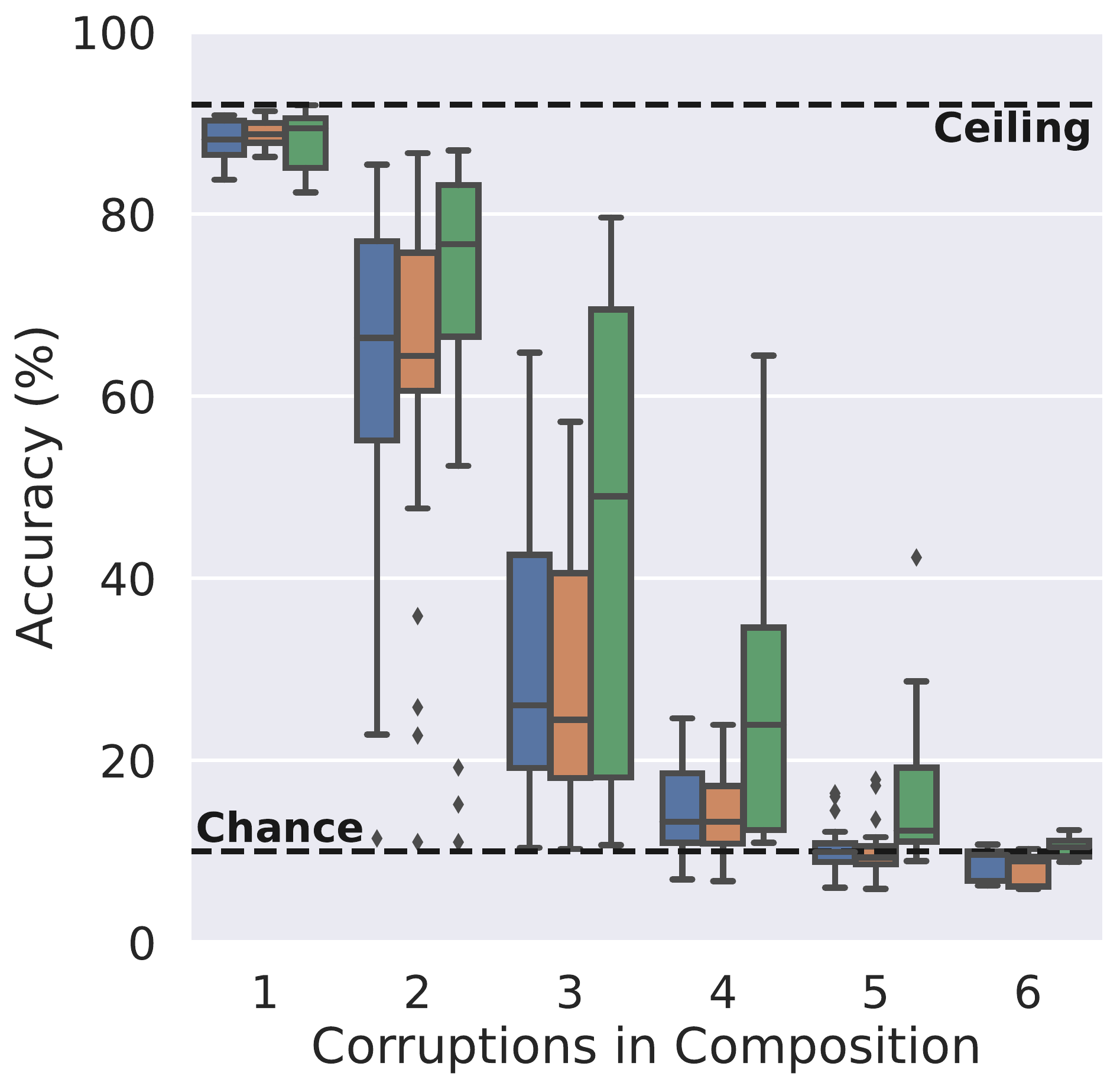} \\
        \caption{CIFAR-10}
        \label{fig:seed3-comp-cf}
    \end{subfigure}%
    \hfill
    \begin{subfigure}[t]{0.32\textwidth}
        \centering
        \includegraphics[width=\linewidth]{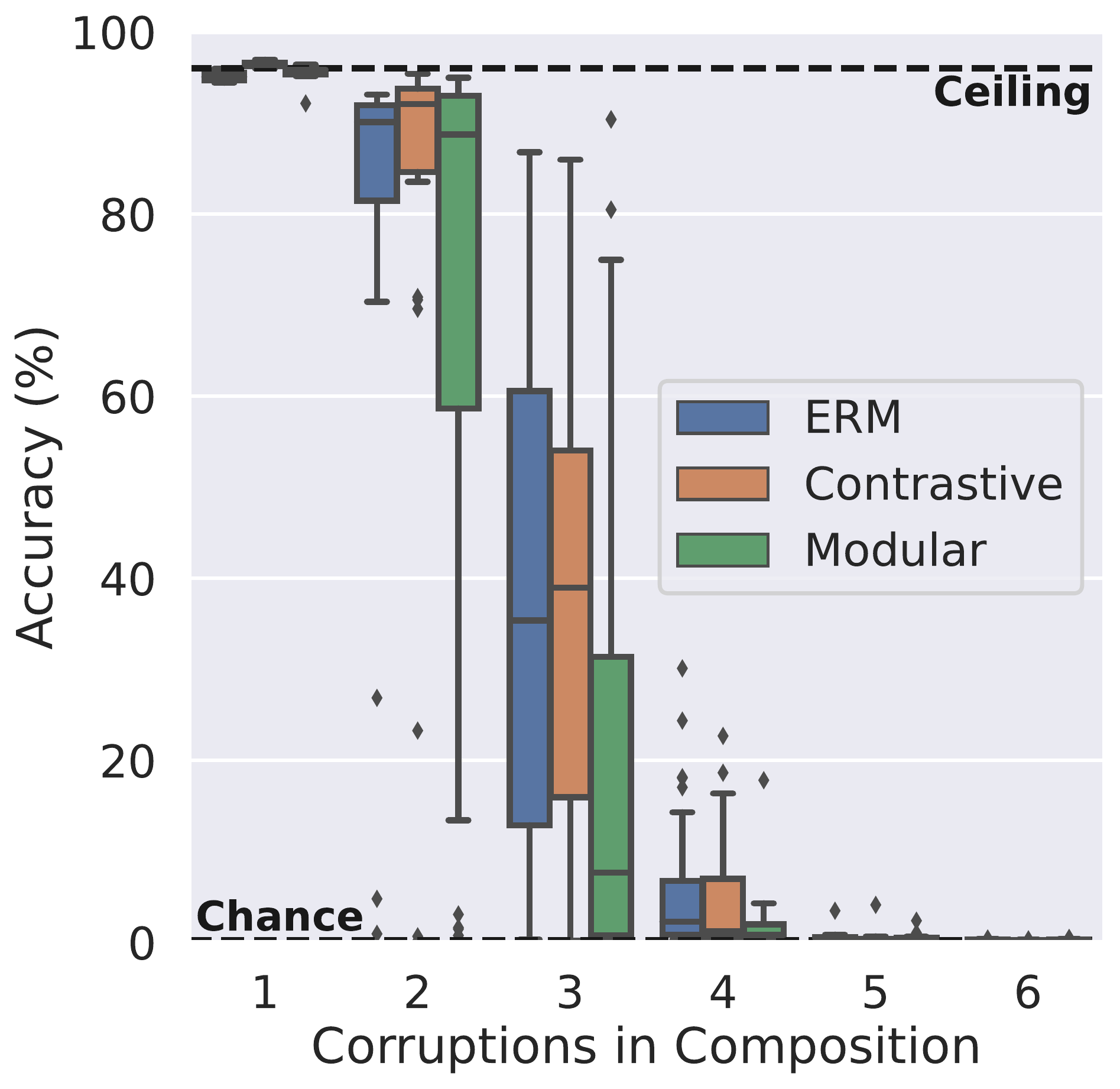} \\
        \caption{FACESCRUB}
        \label{fig:seed3-comp-fs}
    \end{subfigure}
    \vspace{-1.5mm}
    \caption{Evaluating compositional robustness on different datasets (third random seed). This figure is the same as Figure~\ref{fig:comp} with a different random seeding.}
    \label{fig:seed3-comp}
    \vspace{-4.5mm}
\end{figure}

\begin{figure}[tb]
    \centering
    \begin{subfigure}[t]{0.9\textwidth}
        \centering
        \includegraphics[trim={0 60cm 0 0},clip,width=\linewidth]{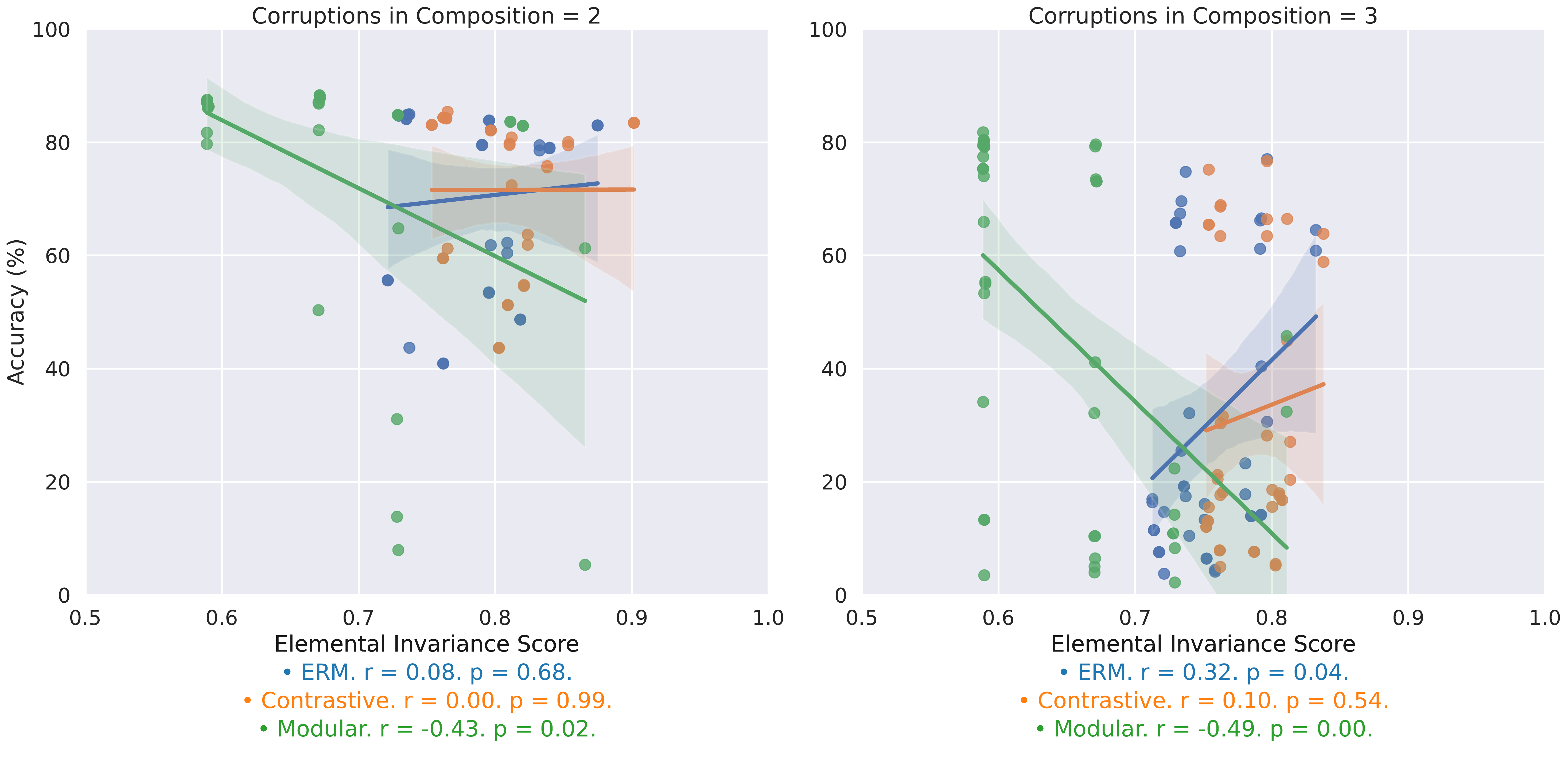} \\
    \end{subfigure}
    \begin{subfigure}[t]{0.9\textwidth}
        \centering
        \includegraphics[trim={0 60cm 0 0},clip,width=\linewidth]{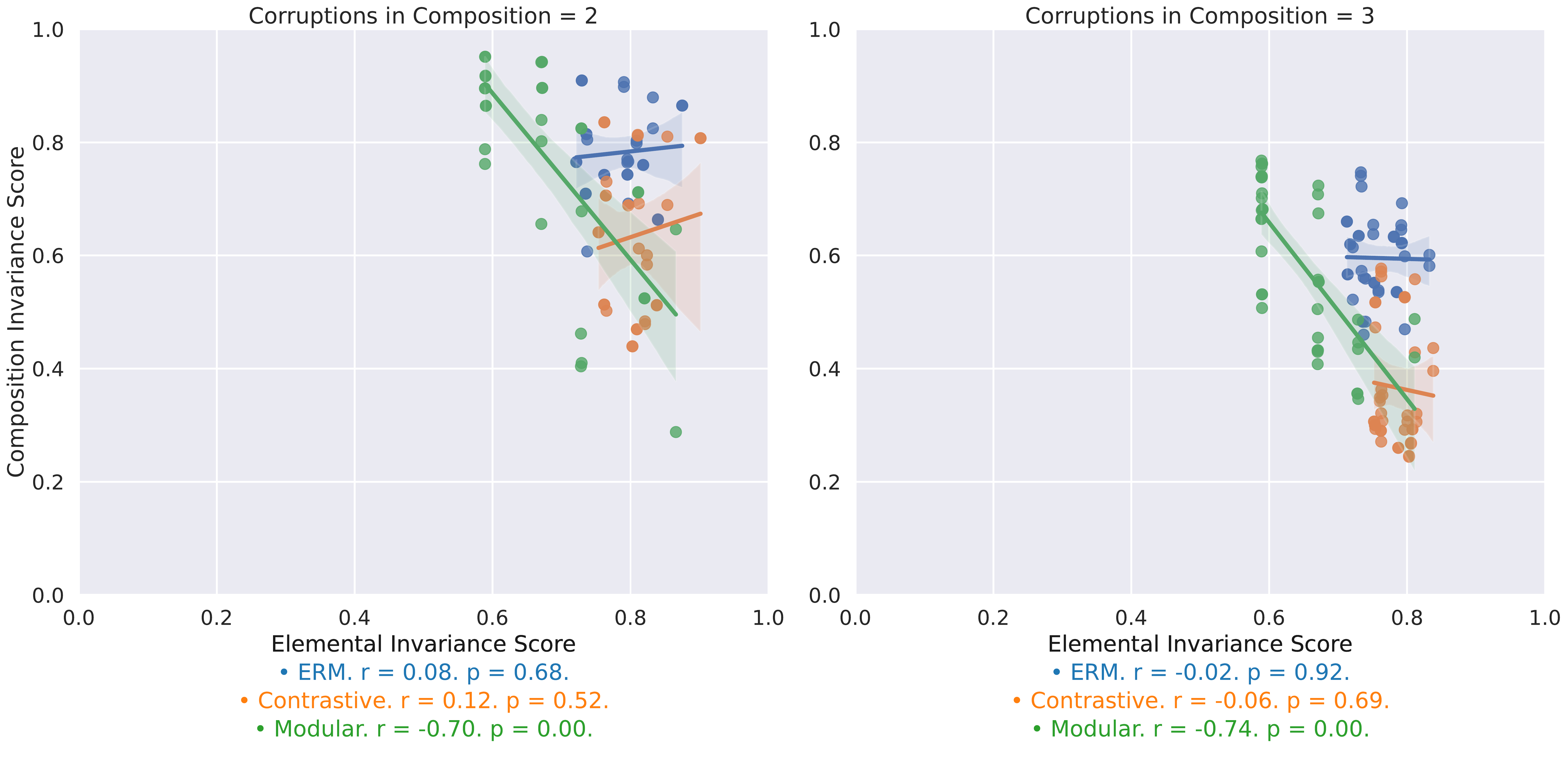} \\
    \end{subfigure}
    \begin{subfigure}[t]{0.9\textwidth}
        \centering
        \includegraphics[trim={0 60cm 0 0},clip,width=\linewidth]{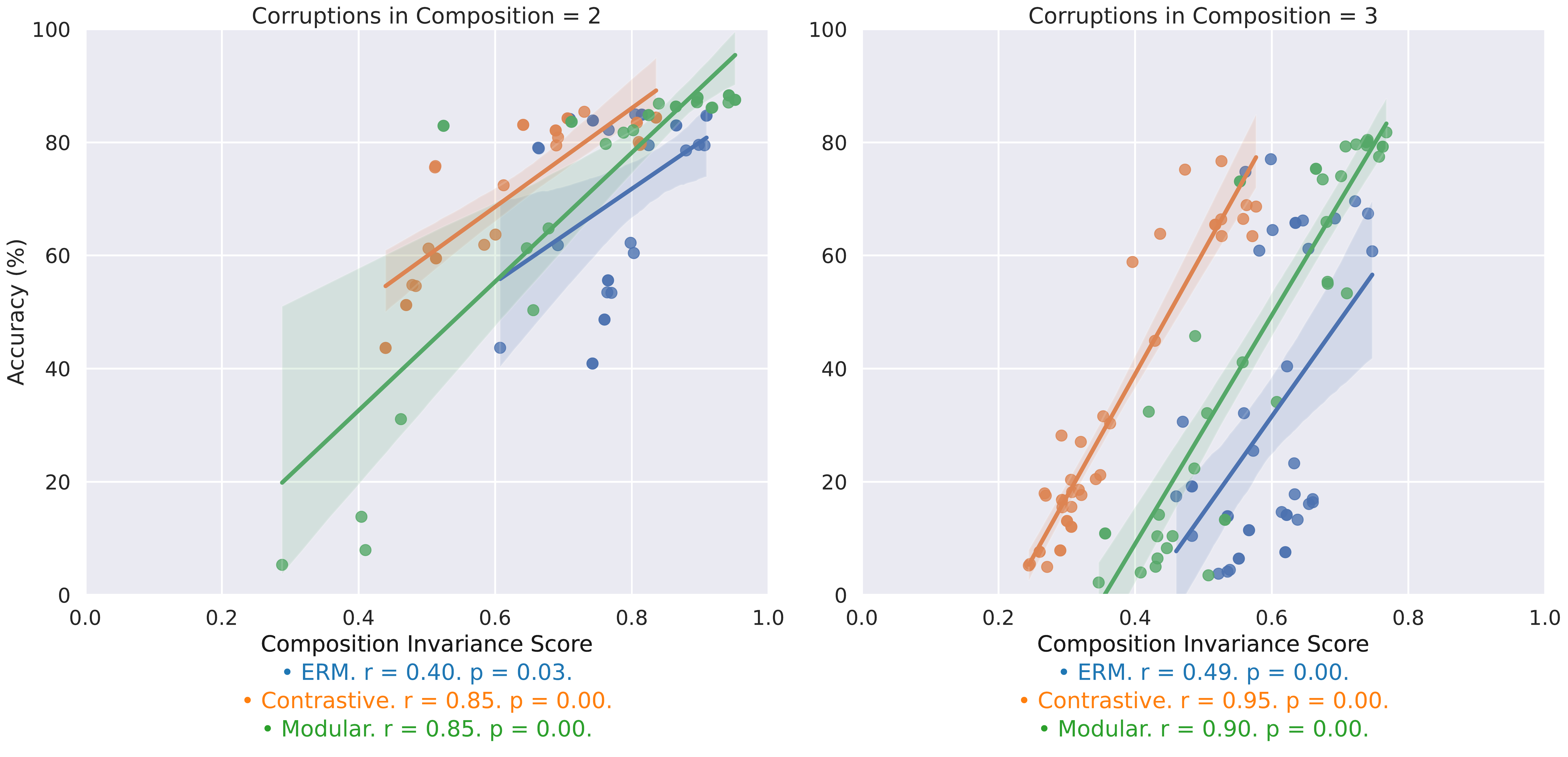} \\
    \end{subfigure}
    \vspace{-1.5mm}
    \caption{Correlating invariance scores with compositional robustness for \emnist\ (third random seed). This is the same as Figure~\ref{fig:invariance-emnist} with a different random seeding.}
    \label{fig:seed3-invariance-emnist}
    \vspace{-6.5mm}
\end{figure}

\begin{figure}[tb]
    \centering
    \begin{subfigure}[t]{0.9\textwidth}
        \centering
        \includegraphics[trim={0 60cm 0 0},clip,width=\linewidth]{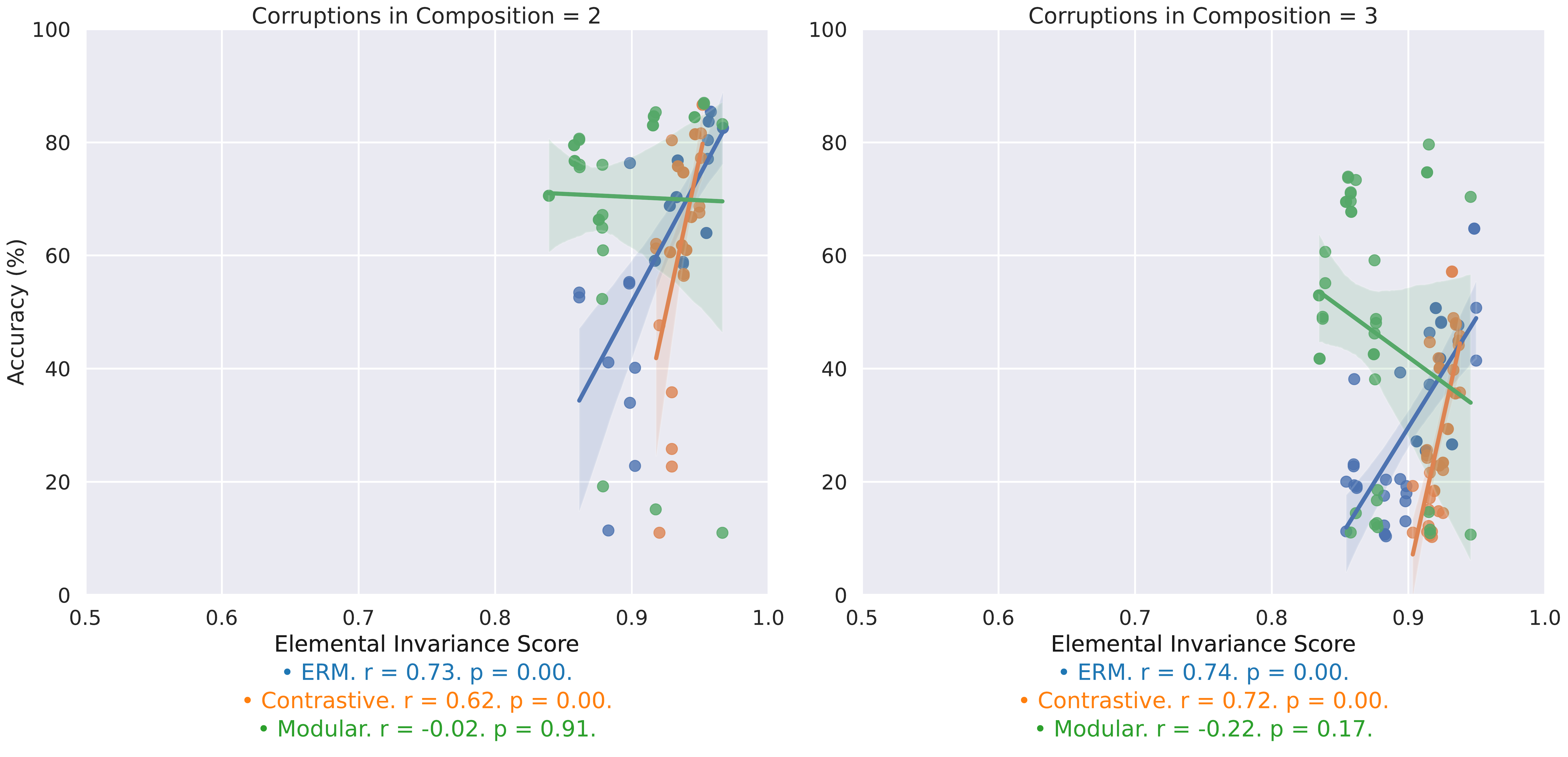} \\
    \end{subfigure}
    \begin{subfigure}[t]{0.9\textwidth}
        \centering
        \includegraphics[trim={0 60cm 0 0},clip,width=\linewidth]{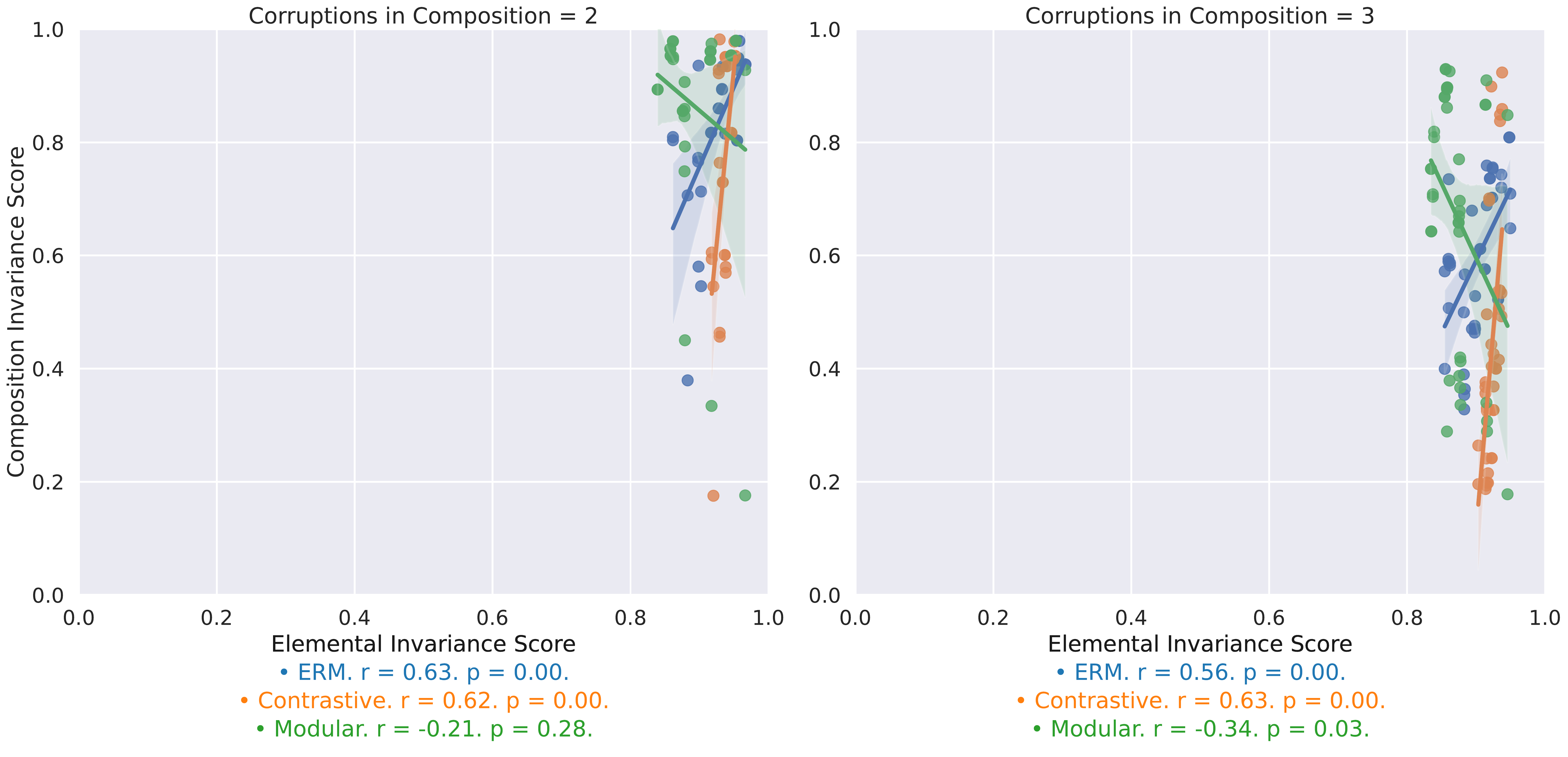} \\
    \end{subfigure}
    \begin{subfigure}[t]{0.9\textwidth}
        \centering
        \includegraphics[trim={0 60cm 0 0},clip,width=\linewidth]{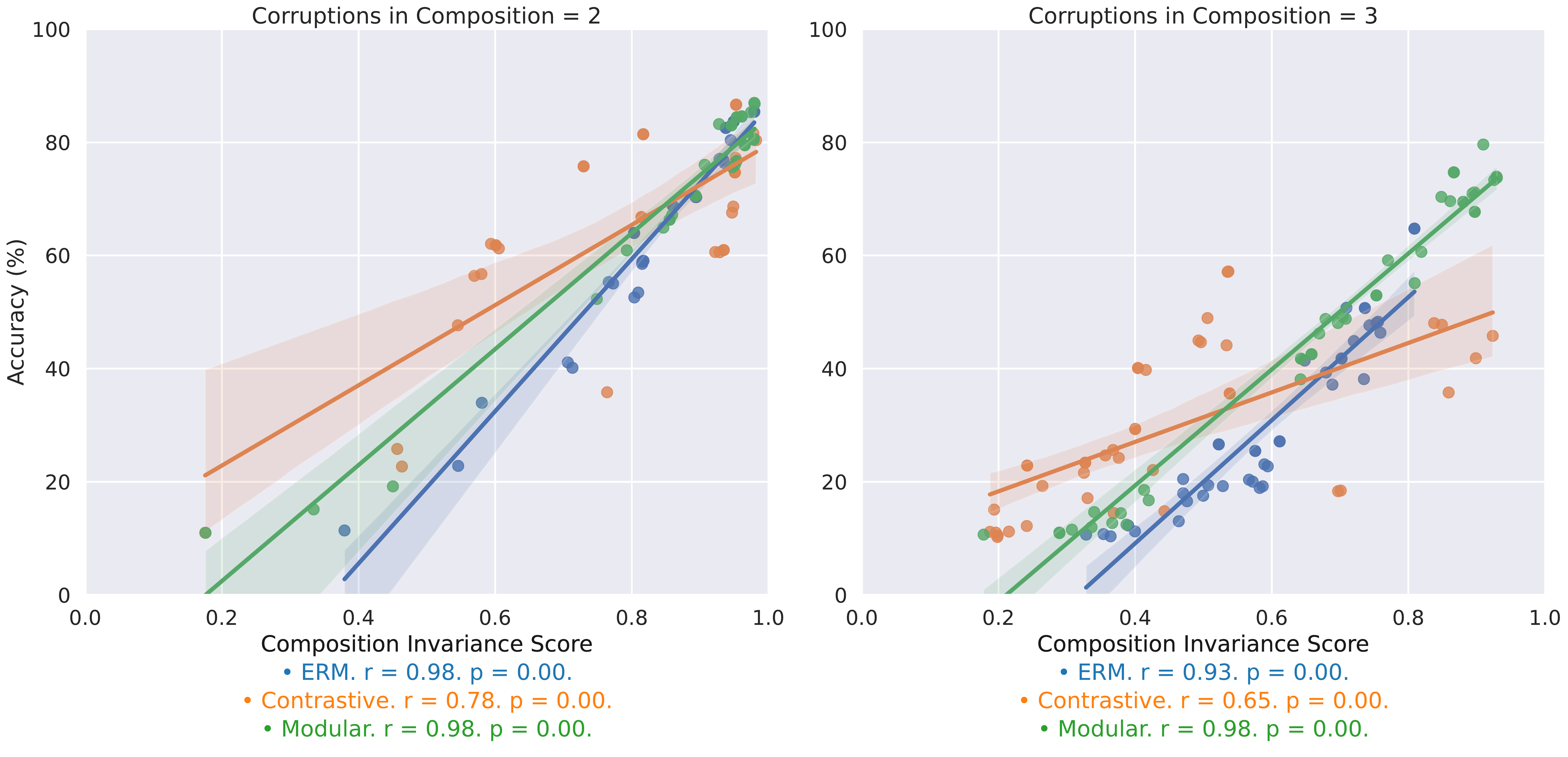} \\
    \end{subfigure}
    \vspace{-1.5mm}
    \caption{Correlating invariance scores with compositional robustness for \cifar\ (second random seed). This is the same as Figure~\ref{fig:invariance-cifar} with a different random seeding.}
    \label{fig:seed3-invariance-cifar}
    \vspace{-6.5mm}
\end{figure}

\begin{figure}[tb]
    \centering
    \begin{subfigure}[t]{0.9\textwidth}
        \centering
        \includegraphics[trim={0 60cm 0 0},clip,width=\linewidth]{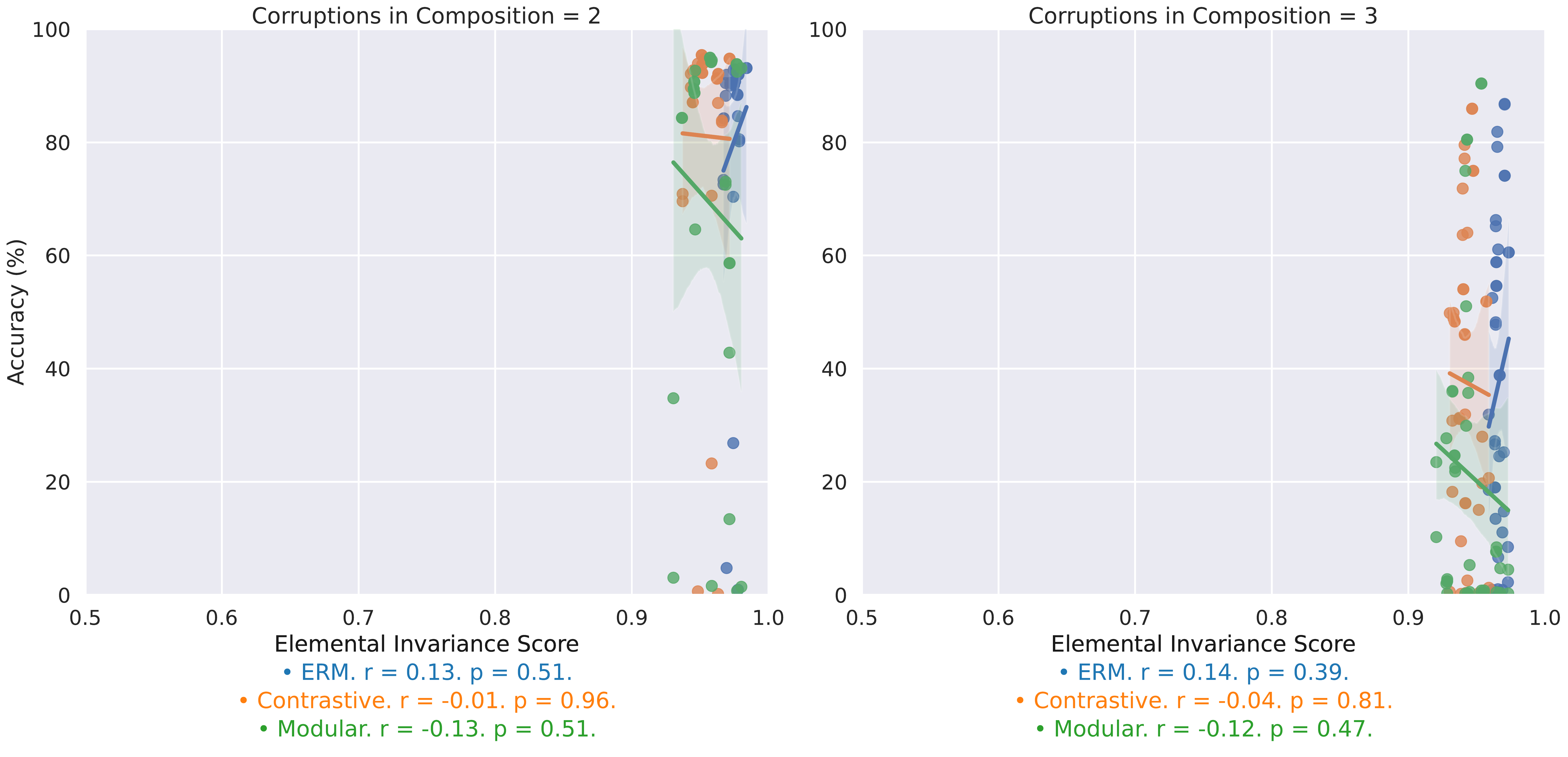} \\
    \end{subfigure}
    \begin{subfigure}[t]{0.9\textwidth}
        \centering
        \includegraphics[trim={0 60cm 0 0},clip,width=\linewidth]{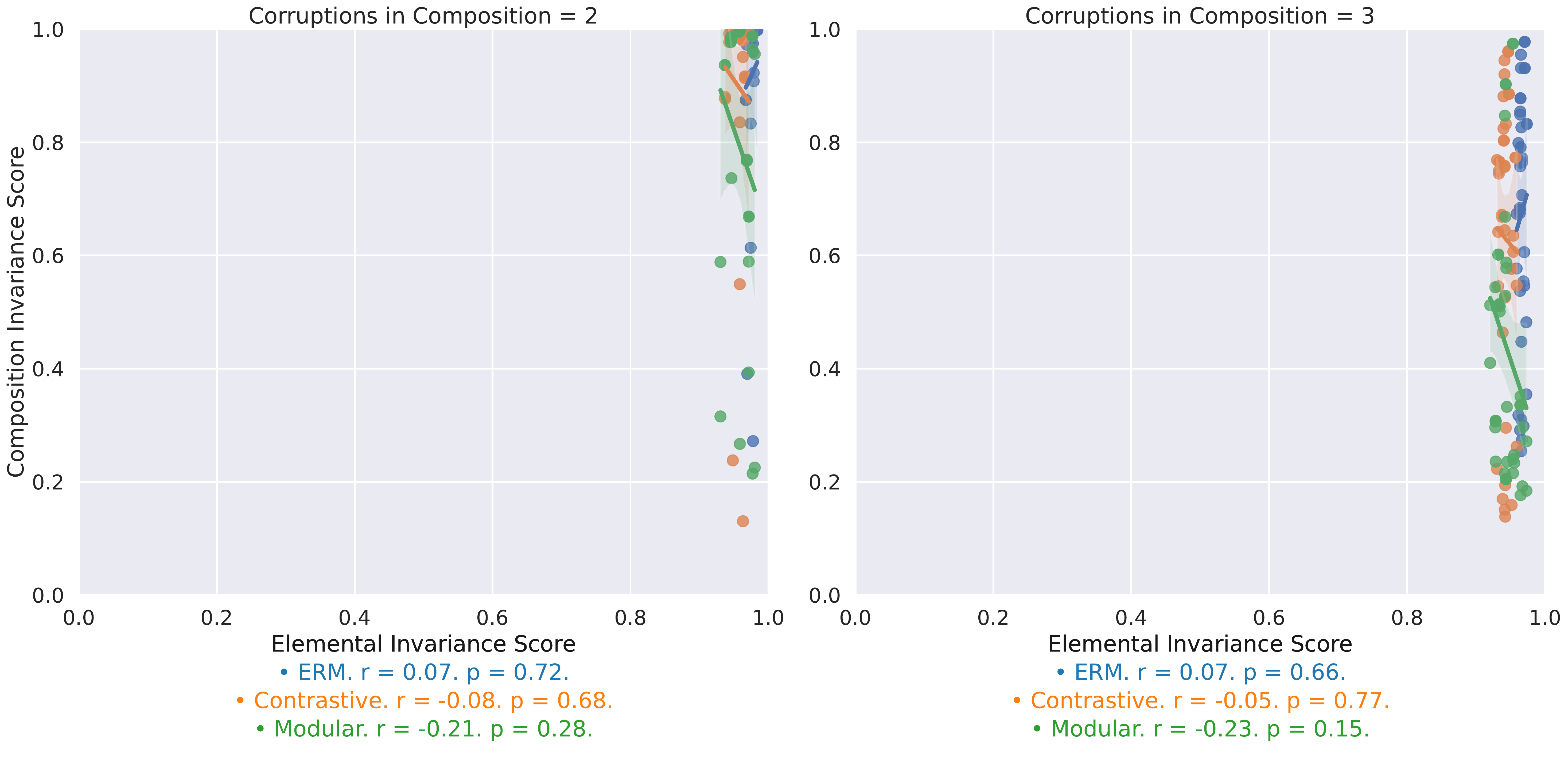} \\
    \end{subfigure}
    \begin{subfigure}[t]{0.9\textwidth}
        \centering
        \includegraphics[trim={0 60cm 0 0},clip,width=\linewidth]{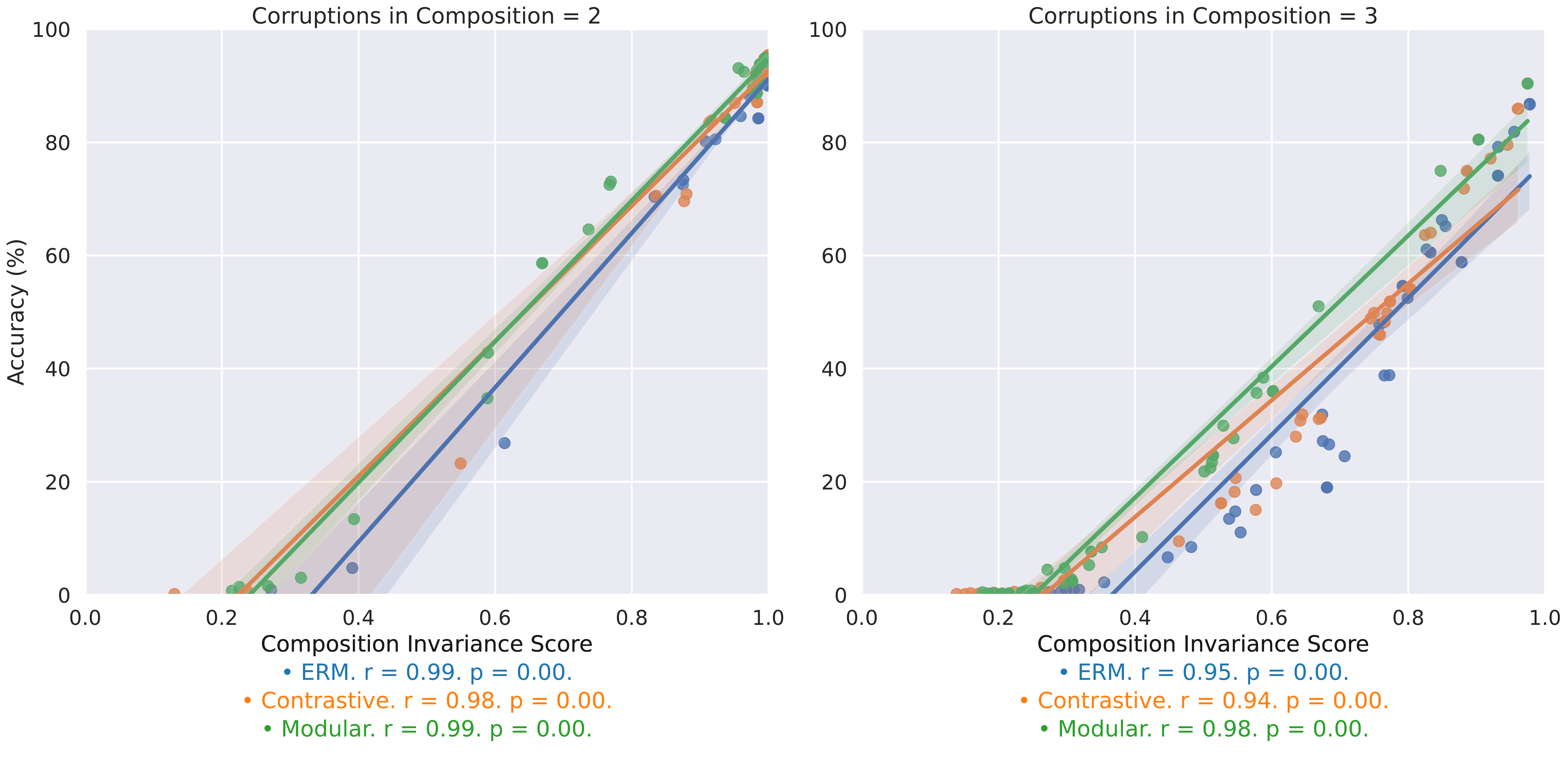} \\
    \end{subfigure}
    \vspace{-1.5mm}
    \caption{Correlating invariance scores with compositional robustness for \facescrub\ (second random seed). This is the same as Figure~\ref{fig:invariance-facescrub} with a different random seeding.}
    \label{fig:seed3-invariance-facescrub}
    \vspace{-6.5mm}
\end{figure}

This appendix shows the results included in the main text for two further random seeds. In particular we replicate Figures~\ref{fig:comp}, \ref{fig:invariance-emnist}, \ref{fig:invariance-cifar} and \ref{fig:invariance-facescrub} in each case. The results in the main text come from the first random seed, Figures~\ref{fig:seed2-comp}-\ref{fig:seed2-invariance-facescrub} show the second random seed and Figures~\ref{fig:seed3-comp}-\ref{fig:seed3-invariance-facescrub} show the third random seed.

% \subsection{Seed 2}

% \subsection{Seed 3}

\section{Heat Maps - Full Granular Results}
\label{app:heat-maps}

\begin{figure}
    \centering
    \includegraphics[trim={0 0 69.4cm 0}, width=\textwidth]{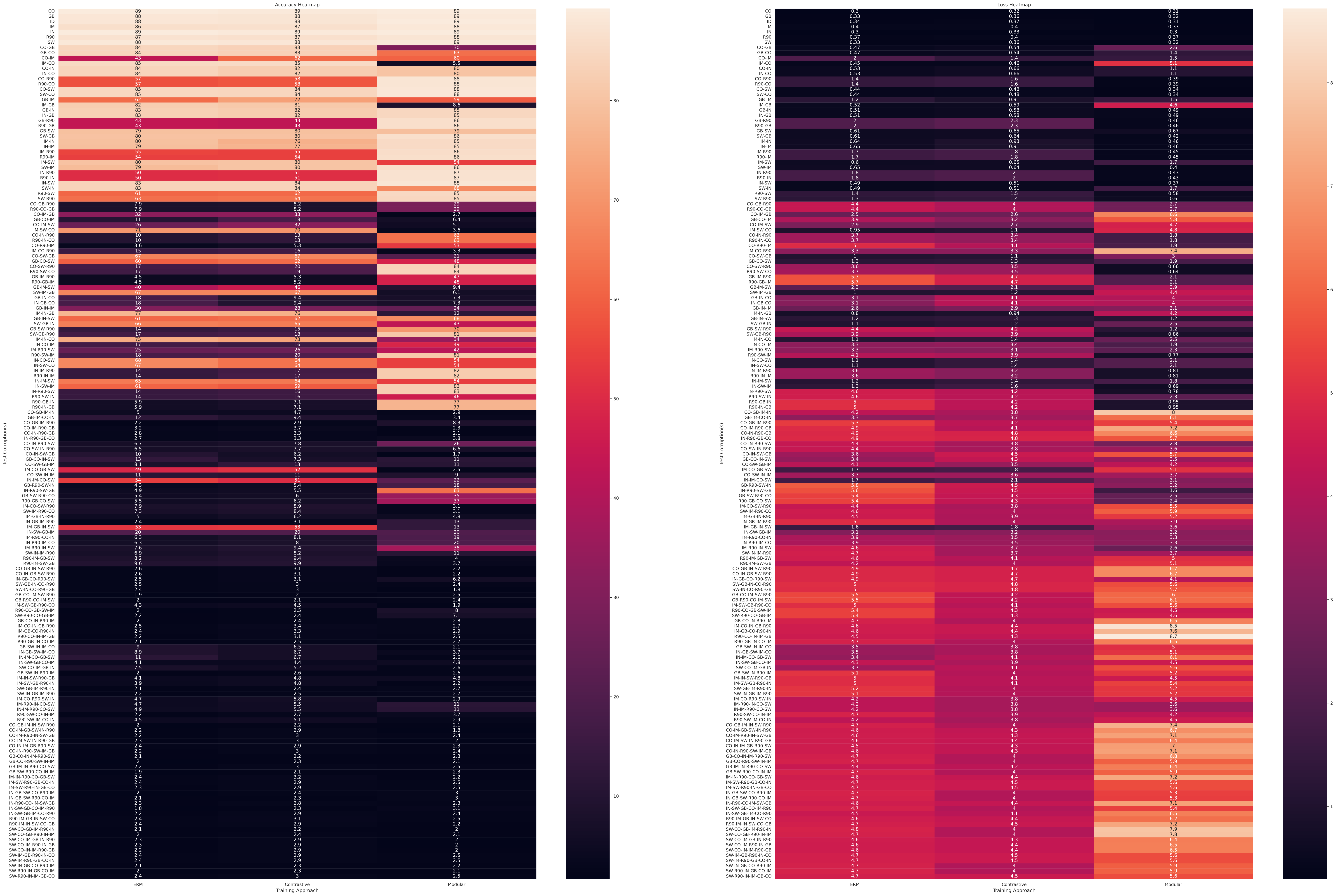}
    \caption{Per-domain heat map for \emnist\ (first random seed). This shows the raw \emnist\ data from Figure~\ref{fig:comp}. Best viewed with zoom.}
    \label{fig:heat-map-emnist}
\end{figure}

\begin{figure}
    \centering
    \includegraphics[trim={0 0 69.4cm 0}, width=\textwidth]{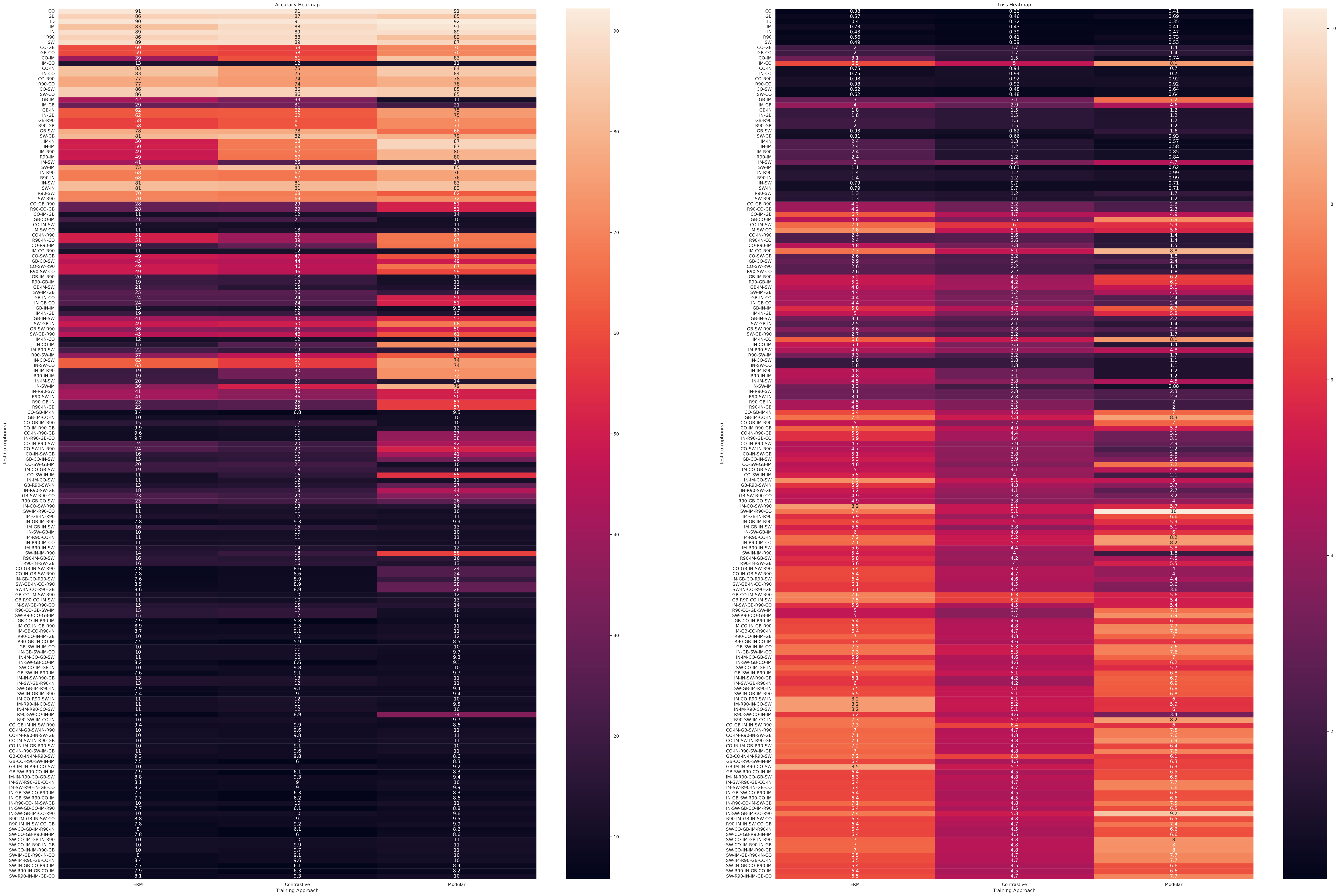}
    \caption{Per-domain heat map for \cifar\ (first random seed). This shows the raw \cifar\ data from Figure~\ref{fig:comp}. Best viewed with zoom.}
    \label{fig:heat-map-cifar}
\end{figure}

\begin{figure}
    \centering
    \includegraphics[trim={0 0 69.4cm 0}, width=\textwidth]{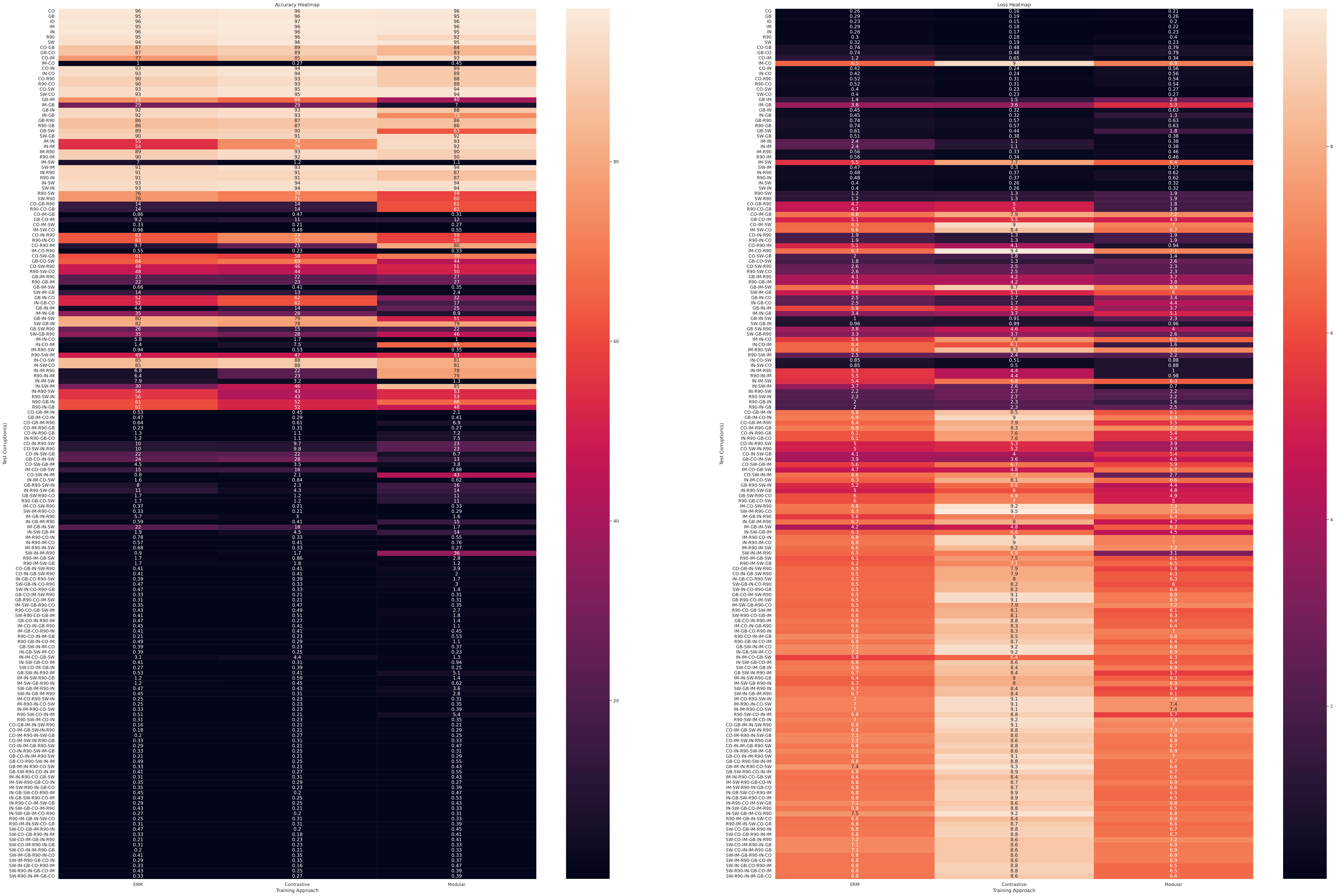}
    \caption{Per-domain heat map for \facescrub\ (first random seed). This shows the raw \facescrub\ data from Figure~\ref{fig:comp}. Best viewed with zoom.}
    \label{fig:heat-map-facescrub}
\end{figure}

\begin{figure}
    \centering
    \includegraphics[trim={0 0 69.4cm 0}, width=\textwidth]{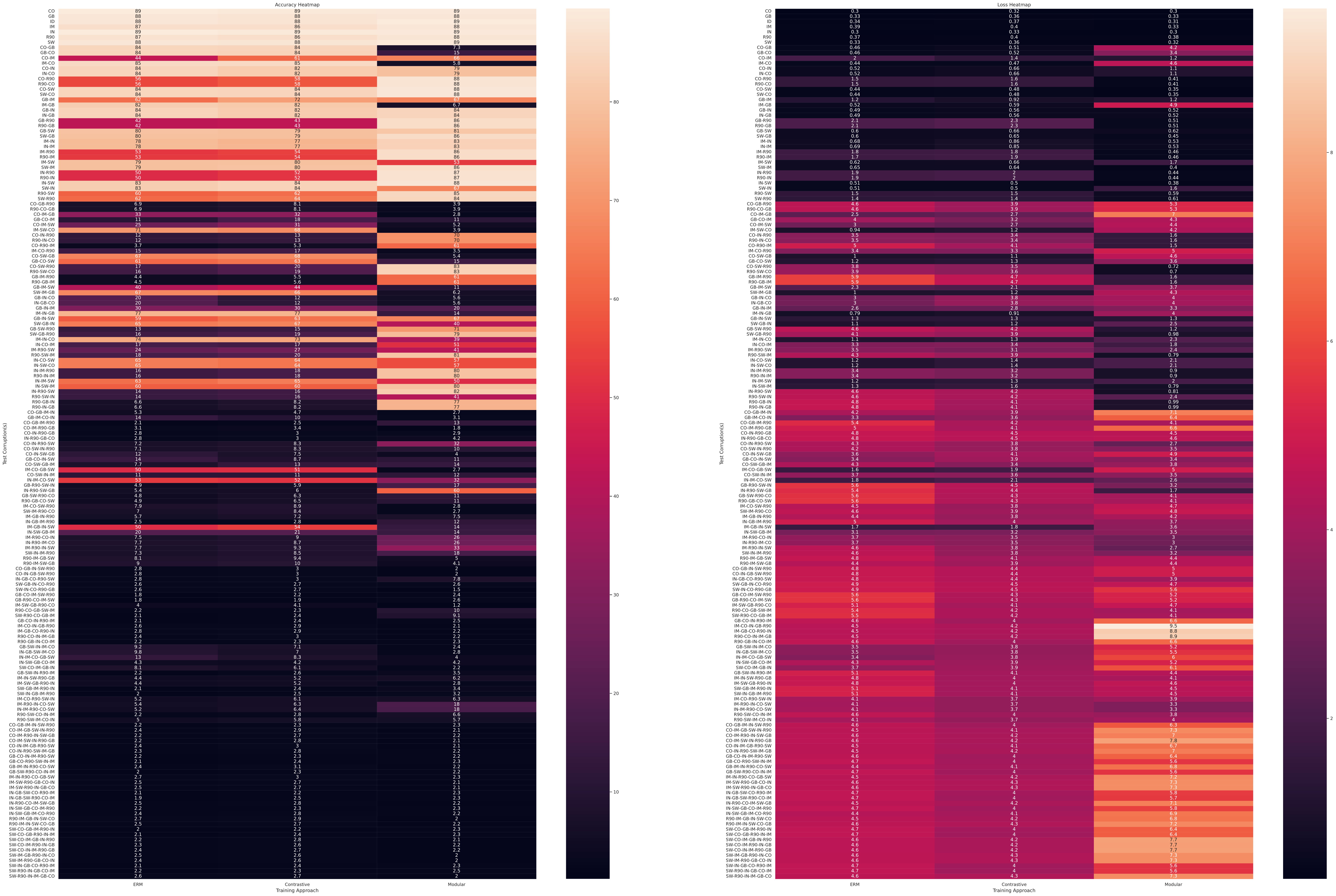}
    \caption{Per-domain heat map for \emnist\ (second random seed). This shows the raw \emnist\ data from Figure~\ref{fig:seed2-comp}. Best viewed with zoom.}
    \label{fig:seed2-heat-map-emnist}
\end{figure}

\begin{figure}
    \centering
    \includegraphics[trim={0 0 69.4cm 0}, width=\textwidth]{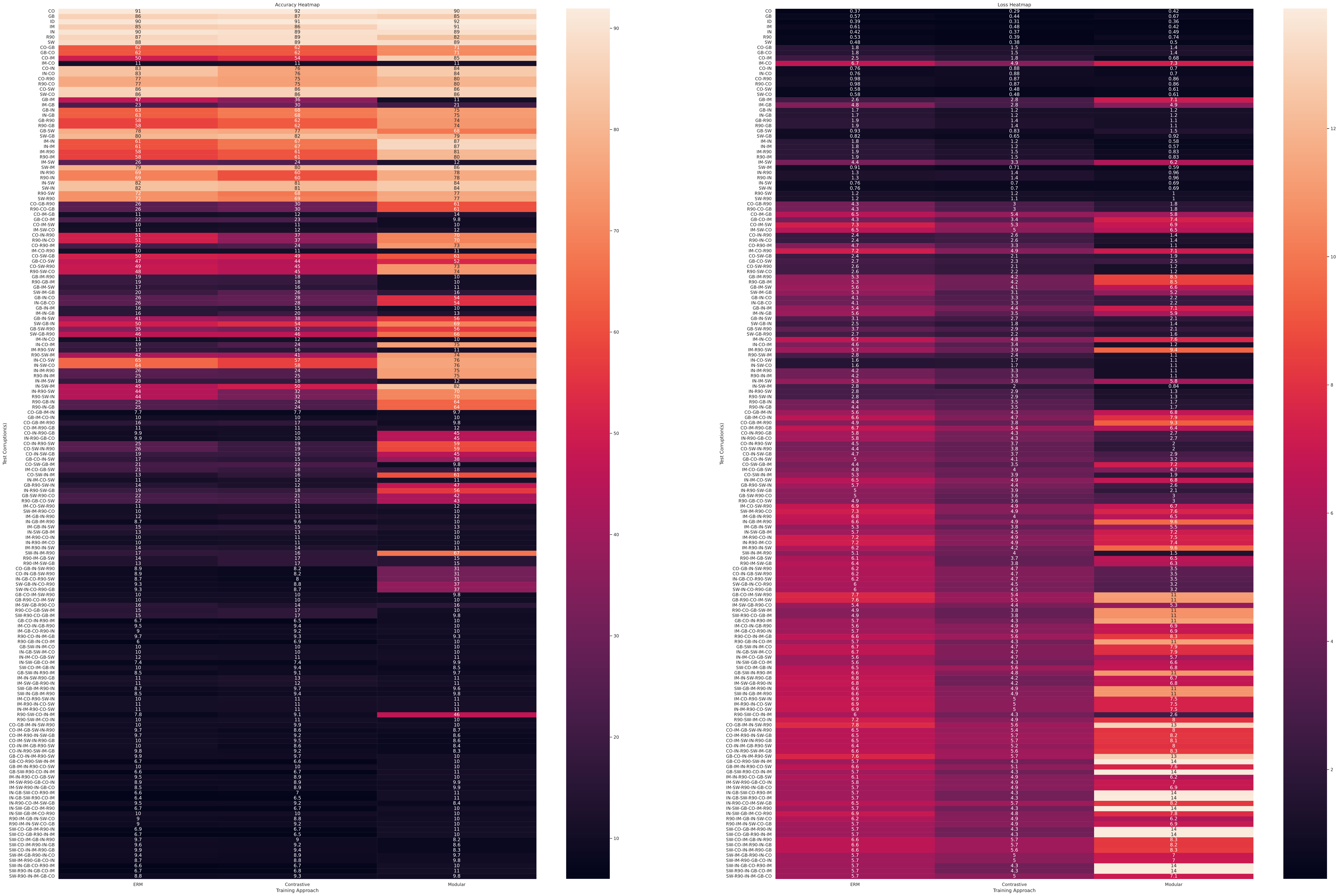}
    \caption{Per-domain heat map for \cifar\ (second random seed). This shows the raw \cifar\ data from Figure~\ref{fig:seed2-comp}. Best viewed with zoom.}
    \label{fig:seed2-heat-map-cifar}
\end{figure}

\begin{figure}
    \centering
    \includegraphics[trim={0 0 69.4cm 0}, width=\textwidth]{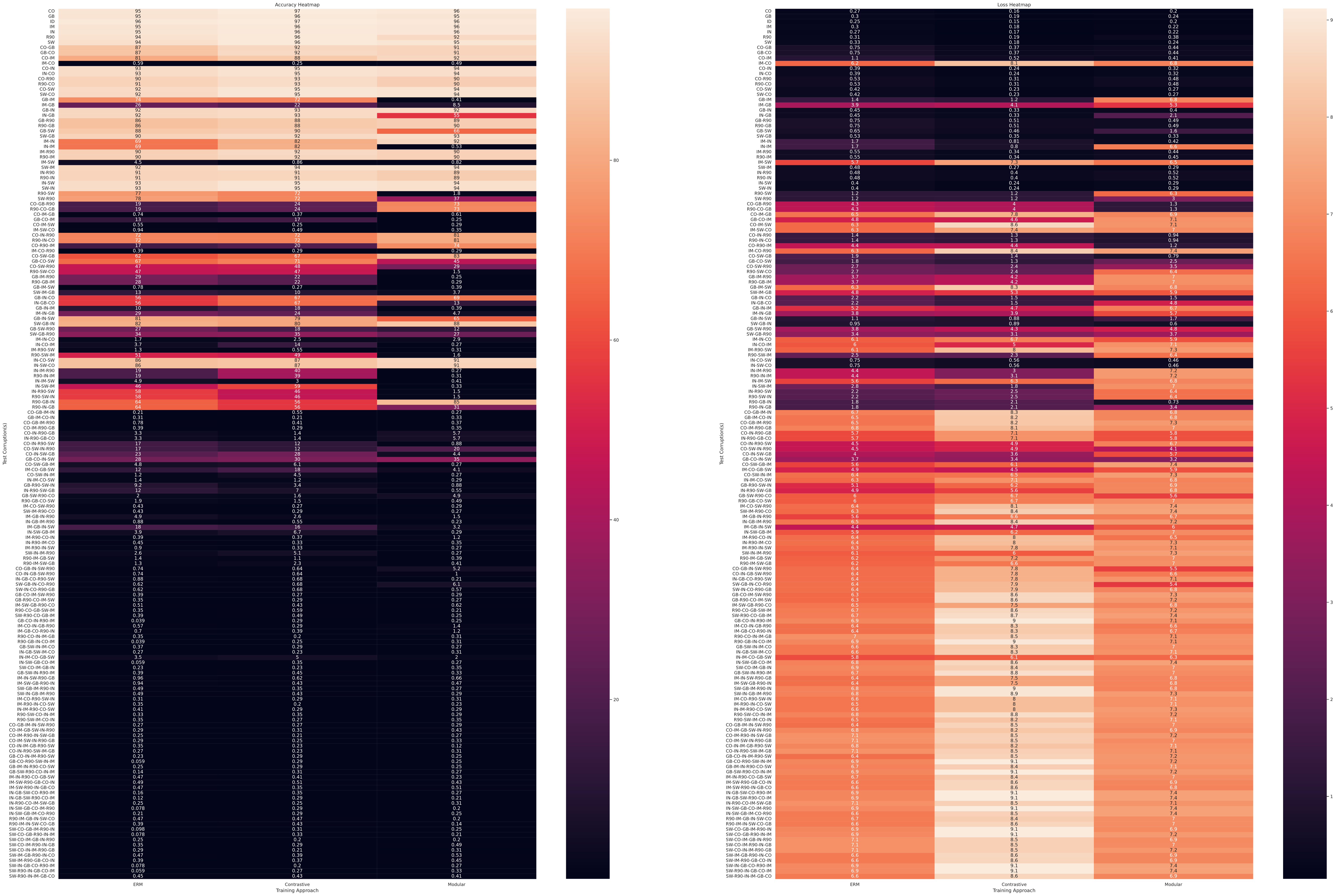}
    \caption{Per-domain heat map for \facescrub\ (second random seed). This shows the raw \facescrub\ data from Figure~\ref{fig:seed2-comp}. Best viewed with zoom.}
    \label{fig:seed2-heat-map-facescrub}
\end{figure}

\begin{figure}
    \centering
    \includegraphics[trim={0 0 69.4cm 0}, width=\textwidth]{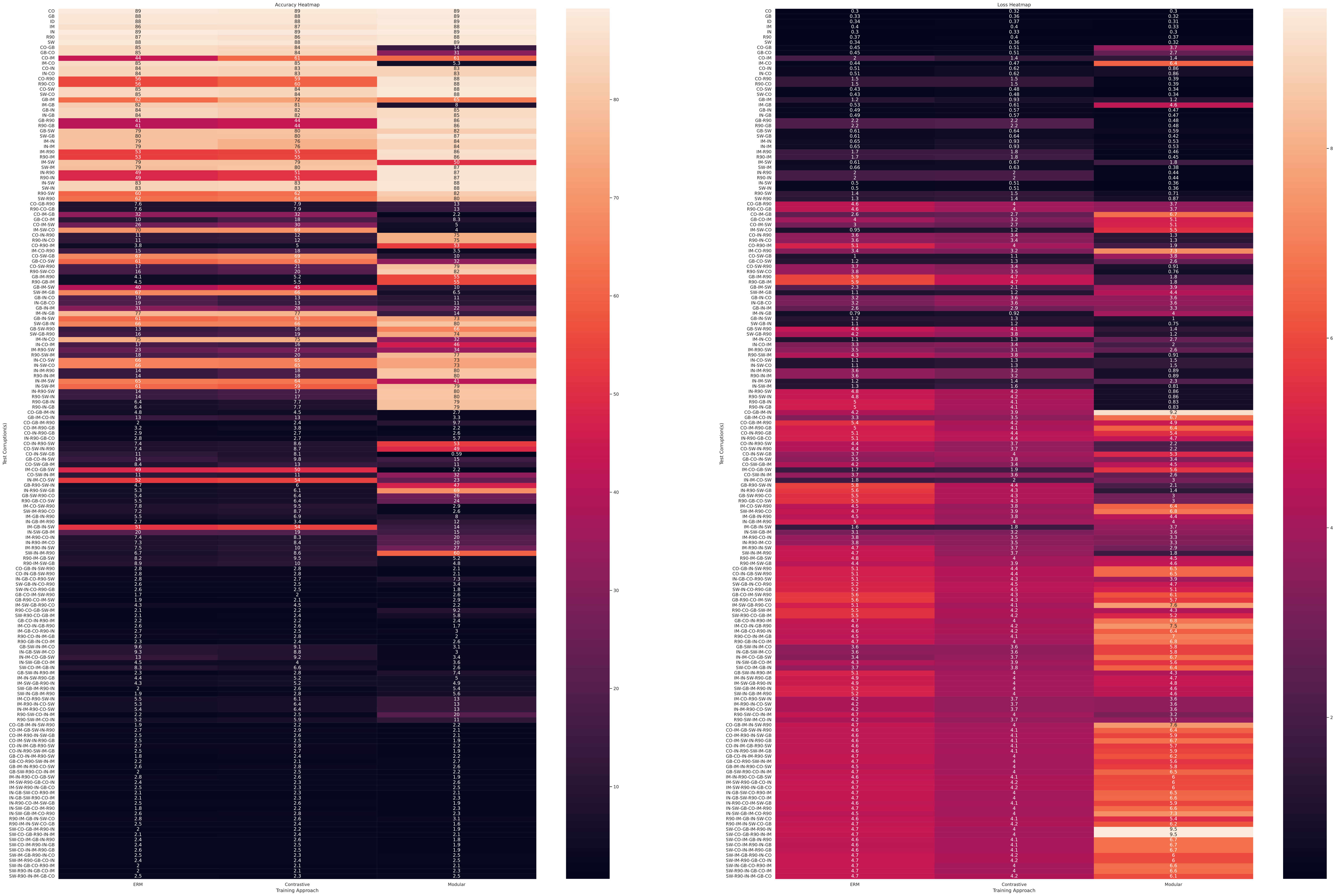}
    \caption{Per-domain heat map for \emnist\ (third random seed). This shows the raw \emnist\ data from Figure~\ref{fig:seed3-comp}. Best viewed with zoom.}
    \label{fig:seed3-heat-map-emnist}
\end{figure}

\begin{figure}
    \centering
    \includegraphics[trim={0 0 69.4cm 0}, width=\textwidth]{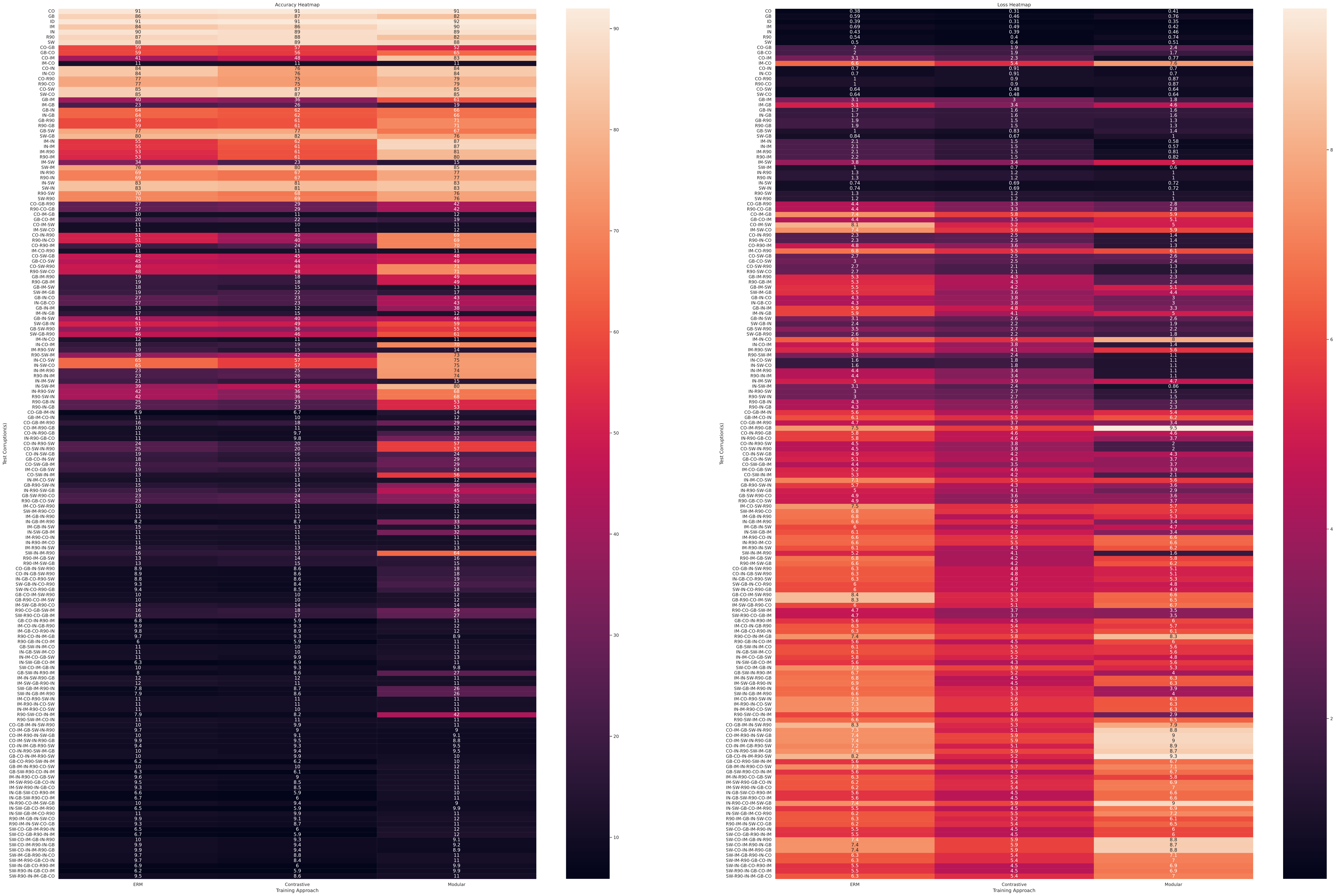}
    \caption{Per-domain heat map for \cifar\ (third random seed). This shows the raw \cifar\ data from Figure~\ref{fig:seed3-comp}. Best viewed with zoom.}
    \label{fig:seed3-heat-map-cifar}
\end{figure}

\begin{figure}
    \centering
    \includegraphics[trim={0 0 69.4cm 0}, width=\textwidth]{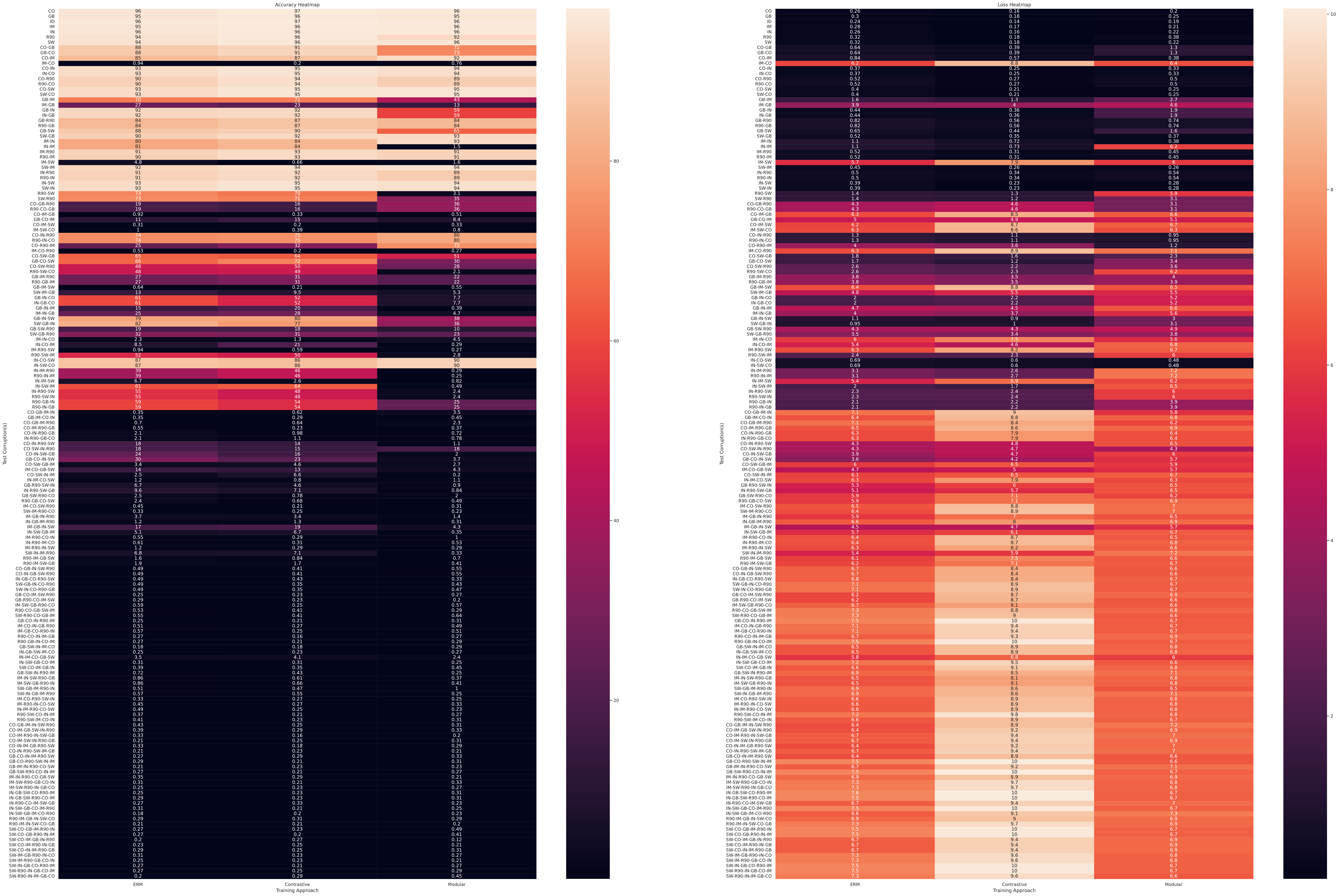}
    \caption{Per-domain heat map for \facescrub\ (third random seed). This shows the raw \facescrub\ data from Figure~\ref{fig:seed3-comp}. Best viewed with zoom.}
    \label{fig:seed3-heat-map-facescrub}
\end{figure}

Finally we show granular results, showing the individual accuracy for every elemental corruption and composition. This is the raw data that is summarized by the box plots in Figures~\ref{fig:comp}, \ref{fig:seed2-comp} and \ref{fig:seed3-comp}. We show heat maps for every dataset and every seed in Figures~\ref{fig:heat-map-emnist}-\ref{fig:seed3-heat-map-facescrub} to give a per-domain view of the differences in behaviors for the different methods for compositional robustness. 

% \subsection{Seed 1}

% \subsection{Seed 2}

% \subsection{Seed 3}

\end{document}